\documentclass[10pt,twocolumn,letterpaper]{article}

\usepackage{wacv}
\usepackage{times}
\usepackage{epsfig}
\usepackage{graphicx}
\usepackage{amsmath}
\usepackage{amssymb}
\usepackage{booktabs}
\usepackage{algorithm}
\usepackage{algpseudocode}
\usepackage{multirow}
\usepackage{pifont}
\usepackage{subcaption}
\usepackage[accsupp]{axessibility}  
\newcommand\blfootnote[1]{%
  \begingroup
  \renewcommand\thefootnote{}\footnote{#1}%
  \addtocounter{footnote}{-1}%
  \endgroup
}

\newcommand{\cmark}{\ding{51}}%
\newcommand{\xmark}{\ding{55}}%

\newcommand{\high}[1]{{\it (#1) }}
\newcommand{\method}{Re:NeRF}
\newcommand{\lightrule}{\specialrule{0.001em}{0em}{0em}}

\def\wacvPaperID{1521} 

\wacvalgorithmstrack   

\wacvfinalcopy 

\ifwacvfinal
\usepackage[breaklinks=true,bookmarks=false]{hyperref}
\else
\usepackage[pagebackref=true,breaklinks=true,colorlinks,bookmarks=false]{hyperref}
\fi

\begin{document}

\title{Compressing Explicit Voxel Grid Representations: fast NeRFs become also small}

\author{Chenxi Lola Deng \qquad Enzo Tartaglione\\~\\
LTCI, T\'el\'ecom Paris, Institut Polytechnique de Paris\\
}

\maketitle
\thispagestyle{empty}

\begin{abstract}
NeRFs have revolutionized the world of per-scene radiance field reconstruction because of their intrinsic compactness. One of the main limitations of NeRFs is their slow rendering speed, both at training and inference time. Recent research focuses on the optimization of an explicit voxel grid (EVG) that represents the scene, which can be paired with neural networks to learn radiance fields. This approach significantly enhances the speed both at train and inference time, but at the cost of large memory occupation.\\
   In this work we propose \method, an approach that specifically targets EVG-NeRFs compressibility, aiming to reduce memory storage of NeRF models while maintaining comparable performance. We benchmark our approach with three different EVG-NeRF architectures on four popular benchmarks, showing \method 's broad usability and effectiveness. \blfootnote{This paper has been accepted for publication at WACV~2023.}
\end{abstract}

\section{Introduction}
\label{sec:intro}

The rising of Neural Radiance Fields (NeRF) techniques has heavily impacted the field of 3D scene modeling and reconstruction in recent years~\cite{mildenhall2020nerf, yu2021pixelnerf, huang2022stylizednerf, kaya2022neural, guo2022fast}. Efficient photo-realistic novel view generation from a fixed set of training images has been a popular area of research in computer vision with broad applications. The ability to distill the essence of the 3D object from 2D representations of it and its compactness is the main reason for making NeRF a high-impact approach in the literature.

The original NeRF~\cite{mildenhall2020nerf} consists of a multi-layer perceptron, which implicitly learns the manifold representing the 3D object. Because of its great generalization for synthesizing novel viewpoints and the high compactness of the model itself, which typically consists of a few MB, NeRF has become a prevalent approach for 3D reconstruction. However, the NeRF's MLP has to be queried million times to render a scene, leading to slow training and rendering time.

\begin{figure}[t!]
\centering
    \includegraphics[width=\columnwidth]{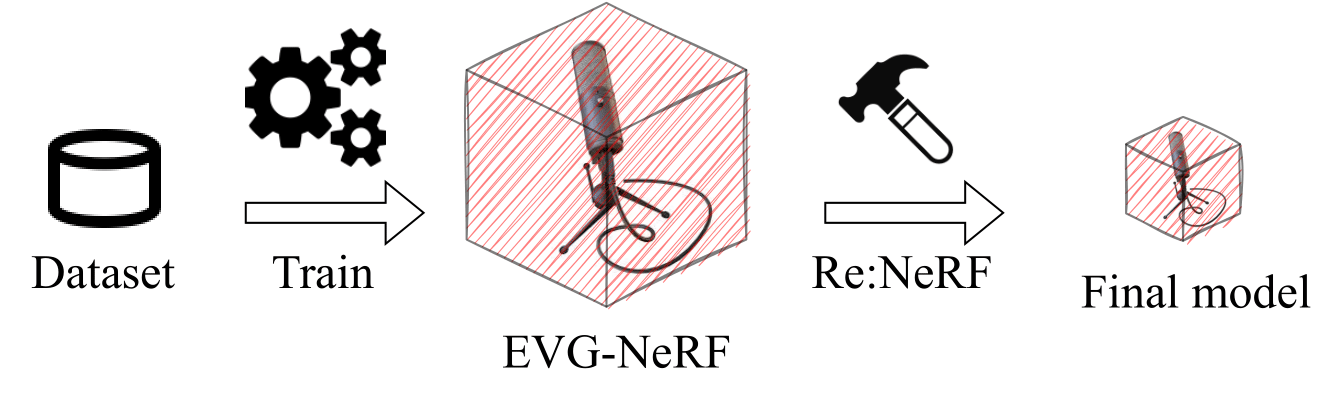}
    \caption{NeRF models with explicit voxel grid representations can be effectively compressed with \method.}
    \label{fig:teaser}
\end{figure}

In an effort to speed up the vanilla NeRF, follow-up works introduced modifications to the original NeRF architecture~\cite{fridovich2022plenoxels, chen2022tensorf, sun2022direct}. One of the popular approaches, for example, encodes features of a scene in an explicit 3D voxel grid, combined with a tiny MLP. This group of methods, which utilizes an ``explicit voxel grid'' (EVG), is gaining more and more popularity due to the high training and rendering speed while maintaining or improving the performance of the original NeRF. Unlike traditional NeRF, EVG-NeRF models require larger memory, limiting their deployment in real-life applications, where models need to be shared through communication channels, or many of these models must be stored on memory-constrained devices.

In this work, we propose \method, a method that reduces memory storage required by trained EVG-NeRF models. Its goal is to accurately separate the object from its background, discarding unnecessary features for rendering the scene, guided by the loss functions designed for training the specific EVG-NeRFs. \method~enables generation of highly compressed models with little or no performance loss: it is specifically designed for EVG-NeRFs as it exploits a spatial locality principle for adding-back voxels to the grid, and in such a sense its working flow resembles the one of a sculptor (Fig.~\ref{fig:teaser}). We observe that \method~enables high-level compression of pre-trained EVG-NeRF models, and that traditional general-purpose approaches, such as blind pruning, perform worse than \method. We test \method~on four datasets with three recent EVG-NeRFs validating the effectiveness of the proposed approach.

\section{Related works}
\label{sec:sota}
\begin{figure}[t!]
\centering
\begin{subfigure}{1.0\columnwidth}
    \includegraphics[width=\textwidth]{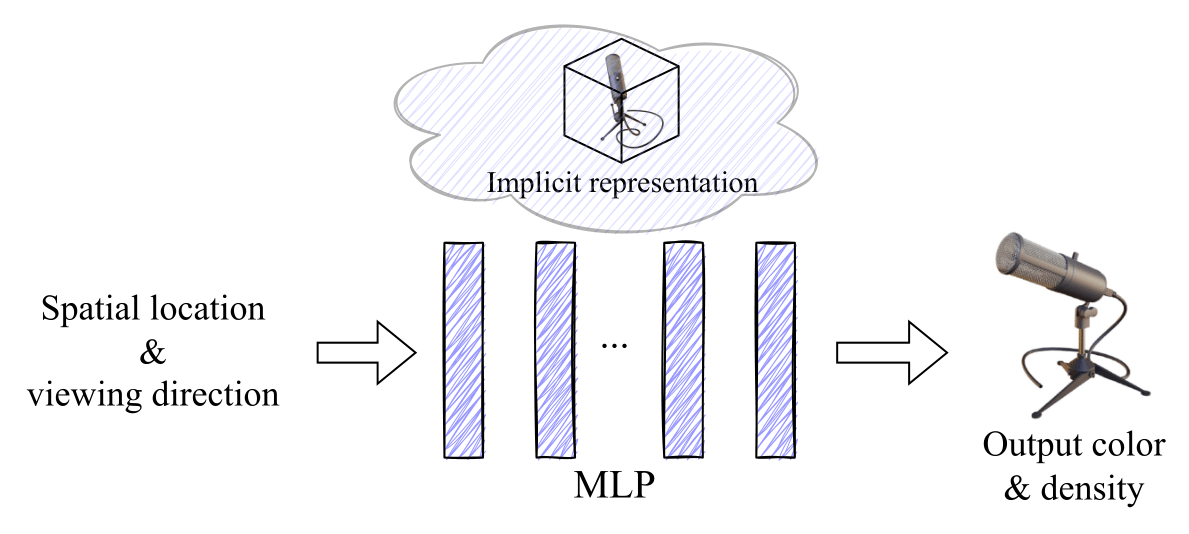}
    \caption{~}
    \label{fig:tradNerf}
\end{subfigure}
\begin{subfigure}{1.0\columnwidth}
    \includegraphics[width=\textwidth]{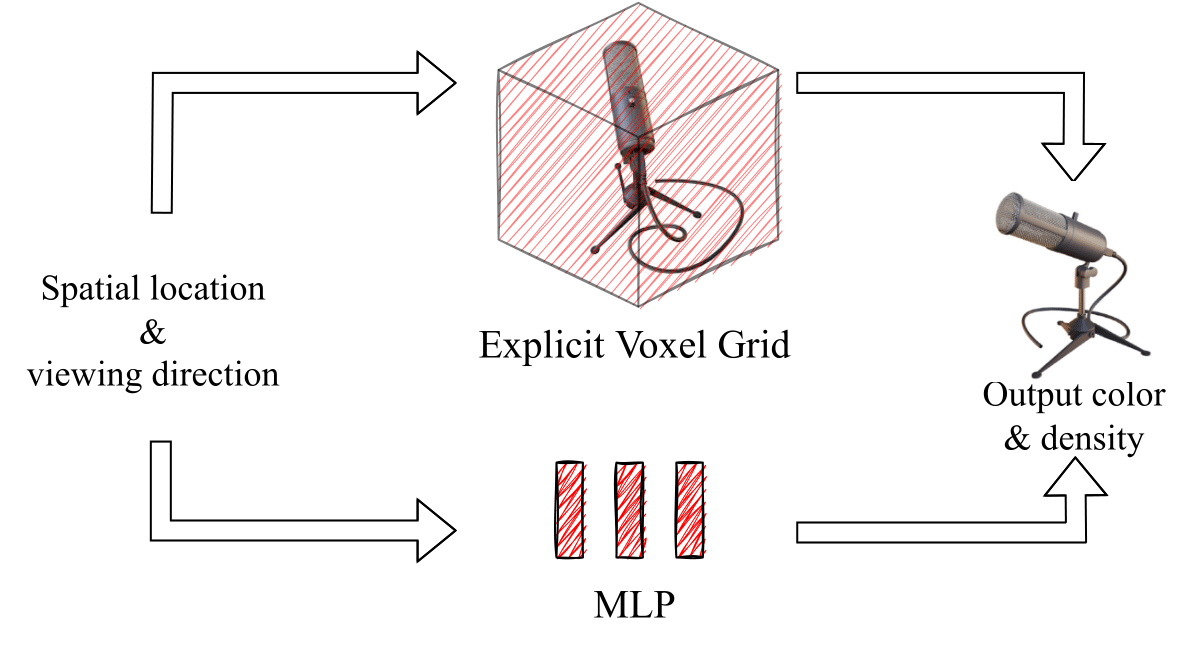}
    \caption{~}
    \label{fig:evgNerf}
\end{subfigure}
\caption{Visualisation of traditional NeRF approach, consisting of a multi-layer perceptron (a) and NeRF-based approaches with explicit voxel grid representation (b). The latter \emph{can} also have a small MLP.}
\label{fig:sota}
\end{figure}

Rendering photo-realistic novel views of a 3D scene from a set of calibrated 2D images of the given scene has been a popular area of research in computer vision and computer graphics. Inspired by Mildenhall~\emph{et~al}'s work in 2020, which proposed to capture the radiance and density field of a 3D scene entirely using a multi-layer perceptron (MLP)~\cite{mildenhall2020nerf}, a large number of follow-up studies have adopted the implicit representation of a scene. Here follows an overview of 3D representation models, neural radiance fields, and follow-up works.\\
\textbf{3D representation for novel view synthesis.} 
Inferring novel views of a scene given a set of images is a long-standing challenge in the field of computer graphics. Various scene representation techniques for 3D reconstruction have been studied in past decades. Light field rendering~\cite{davis2012unstructured, levin2010linear, shi2014light} directly synthesizes unobserved viewpoints by interpolating between sampled rays but it is slow to render and requires substantial computational resources. Meshes are another common technique that is easy to implement and allows rendering in real-time~\cite{debevec1996modeling, thies2019deferred, waechter2014let}. However, it struggles to capture fine geometry and topological information and its rendering quality is limited to mesh resolution. Differentiable methods have been recently proposed to perform scene reconstruction~\cite{flynn2019deepview, li2020crowdsampling, srinivasan2019pushing}. They use a differentiable ray-marching operation to encode and decode a latent representation of a scene and achieve excellent rendering quality.\\
\textbf{Neural Radiance Fields.} 
Unlike traditional explicit volumetric representation techniques, NeRF~\cite{mildenhall2020nerf} stands out in recent years to be the most prevalent method for novel view rendering that infers photo-realistic views given a moderate number of input images. It encodes the entire content of the scene including view-dependent color emission and density into a single multi-layer perceptron (Fig.~\ref{fig:tradNerf}) and achieves state-of-the-art quality. Besides, Neural Radiance Field-based approaches are proving on-the-field to have good generalization when undergoing several transformations, like changing environmental light~\cite{boss2021nerd, srinivasan2021nerv}, image deformation~\cite{gafni2021dynamic, noguchi2021neural, tretschk2021non} and are even usable in more challenging setups including meta learning~\cite{tancik2021learned}, learn dynamically-changing scenes~\cite{gao2021dynamic, li2021neural, martin2021nerf, xian2021space} and even in generative contexts~\cite{chan2021pi, kosiorek2021nerf, schwarz2020graf}. Compared to explicit representations, NeRF requires very little storage space, but on the contrary suffers from lengthy training time and very slow rendering speed, as the MLP is queried an extremely high number of times for rendering a single image.\\
\textbf{NeRF with explicit voxel grids.} 
To reduce inference and training time, explicit prior on the 3D object representation can be imposed. The most intuitive yet effective approach relies on splitting the 3D volume into small blocks, each of which is learned by a tiny NeRF model. With KiloNeRF~\cite{reiser2021kilonerf}, the advantage of doing this is twofold: the size of a single NeRF model is much smaller than the original one, reducing the latency time; secondly, the rendering process itself becomes parallelizable, as multiple pixels can be rendered simultaneously. The downside of this approach is that the granularity of the KiloNeRFs needs to be properly tuned, and the distillation of the single tinier NeRFs can be quite an expensive process. An interesting approach that leverages radiance fields with no explicit neural component is Plenoxels~\cite{fridovich2022plenoxels}. In this case, a sparse feature grid is encoded with 3D spherical harmonics (it belongs to EVG approaches without the MLP component in Fig.~\ref{fig:evgNerf}). Hence, both the training time and the inference times are drastically improved, however, at the cost of a significant increment in-memory storage for the learned model, despite its sparse representation. Showing similar convergence time but maintaining an MLP component for complex view-dependent appearances, DVGO~\cite{sun2022direct} proposes post-activation interpolation. Recently, in order to further improve the execution speed, TensoRF~\cite{chen2022tensorf} has been proposed, which decomposes a 4D tensor into low-rank components prior to training. With the lower-quality rendering setup, the authors deliver a model of size comparable to the original NeRF, but with higher-quality rendering the memory discrepancy with the vanilla NeRF model is still quite wide.\\
\textbf{Compressing EVG-NeRF.} 
Whilst dense voxel-based representations increase rendering speed drastically, they require an order of magnitude more memory than implicit volumetric representations to achieve comparable rendering quality. Hierarchical structure representations using octrees allow the 3D scene to be encoded in a sparse manner, but the memory occupancy still remains high. Recent work addressed the problem of training a model with neural sparse voxel fields~\cite{liu2020neural} progressively reducing the granularity of voxels and skipping the rendering for empty voxels. This approach, however, is designed for resource reallocation. While it improves the rendering speed, it still suffers from a long training time. To the best of our knowledge, \method~is the first approach focusing on compression specifically for EVG-NeRFs. While other works leverage the knowledge of sparsity of the 3D scene \cite{liu2020neural, fridovich2022plenoxels, sun2022direct}, they are focused on performance enhancement (fighting against artifacts which might appear in the empty space) and are not specific for compression. In this work, we are NeRF architecture agnostic, and our goal is to preserve the performance while reducing the model's size.

\section{\method}
\label{sec:method}
In this section, we present \method, our approach towards storage memory reduction for EVG-NeRFs. To reduce the model size, we iteratively remove parameters with the least ranked importance. Following each round of pruning, we design a strategy that adds back neighbor voxels to avoid a drop in performance. 

\subsection{Which parameters are important?}
\label{sec:importance}
One of the key characteristics making EVG-NeRF an effective approach is the possibility of end-to-end training: given some target loss function $\mathcal{L}$ evaluated on the rendered image, using back-propagation, it is possible to train all the parameters$\boldsymbol{w}$ of the model. This learning approach is common with any standard deep neural network, which allows us to build on top of the existing technique with the same set of optimizers (such as SGD and Adam). Methods based on mini-batches of samples have gained popularity, as they allow better generalization than stochastic learning while being memory and time efficient. They also benefit from libraries that exploit parallel computation on GPUs. In such a framework, a network parameter $w_i$is updated towards the averaged direction which minimizes the averaged loss for the mini-batch. Evidently, if the gradient's magnitude is zero, the parameter is not updated, meaning that the local loss landscape for it is \emph{flat}. A typical approach to reduce the number of parameters in a deep neural network is to \emph{threshold} the parameters according to some hyper-parameters that determine the amount to be removed~\cite{tartaglione2022loss, Frankle2019TheLT}:
\begin{equation}
    \label{eq:magprune}
    w_i=\left\{
    \begin{array}{ll}
        w_i & if |w_i| > \mathcal{Q}_{|\boldsymbol{w}|}(\gamma)\\
        0 & otherwise,
    \end{array}
    \right .
\end{equation}
where $\mathcal{Q}_{|\boldsymbol{w}|}(\cdot)$ is the quantile function for the $\ell_1$ norm of the parameters and $\gamma \in [0; 1]$ is the percentage of parameters to be removed. Despite its simplicity and broad application, this approach has a potential issue: parameters having very low magnitude can be important for the model. For example, a parameter can have a very low magnitude but a high gradient: hard-setting it to zero according to \eqref{eq:magprune} can significantly influence the loss value/performance. Because of this, other works have suggested evaluating the importance of a parameter using the gradient of a parameter as a criterion~\cite{lecun1989optimal, tartaglione2021serene}. A parameter $w_i$ can have a low gradient locally, but removing it may potentially impose a drastic change in both the loss value and its gradient. It is necessary, hence, to find a compromise between these two conditions. We can estimate the variation of the loss value using a Taylor series expansion truncated to the first order:
\begin{equation}
    \label{eq:taylor}
    \Delta \mathcal{L}(w_i) \approx \frac{\partial \mathcal{L}}{\partial w_i} w_i,
\end{equation}
and from \eqref{eq:taylor} we can define how to remove the parameters according to
\begin{equation}
    \label{eq:taylorprune}
    w_i=\left\{
    \begin{array}{ll}
        w_i & if |\mathcal{L}(w_i)| > \mathcal{Q}_{|\Delta \mathcal{L}(w)|}(\gamma)\\
        0 & otherwise .
    \end{array}
    \right .
\end{equation}
It is a known fact, however, that both gradient and weight magnitudes for the parameters change depending on the typology of layers taken into consideration~\cite{lee2018snip}. Hence, in order to address a parameter-removing strategy that could be applied globally (hence, removing a given ratio of the parameters from the whole model, without imposing uniformity in this removal), the quantile function should be evaluated on the layer-normalized quantity
\begin{equation}
    \label{eq:taylornorm}
    \Delta \hat{\mathcal{L}}(w_i) = \frac{\frac{\partial \mathcal{L}}{\partial w_i} w_i}{\max\left| \frac{\partial \mathcal{L}}{\partial w_j} w_j \right|}, w_j\text{ in same layer as }w_i.
\end{equation}
Consequently, \eqref{eq:taylorprune} becomes
\begin{equation}
    \label{eq:taylorprunenormalized}
    w_i=\left\{
    \begin{array}{ll}
        w_i & if |\mathcal{L}(w_i)| > \mathcal{Q}_{|\Delta \hat{\mathcal{L}}(w)|}(\gamma)\\
        0 & otherwise .
    \end{array}
    \right .
\end{equation}
This strategy, however, evaluates the loss variation for each parameter $w_i$ independently as in \eqref{eq:taylor}, which is known to be sub-optimal, as there is a dependence between parameters inside the model. How can we correct a potential ``excessive'' removal of parameters?

\subsection{Removing only?}
Removing parameters from a model is always a matter of delicacy: if the parameters are removed too fast, at some point the performance can not be recovered. On the contrary, if they are removed too slowly, the training complexity becomes large. Furthermore, the strategy to identify which parameters can be removed from the model, for a matter of efficiency, is limited to a first-order approximation in~\eqref{eq:taylor}, making the parameter removal mechanism potentially prone to approximation errors. How can we identify the parameters, which have been removed, and should be added back in order not to degrade the performance excessively?\\
Let us consider the subset of parameters $\overline{\mathcal{W}}$ which have been removed. Since these parameters have been removed, according to \eqref{eq:taylor}, $\Delta \mathcal{L}(w_i) = 0 \forall w_i \in \overline{\mathcal{W}}$, meaning that this metric cannot be used to eventually re-include parameters in the model.

In order to determine whether the re-inclusion of a previously removed parameter will enhance the performance further (or in other words, will cause the minimization of the evaluated loss function) we can, for instance, look at the value for its gradient. If the gradient is above a given threshold, the parameter is added back. A simple threshold could be defined by the distribution of the magnitude of the gradients for the remaining parameters $\mathcal{W}$:
\begin{equation}
    \label{eq:reinc}
    \left|\frac{\partial \mathcal{L}}{\partial w_i}\right| \geq \mathcal{Q}_{\left|\frac{\partial \mathcal{L}}{\partial w}\right|, w\in \mathcal{W}}(\delta) \Rightarrow w_i \in \mathcal{W} ,
\end{equation}
where $\delta\in [0;1]$ determines the relative threshold for the re-inclusion.

\begin{figure}[t]
\centering
\begin{subfigure}{.45\columnwidth}
    \includegraphics[width=\textwidth]{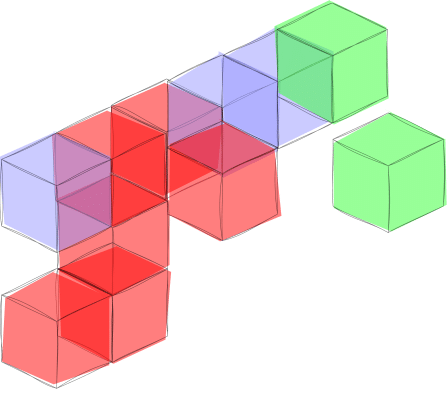}
    \caption{~}
    \label{fig:beforeRE}
\end{subfigure}
\hfill
\begin{subfigure}{.45\columnwidth}
    \includegraphics[width=\textwidth]{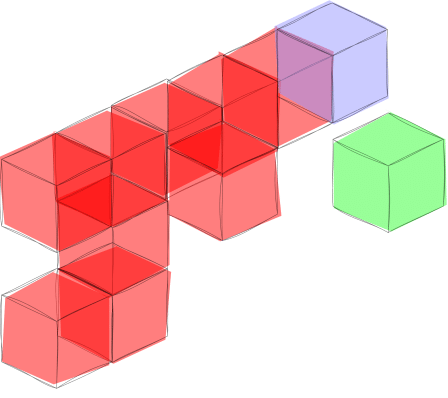}
    \caption{~}
    \label{fig:afterRE}
\end{subfigure}
\caption{Effect of RE-INCLUDE before (a) and after running one iteration (b). In red: voxels already in the model; in green: non-neighbor voxels satisfying the re-inclusion rule; in blue: neighbor voxels satisfying the re-inclusion rule. }
\label{fig:neighb}
\end{figure} 

Although \eqref{eq:reinc} is a general rule and is potentially applicable to all the layers for EVG-NeRFs, we can leverage the voxel grid structure, imposing a prior over the 3D manifold representation for the object itself. We expect it to be \emph{compact} and \emph{the least sparse possible}. Towards this end, we add, as an additional constraint to \eqref{eq:reinc}, that a parameter $w_i \in \overline{\mathcal{W}}$, in order to be re-included, it should also be connected, or should be a \emph{neighbor} of some $w_j \in \mathcal{W}$. Hence, the re-inclusion rule becomes
\begin{equation}
    \label{eq:reinc-final}
    \begin{array}{cl}
        \left|\frac{\partial \mathcal{L}}{\partial w_i}\right| \geq \mathcal{Q}_{\left|\frac{\partial \mathcal{L}}{\partial w}\right|, w\in \mathcal{W}}(\delta) \\
        \wedge & \Rightarrow w_i \in \mathcal{W}\\
        \exists w_j \in \mathcal{W} | w_j \in \Omega(w_i),
    \end{array}
\end{equation}
where $\Omega(w_i)$ is the subset of parameters that are neighbors of $w_i$. Fig.~\ref{fig:beforeRE} displays a practical case where there are some voxels not included (white space), voxels in the model (red), voxels removed which satisfy \eqref{eq:reinc-final} (blue) and voxels which satisfy the condition on the gradient, but are not neighbors of any voxel in the model (green). After one re-inclusion iteration, the blue voxels are included, and some green voxels (the neighbors of the blue ones) become the new candidates for the re-inclusion (Fig.~\ref{fig:afterRE}). In order to find the whole subset of voxels to be added-back, it is necessary to iterate over the re-inclusion mechanism, until there are no voxels in blue to add. Follows an overview on \method.

\subsection{Overview on the \method~scheme}
\begin{algorithm}[ht]
\caption{\method.}
\label{alg:ourmethod}
\begin{algorithmic}[1]
\Procedure{\method ($\mathcal{W}_{beg}$, $\gamma$, $\delta$)}{}
    \State $T_{rem} \gets \mathcal{Q}_{|\Delta\hat{\mathcal{L}}(w)|, w\in \mathcal{W}_{beg}}(\gamma)$
    \State $\mathcal{W}, \overline{\mathcal{W}} \gets \text{REMOVE}(\mathcal{W}_{beg}, T_{rem})$\label{line:rem}
    \State $T_{inc} \gets \mathcal{Q}_{\left|\frac{\partial \mathcal{L}}{\partial w}\right|, w\in \mathcal{W}}(\delta)$
    \State $\mathcal{W}_{end} \gets \text{RE-INCLUDE}(\mathcal{W}, \overline{\mathcal{W}}, T_{inc})$\label{line:add}
    \State \Return $\mathcal{W}_{end}$
\EndProcedure

\Procedure{REMOVE($\mathcal{W}_{beg}$, $T_{rem}$)}{}
    \State $\mathcal{W} \gets \emptyset$
    \State $\overline{\mathcal{W}}\gets \emptyset$
    \For{$w_i \in \mathcal{W}_{beg}$}
        \If{$\left|\Delta\mathcal{L}(w_i) \right|\geq T_{rem}$}
            \State $\mathcal{W} \gets \mathcal{W} \cup \{w_i\}$
        \Else
            \State $\overline{\mathcal{W}} \gets \overline{\mathcal{W}} \cup \{w_i\}$
        \EndIf
    \EndFor
\State \Return $\mathcal{W}, \overline{\mathcal{W}}$
\EndProcedure

\Procedure{RE-INCLUDE($\mathcal{W}$, $\overline{\mathcal{W}}$, $T_{inc}$)}{}
    \State $one\_added \gets True$
    \While{$one\_added$}\label{line:oneadd}
        \State $one\_added \gets False$
        \For{$w_i \in \overline{\mathcal{W}}$}
            \If{$\left|\frac{\partial \mathcal{L}}{\partial w_i}\right|\geq T_{inc}$}
                \State $\Omega \gets \text{NEIGHBORS}(w_i)$
                \If{$\Omega \cap \mathcal{W} \neq \emptyset$}\label{line:neightest}
                    \State $\mathcal{W} \gets \mathcal{W} \cup \{w_i\}$
                    \State $one\_added \gets True$
                \EndIf
            \EndIf
        \EndFor
    \EndWhile
    \State \Return $\mathcal{W}$
\EndProcedure

\end{algorithmic}
\end{algorithm}
\begin{figure}
    \includegraphics[width=\columnwidth]{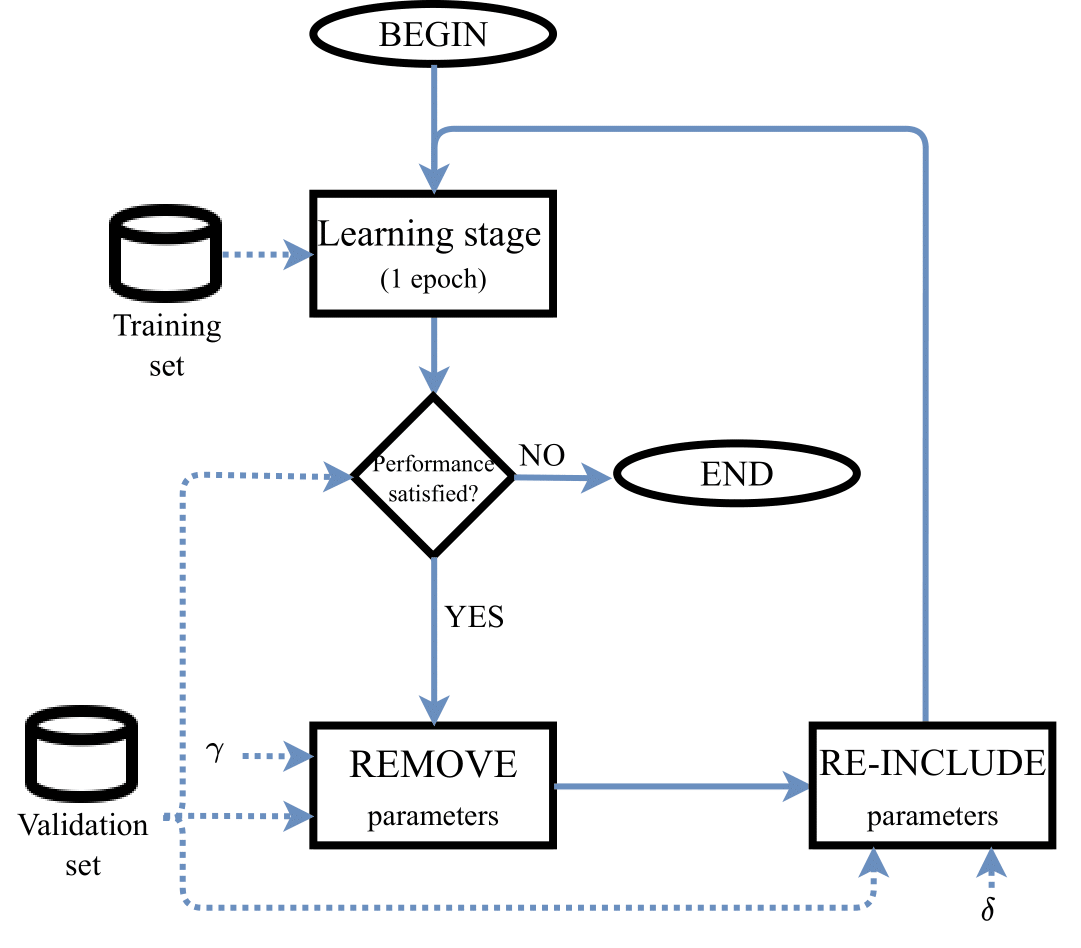}
    \caption{Overview on \method. The dashed arrows indicate usage of some specific dataset/hyper-parameter at every stage.}
    \label{fig:overview}
\end{figure}

In this section, we provide an overview of \method, which is displayed in Fig.~\ref{fig:overview}. Given a pre-trained model, we perform a one-epoch fine-tuning on the model with the same policy as in the original NeRF model, moving then to the parameters removal/re-inclusion to determine the subset $\mathcal{W}$ of parameters belonging to the model.  Every time we perform a step of parameter removal, we follow the steps as in Algorithm~\ref{alg:ourmethod}. In particular, we are asked a subset of parameters to belong to the model $\mathcal{W}_{beg}$ and two hyper-parameters $\gamma\in [0;1]$ and $\delta\in [0;1]$: while $\gamma$ determines how many parameters are (tentatively) removed after every step, $\delta$ determines how many parameters are (eventually) added back. Hence, we distinguish two phases for \method: one (REMOVE) splits the model parameters into those dropping below or staying above a given threshold (line~\ref{line:rem}), and the other (RE-INCLUDE) re-includes the tentatively removed parameters that have both high derivative and are neighbors of other
parameters in the model. This will favor lower loss (line~\ref{line:add}). In particular, the latter might need to be run multiple times, every time at least one parameter is re-included (line~\ref{line:oneadd}). This is necessary as, every time a new parameter is added to $\mathcal{W}$, the neighbor test as in line~\ref{line:neightest} potentially gives a different outcome.\\
We iterate over this until the performance does not drop below some pre-fixed performance threshold $\Delta T$ (from the original performance): when this happens, we end our training process. In order to save the model, the state dictionary is first quantized on 8-bits with a uniform quantizer and successively compressed using LZMA.\\
In the next section, we are going to present the results obtained on some common benchmarks for NeRFs.
\section{Results}
\label{sec:results}

In this section, we present the empirical results obtained on state-of-the-art datasets and three different EVG-NeRF approaches, on top of which \method~has been executed in order to reduce the storage memory. For all the experiments the models have been pre-trained using the hyper-parameter setup indicated in the respective original work. As a common stop criterion, we impose a maximum worsening in performance $\Delta T$ of $1$dB on the original model's PSNR. All the other hyper-parameters have been optimized using a grid-search algorithm. Although every technique requires a specific CUDA and PyTorch version, the \method~code is compatible with pytorch~1.12 and back-compatible with PyTorch~1.6. For all the experiments an NVIDIA A40 equipped with 40~GB has been used.\footnote{The source code will be made available at the conference's dates.}

\begin{table*}
    \caption{Results obtained on low compressibility regime (LOW) and high compressibility (HIGH). The first line indicates the baseline. In every dataset, the various metrics are averaged for the samples in them.}
    \label{tab:results}
	\renewcommand{\arraystretch}{1.2}
    \resizebox{\textwidth}{!}{
	\centering
		\begin{tabular}{@{\hskip1pt}c c c c c c c c c c c c c c @{\hskip1pt}}
        \toprule
        
        \multirow{3}{*}{\bf \large Approach} & \multirow{3}{*}{\bf \large Compress}   &\multicolumn{3}{c}{\bf \large Synthetic-NeRF}&\multicolumn{3}{c}{\bf \large Synthetic-NSVF} &\multicolumn{3}{c}{\bf \large Tanks\&Temples} &\multicolumn{3}{c}{\bf \large LLFF-NeRF}\\
        
        & &\bf PSNR  & \bf SSIM & \bf Size &\bf PSNR  & \bf SSIM & \bf Size&\bf PSNR  & \bf SSIM & \bf Size  &\bf PSNR  & \bf SSIM & \bf Size\\
        
        & & [dB]($\uparrow$)  &($\uparrow$)& [MB]($\downarrow$)& [dB]($\uparrow$)  &($\uparrow$)& [MB]($\downarrow$)& [dB]($\uparrow$)  &($\uparrow$)& [MB]($\downarrow$)& [dB]($\uparrow$)  &($\uparrow$)& [MB]($\downarrow$) \\
        \midrule 
        NSVF~\cite{liu2020neural} & - & 31.74 & 0.953& $\sim$ 16 & 35.13& 0.979 & $\sim$16&28.40 &0.900 & $\sim$16 & - & - & - \\
        Instant-NGP~\cite{mueller2022instant} & - & 33.04 & 0.934&  28.64& 36.11& 0.966 & 46.09&28.81 &0.917 & 46.09& 20.18 & 0.662 & 46.09  \\
        
        \midrule 
          & - &31.92 &0.957&160.09 & 35.42 &0.979&104.12& 28.26 &0.909&106.48  & - & - & -\\
         DVGO~\cite{sun2022direct} & LOW & 31.47& 0.952& 3.99& 35.29&0.974& 4.37& 28.22 &0.910& 4.69& - & - & -\\
          & HIGH & 31.08&0.944 & 2.00& 34.90&0.969& 2.46& 27.90 &0.894& 1.62& - & - & -\\

         \midrule
          & - &33.14 &0.963&69.26 &36.52 & 0.982& 69.05 & 28.56&0.920 &64.04& 26.73 & 0.839&151.79 \\
         TensoRF~\cite{chen2022tensorf} & LOW & 33.26&0.962 &11.47 & 36.44 &0.982&11.60 & 28.50 &  0.916& 9.99& 26.80 & 0.820 & 32.34\\
          & HIGH & 32.81&0.956&7.94& 36.14&0.978&8.52& 28.24 & 0.907 &6.70 & 26.55 & 0.797 & 20.27\\
         \midrule
         & - & 31.48 & 0.956 & 189.08 & - & - & - & 27.37 & 0.904 & 147.96
         & 25.90 & 0.838 & 1484.96\\
         Plenoxels~\cite{fridovich2022plenoxels} 
         & LOW & 31.52 & 0.952 & 91.77 & - & - & - & 27.66 & 0.909 & 102.26 & 26.24 & 0.838 & 457.23\\
         & HIGH & 30.97 & 0.944 & 54.68 & - & - & - & 27.34 & 0.896 & 85.47& 25.95 & 0.828 & 338.02\\
         
        \bottomrule
	\end{tabular}
    }
\end{table*}
\begin{figure*}
    \begin{subfigure}{0.24\textwidth}
        \begin{subfigure}{0.52\textwidth}
        \includegraphics[width=\textwidth, trim={50 100 150 100},clip]{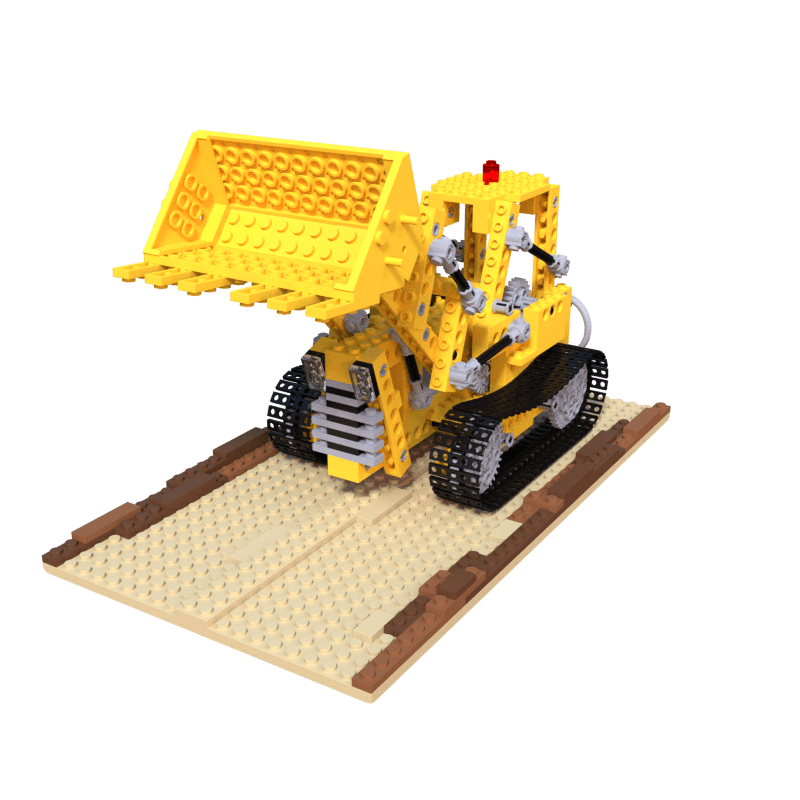}
        \includegraphics[width=1.01\textwidth]{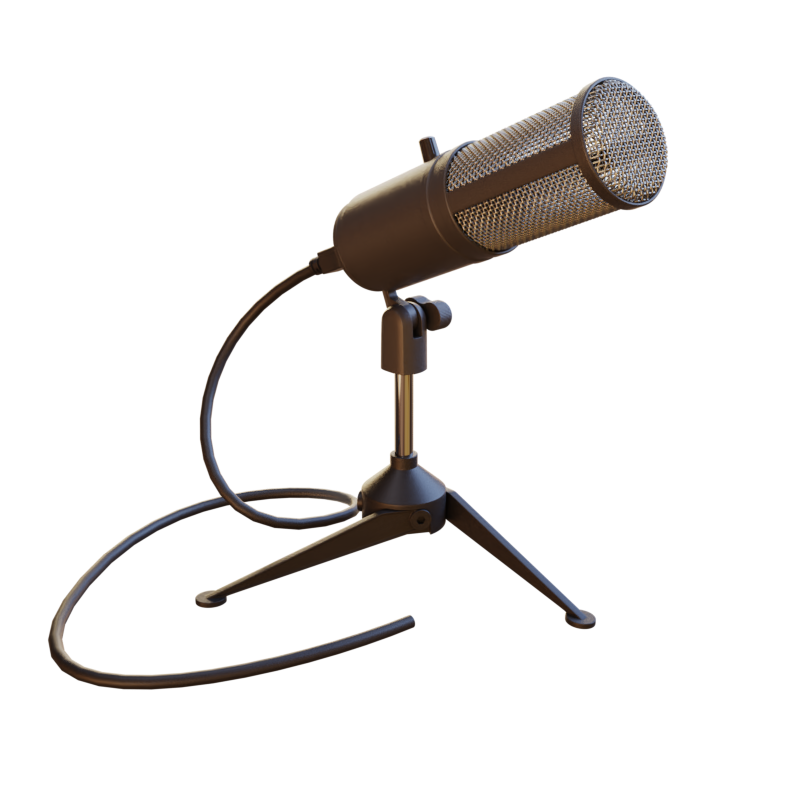}
       
        \end{subfigure}
        \begin{subfigure}{0.45\textwidth}
            \includegraphics[width=\textwidth]{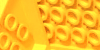}
            \includegraphics[width=\textwidth]{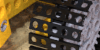}
            \includegraphics[width=\textwidth]{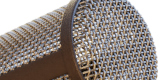}
            \includegraphics[width=\textwidth]{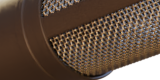}
        \end{subfigure}
        \begin{subfigure}{\textwidth}
             \includegraphics[width=\textwidth]{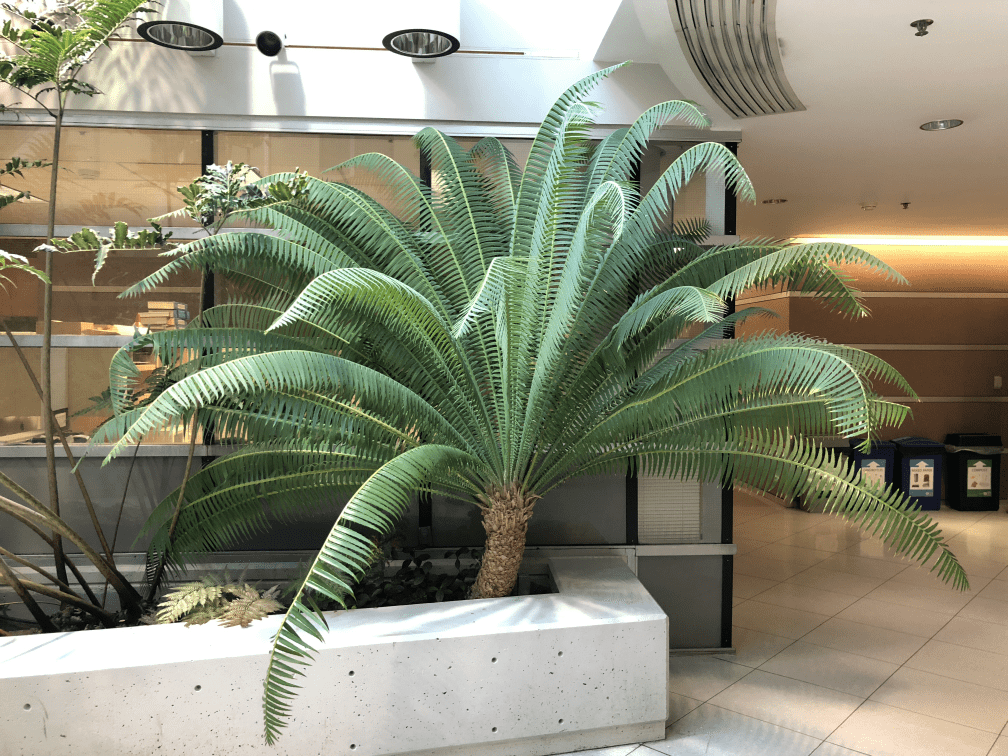}
        \end{subfigure}
        \caption{Ground truth.}
    \end{subfigure}
    \begin{subfigure}{0.24\textwidth}
        \begin{subfigure}{0.52\textwidth}
        \includegraphics[width=\textwidth, trim={50 100 150 100},clip]{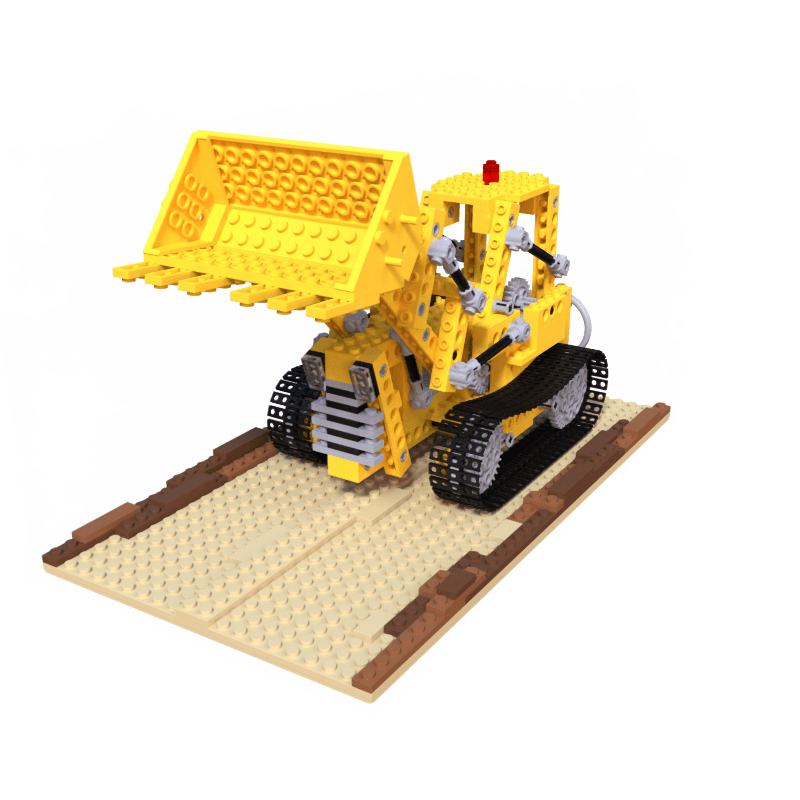}
        \includegraphics[width=1.01\textwidth]{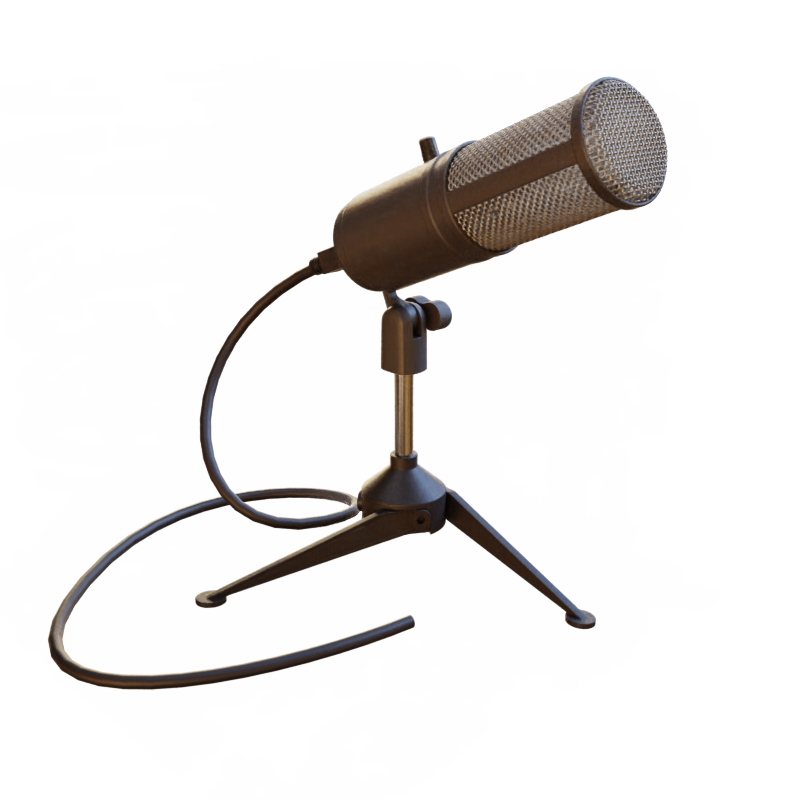}
        \end{subfigure}
        \begin{subfigure}{0.45\textwidth}
            \includegraphics[width=\textwidth]{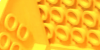}
            \includegraphics[width=\textwidth]{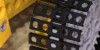}
            \includegraphics[width=\textwidth]{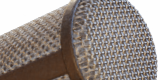}
            \includegraphics[width=\textwidth]{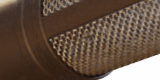}
        \end{subfigure}
              \begin{subfigure}{\textwidth}
             \includegraphics[width=\textwidth]{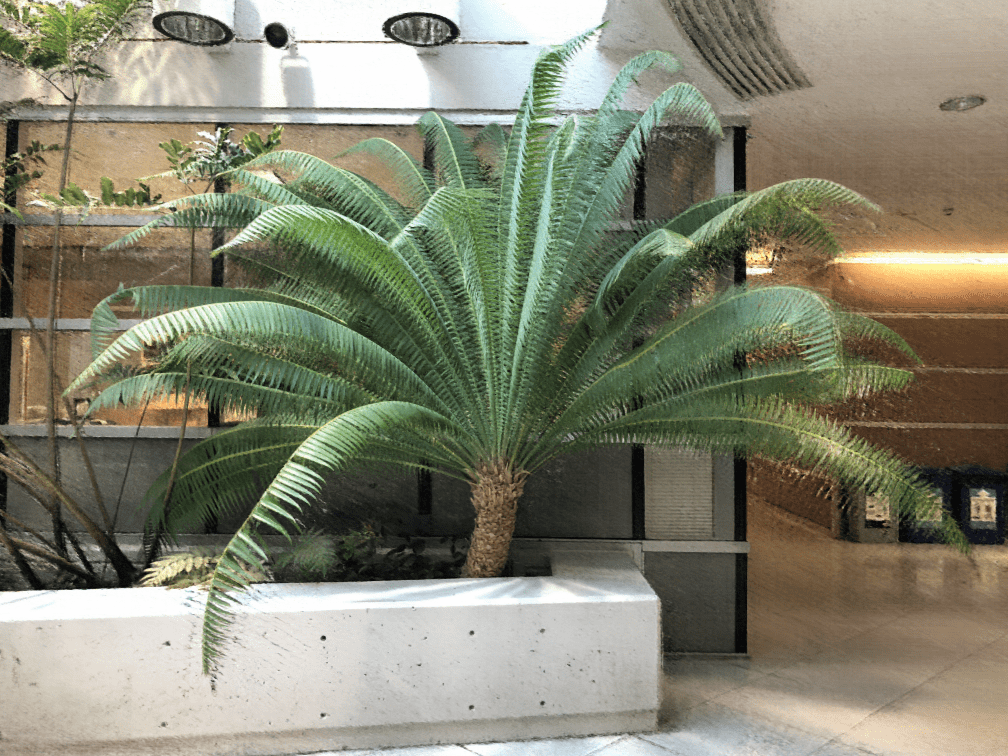}
        \end{subfigure}
        \caption{TensoRF (baseline).}
    \end{subfigure}
    \begin{subfigure}{0.24\textwidth}
        \begin{subfigure}{0.52\textwidth}
        \includegraphics[width=\textwidth, trim={50 100 150 100},clip]{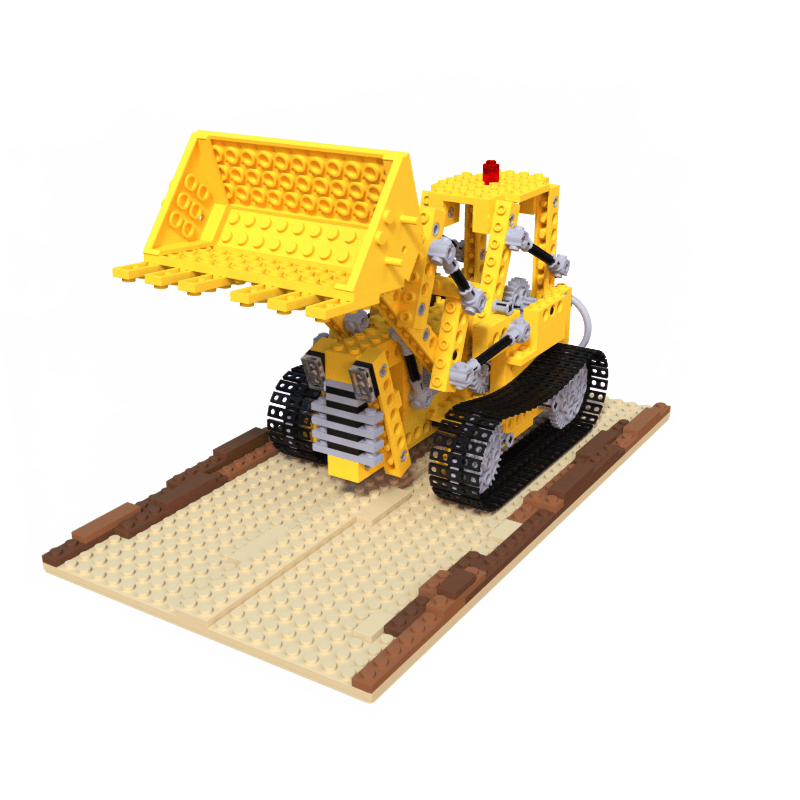}
        \includegraphics[width=1.01\textwidth]{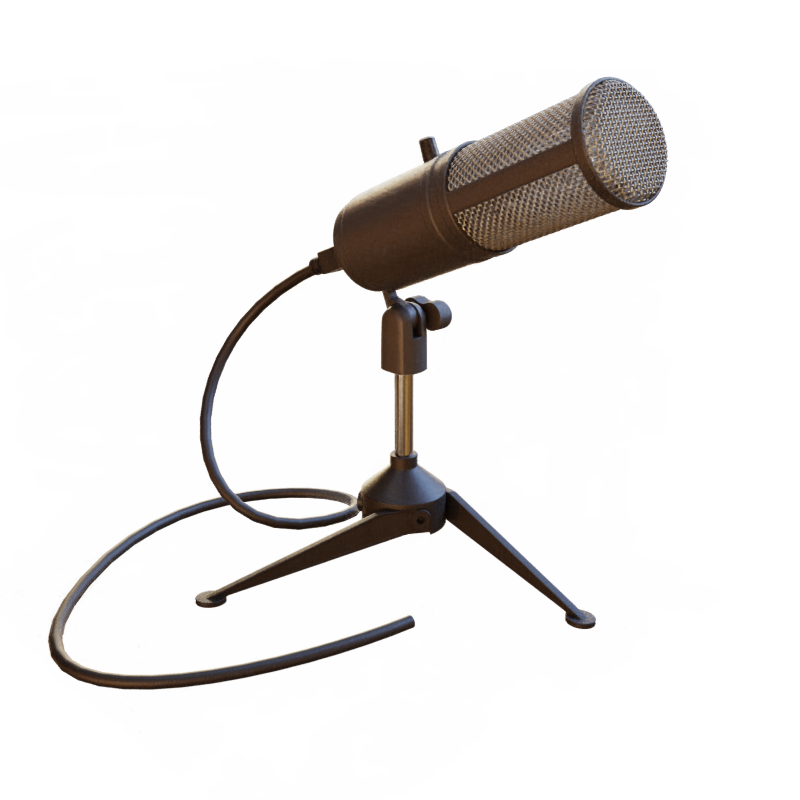}
        \end{subfigure}
        \begin{subfigure}{0.45\textwidth}
            \includegraphics[width=\textwidth]{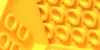}
            \includegraphics[width=\textwidth]{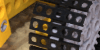}
            \includegraphics[width=\textwidth]{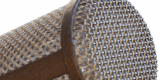}
            \includegraphics[width=\textwidth]{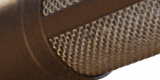}
        \end{subfigure}
                     \begin{subfigure}{\textwidth}
             \includegraphics[width=\textwidth]{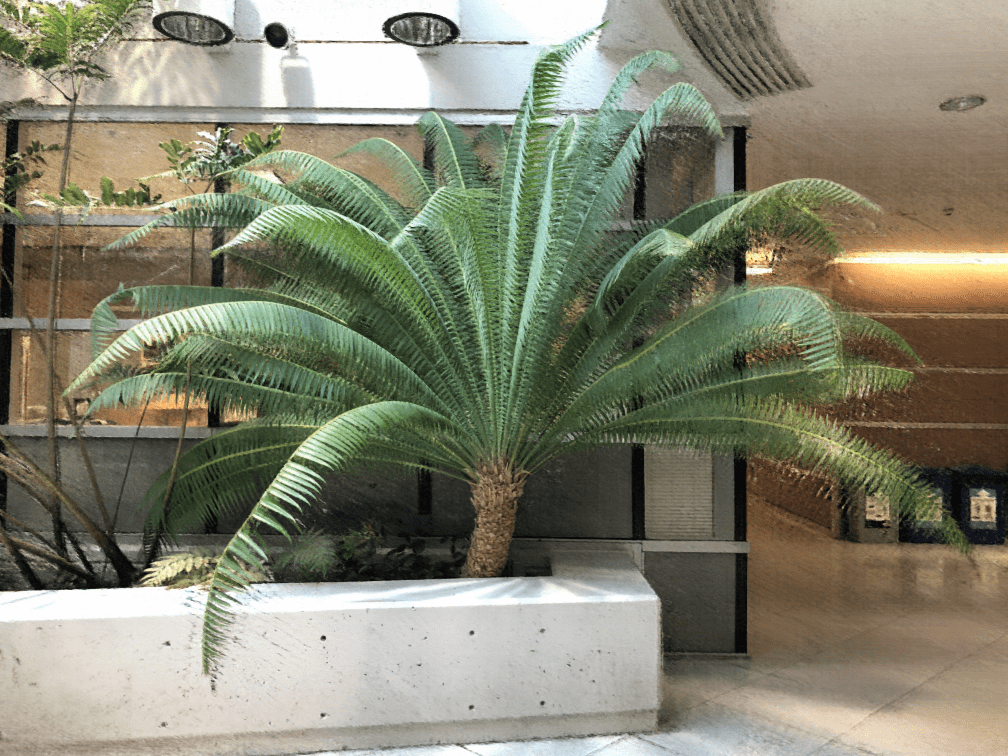}
        \end{subfigure}
        \caption{TensoRF low compress.}
    \end{subfigure}
    \begin{subfigure}{0.24\textwidth}
        \begin{subfigure}{0.52\textwidth}
        \includegraphics[width=\textwidth, trim={50 100 150 100},clip]{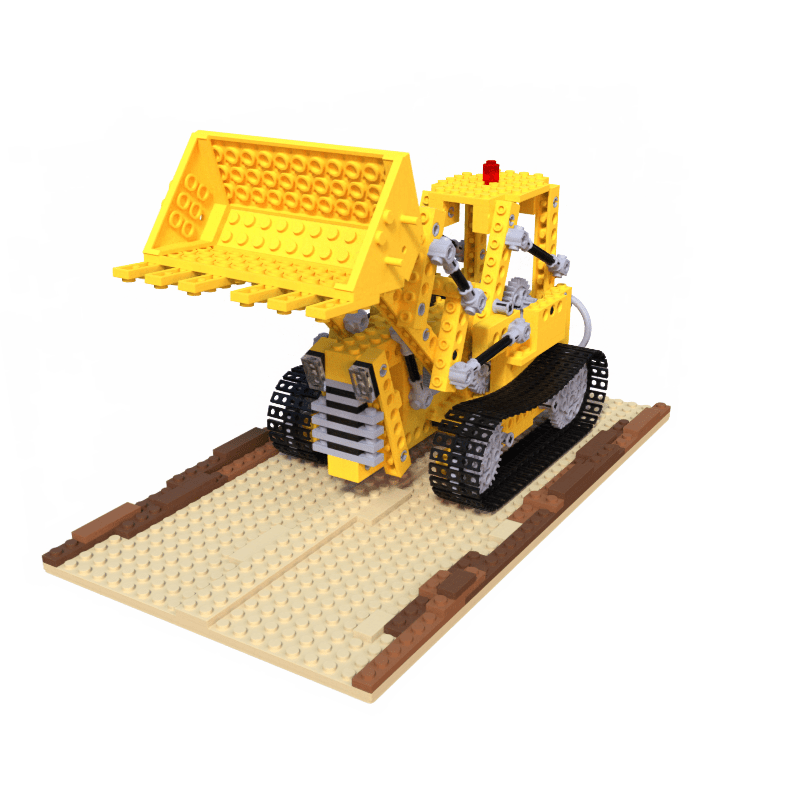}
        \includegraphics[width=1.01\textwidth]{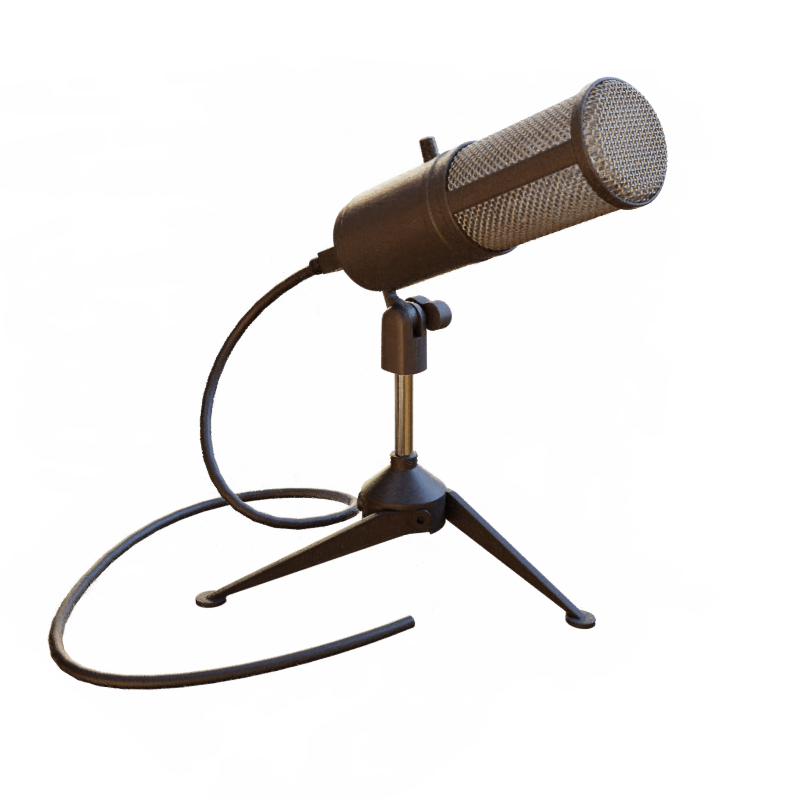}
        \end{subfigure}
        \begin{subfigure}{0.45\textwidth}
            \includegraphics[width=\textwidth]{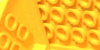}
            \includegraphics[width=\textwidth]{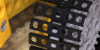}
               \includegraphics[width=\textwidth]{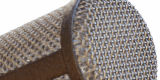}
            \includegraphics[width=\textwidth]{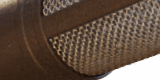}
                \end{subfigure}
                     \begin{subfigure}{\textwidth}
             \includegraphics[width=\textwidth]{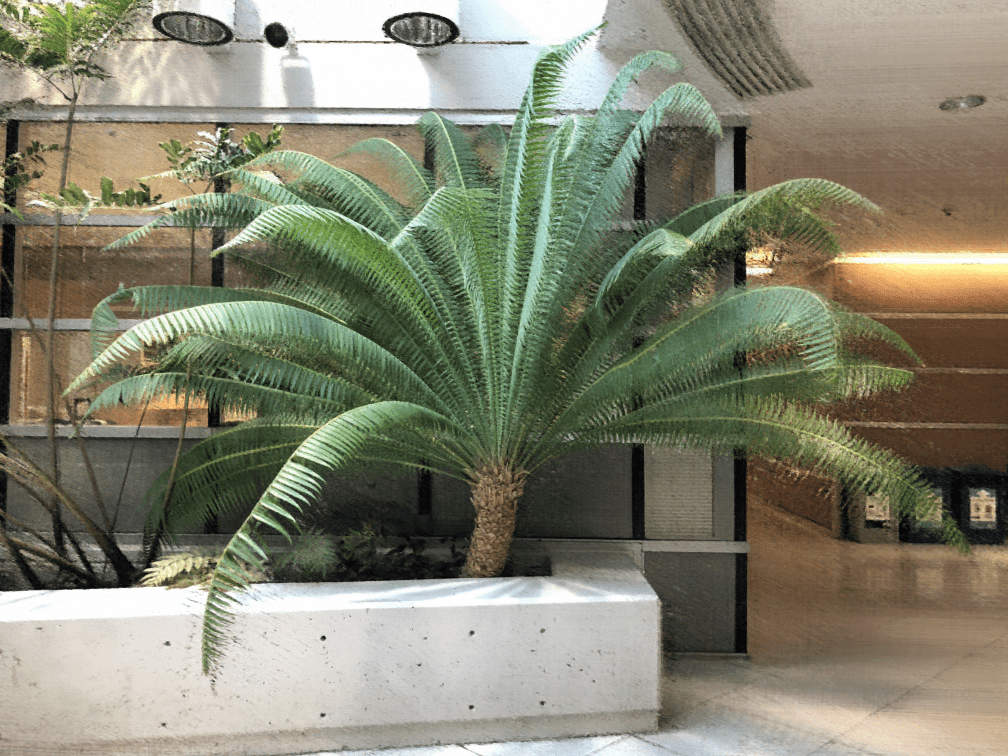}
        \end{subfigure}
        \caption{TensoRF high compress.}
    \end{subfigure}
    \caption{Qualitative results for ``lego'' (top), ``mic'' (middle) and ``fern'' (bottom).}
    \label{fig:highfreq}
\end{figure*}

\subsection{Setup}

\textbf{Datasets.} We have evaluated our approach on four datasets. Synthetic-NeRF~\cite{mildenhall2020nerf} and Synthetic-NSVF~\cite{liu2020neural} are two popular datasets, containing 8 different realistic objects each, which are synthesized from NeRF (\textit{chair, drums, ficus, hotdog, lego, materials, mic} and \textit{ship}) and NSVF (\textit{bike, lifestyle, palace, robot, spaceship, steamtrain, toad} and \textit{wineholder}), respectively. For both, the image resolution has been set up to 800 × 800 pixels, having 100 views for training, 100 for validation, and 100 for testing. 
The third dataset we have tested is Tanks\&Temples~\cite{knapitsch2017tanks}: our choice fell on this dataset as it is a collection of real-world images. Here we use a subset of the provided samples (namely: \textit{ignatius, truck, barn, caterpillar} and \textit{family}). We use here FullHD resolution, and we use also in this case $10\%$ of the images used for validation and $10\%$ for testing. Finally, the fourth dataset we we run our experiments is LLFF-NeRF~\cite{mildenhall2019local}. Differently from the other three datasets, this datasets contains realistic images, and non blank background. Each scene consists of 20 to 60 forward-facing images with resolution 1008 × 756. In this case, we have used all the 8 available samples (\textit{fern, flower, fortress, horns, leaves, orchids, room} and \textit{trex}).\\
\textbf{Architectures and compressibility configuration.} We have tested \method~on three very different EVG-NeRF approaches: DVGO~\cite{sun2022direct}, TensoRF~\cite{chen2022tensorf} and Plenoxels~\cite{fridovich2022plenoxels}.\footnote{Although Plenoxels is a method for learning radiance fields and does not have any ``neural network'', it still leverages the same optimization tools. We include it in our experimental setup to show the even broader adaptability of \method~to any approach minimizing a differentiable loss function.} DVGO models are trained using the same configuration as in the paper, in the $160^3$ voxel grid size configuration. TensoRF models were obtained with their default 192-VM configuration, which factorizes tensors into 192 low-rank components and optimizes the model for 30k steps. Plenoxel models have obtained training first on $128^3$ grid, up-sampled to $256^3$, and finally to $512^3$. For all the architectures and datasets we have used $\gamma=0.5$ and  $\delta=0.5$, except for Plenoxel trained on the Synthetic-NeRF and LLFF-NeRF datasets, where $\gamma=0.66$ has been used. For a matter of comparison with other efficiencing approaches, we compare our results also with NSVF~\cite{liu2020neural} and with Instant-NGP~\cite{mueller2022instant}.

\subsection{Discussion}
All the results are reported in Table~\ref{tab:results}. Here the ``LOW'' compressibility refers to compressibility achieved with the best PSNR evaluated on the validation set, while ``HIGH'' refers to the model achieved right before reaching the stop criterion (which consists of a worsening of the original performance of at most 1dB on the original PSNR). Some qualitative results are also displayed in Fig.~\ref{fig:highfreq}.

In general, we observe that \method~effectively reduces the size of the models in all the combinations of tested datasets/EVG-NeRFs, with different impacts depending on the EVG-NeRF it is applied. In general, the approach having higher average sizes while also having slightly worse performance is Plenoxels~\cite{fridovich2022plenoxels}, which is an EVG approach with no neural elements in it. Nevertheless, \method~ is able to compress it effectively. In particular, in the low compressibility setup, the performance is improved with an overall size reduction. DVGO~\cite{sun2022direct}, consisting of a voxel grid and of an MLP component, is massive, achieving for example compression ratios of $80\times$ for Synthetic-NeRF and $65\times$ within the 1dB performance loss. The approach generally achieving better performance is TensoRF~\cite{chen2022tensorf}, where the low compression setup maintains almost the same performance still enabling $6\times$ compression. When compared to DVGO, TensoRF occupies less memory as it relies on factorized neural radiance fields in the 4D voxel grid, namely it is by design more efficient at training time, and of course in order to maintain such a higher performance the possible compressibility of the model is relatively limited. When compared with other approaches, we observe in general a significant improvement in performance for similar model's size (LLFF-NeRF) or a significantly lower memory for similar performance (in the other three cases). 

\subsection{Ablation study}
\begin{figure*}[t]
\centering
\begin{subfigure}{0.3\textwidth}
    \includegraphics[width=\textwidth, trim={120 30 60 60},clip]{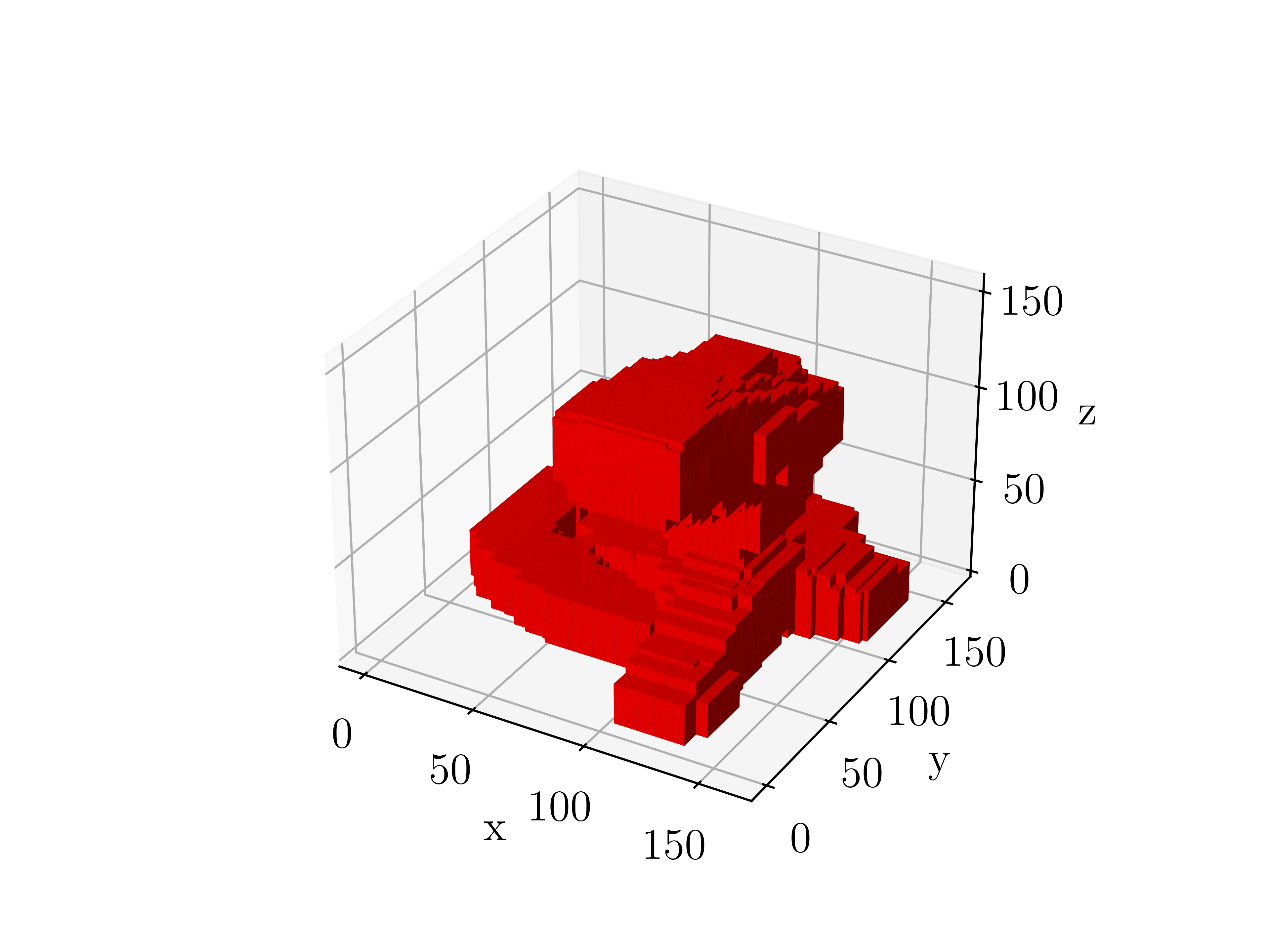}
    \caption{~}
    \label{fig:abloccbase}
\end{subfigure}
\begin{subfigure}{0.3\textwidth}
    \includegraphics[width=\textwidth, trim={120 30 60 60},clip]{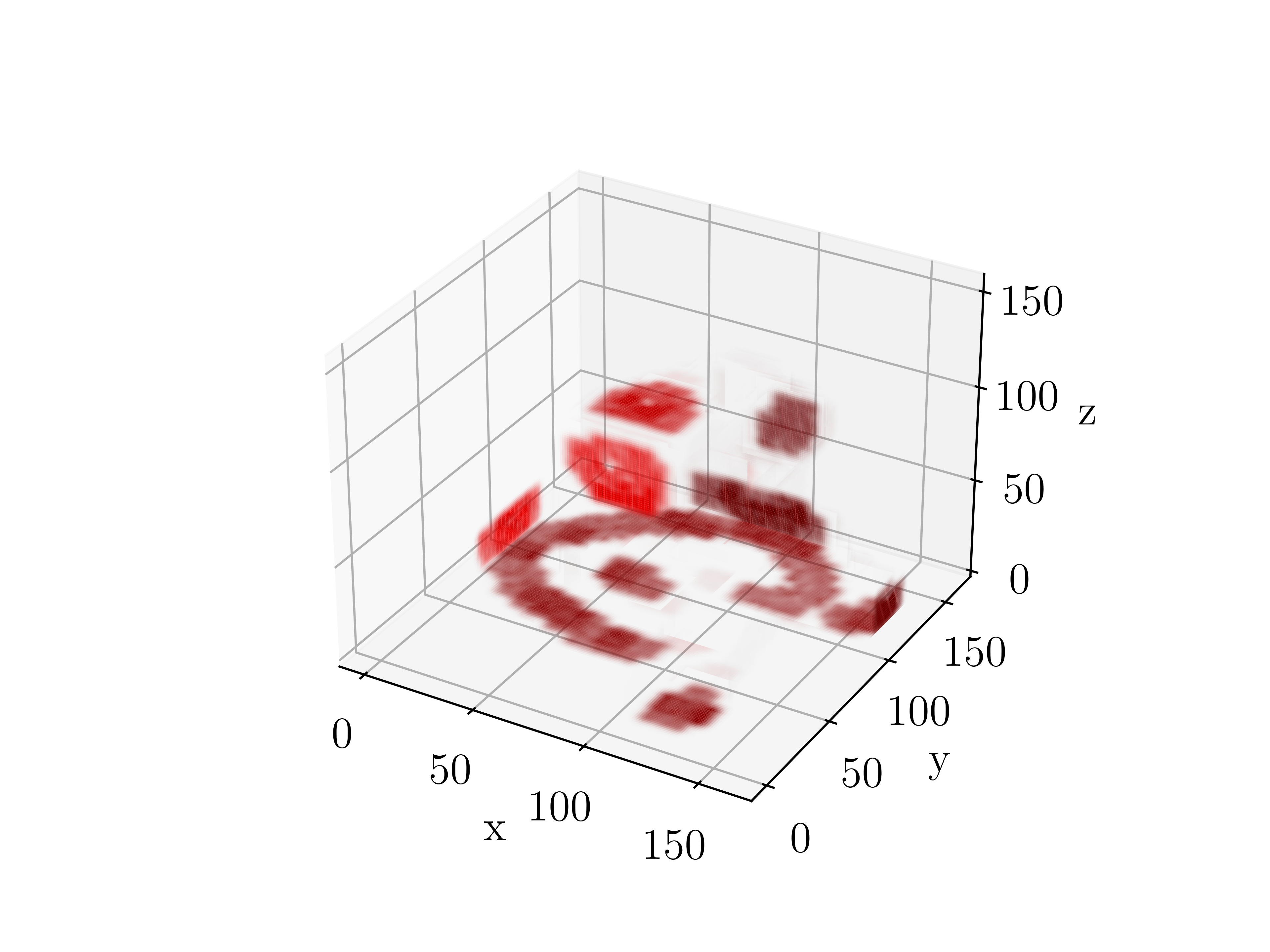}
    \caption{~}
    \label{fig:bbloccbase}
\end{subfigure}
\begin{subfigure}{0.35\textwidth}
    \includegraphics[width=\textwidth]{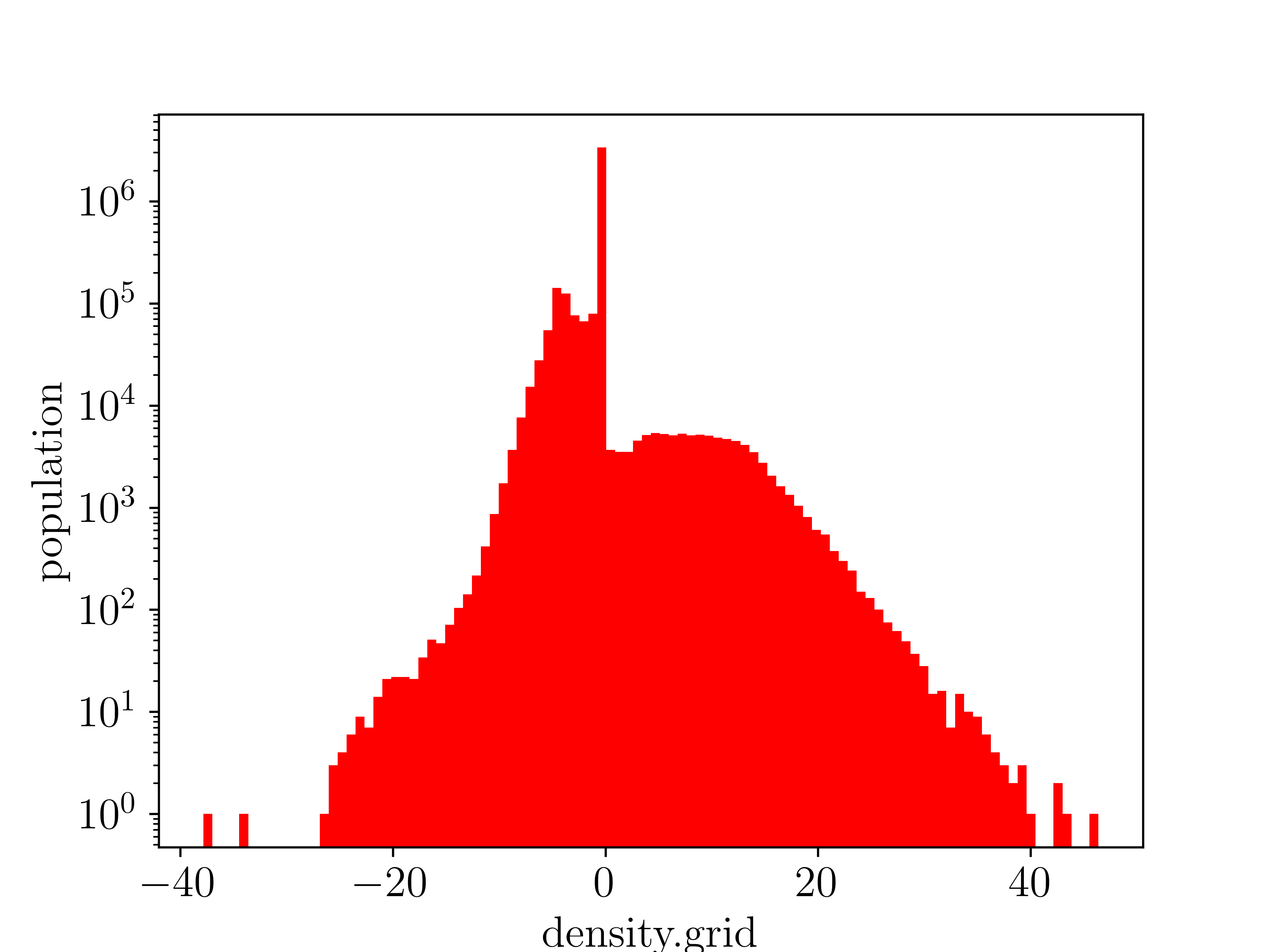}
    \caption{~}
    \label{fig:cbloccbase}
\end{subfigure}
\begin{subfigure}{0.3\textwidth}
    \includegraphics[width=\textwidth, trim={120 30 60 60},clip]{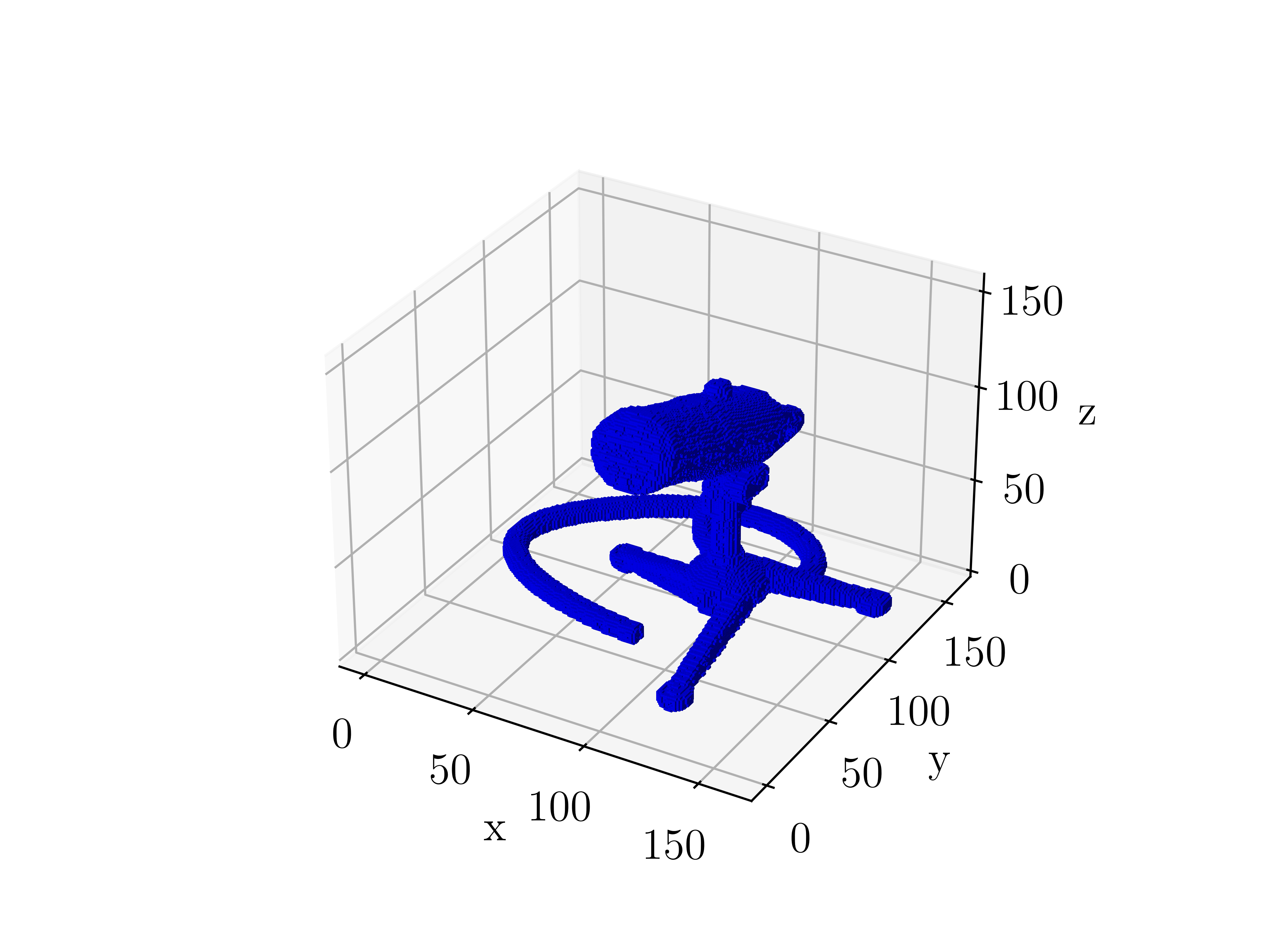}
    \caption{~}
    \label{fig:abloccmethod}
\end{subfigure}
\begin{subfigure}{0.3\textwidth}
    \includegraphics[width=\textwidth, trim={120 30 60 60},clip]{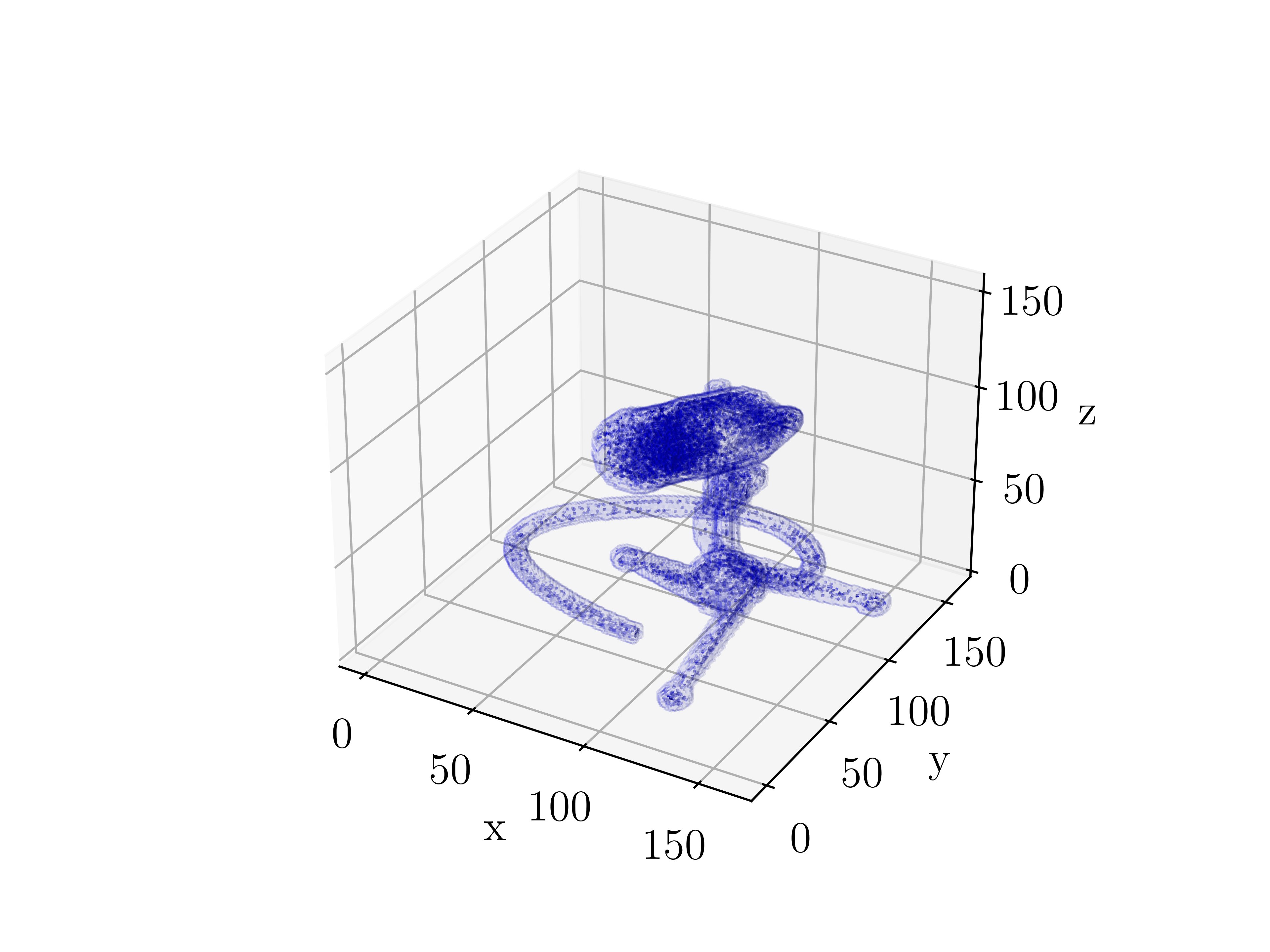}
    \caption{~}
    \label{fig:bbloccmethod}
\end{subfigure}
\begin{subfigure}{0.35\textwidth}
    \includegraphics[width=\textwidth]{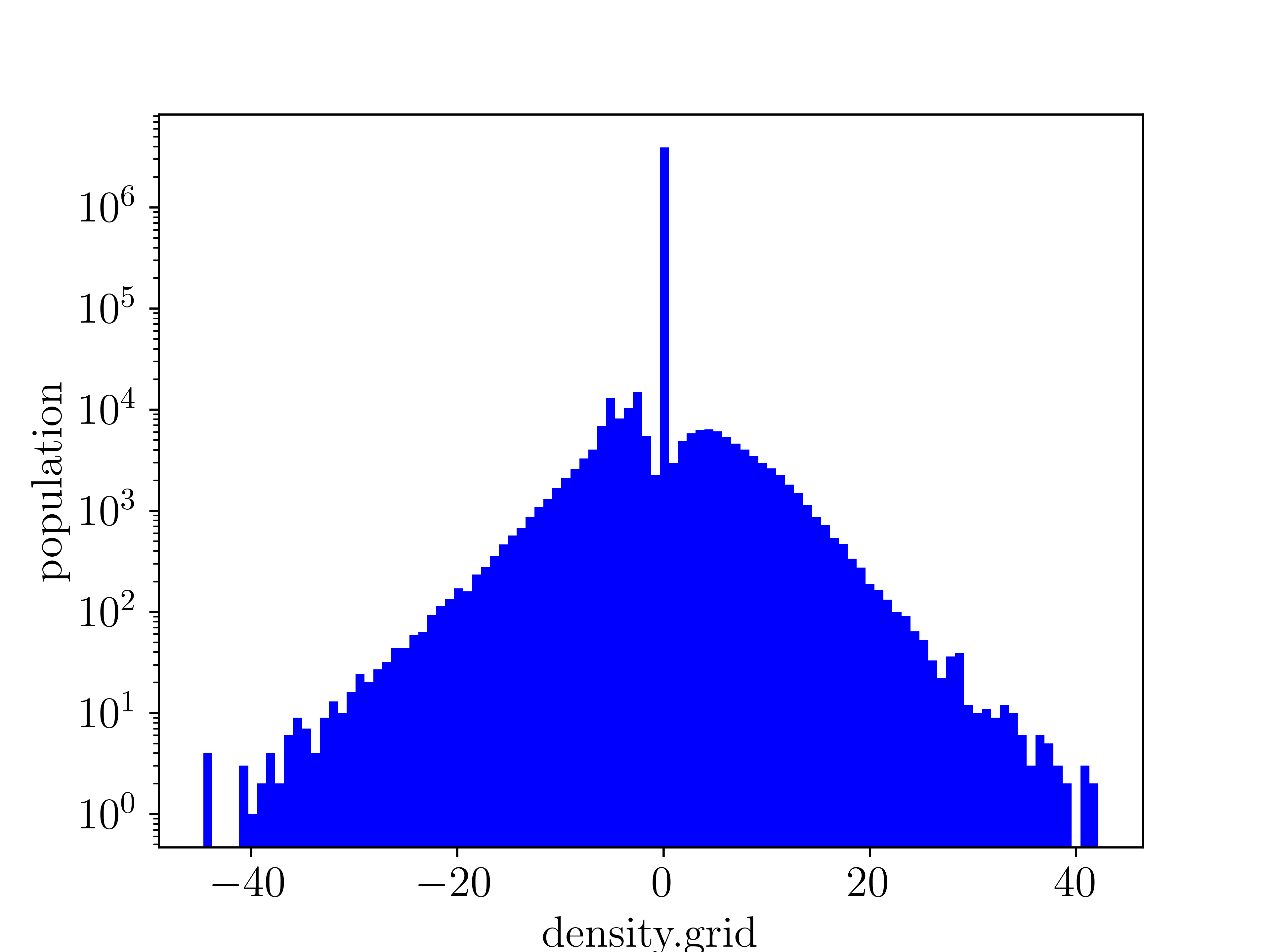}
    \caption{~}
    \label{fig:cbloccmethod}
\end{subfigure}
\caption{Visualization of the \texttt{density.grid} layer for DVGO~\cite{sun2022direct} trained on ``Mic'' (Synthetic-NeRF). Up: baseline model; down: \method~applied. Here are visualized the non-empty voxels (a, d), their effective value (b, e), and the distribution of their values, in log scale (c, f). For visualization, the values in (b) have been amplified by a factor 10$\times$.}
\label{fig:ablation}
\end{figure*}
\begin{table*}
    \caption{Ablation study conducted on ``Mic'' from Synthetic-NeRF. The approach used here is DVGO~\cite{sun2022direct}. The first line is the reference baseline.}
    \label{tab:ablation}
    \centering
    \renewcommand{\arraystretch}{1.2}
    \begin{tabular}{c c c c c c c c c}
        \toprule
            &&&&\multicolumn{2}{c}{\bf LOW compressibility}&\multicolumn{2}{c}{\bf HIGH compressibility}\\
        \bf Layers & \bf Remove & \bf Re-include & \bf Quantization & \textbf{PSNR}[dB]  & \textbf{Size}[MB]& \textbf{PSNR}[dB]  & \textbf{Size}[MB]\\
        \midrule
        \xmark &\xmark  & \xmark & \xmark &33.15&67.69 &- &-  \\
        All & \cmark & \xmark & \xmark & 26.72&6.88 & 25.09 & 0.87\\
        Voxels & \cmark & \xmark & \xmark &  33.19 & 7.02 & 27.67 & 1.08\\
        Voxels & \cmark & \cmark & \xmark &33.26 & 7.12 & 29.54 & 1.24 \\
        Voxels & \cmark & \cmark & \cmark  & 33.10 & 1.48 & 29.41 & 0.34 \\
        \bottomrule
    \end{tabular}
\end{table*}

In this section, we propose the ablation study for \method. In particular, we want to evidence the single contributions of the proposed technique, emphasizing their effect. Towards this end, we have conducted experiments on ``Mic'' from the Synthetic-NeRF dataset and used DVGO~\cite{sun2022direct} as the EVG-NeRF approach. The summary for the ablation study is enclosed in Table~\ref{tab:ablation}. All the measures here proposed are averaged on 3 different runs.\\
\textbf{Remove all the layers or a subset of them?} 
Considering the heterogeneity of the layers in the EVG-NeRF approaches, it is not straightforward that removing parameters from all the layers is the best approach. Indeed, we observe that focusing on the layers with explicit voxel representation (indicated as ``Voxel'') leads to a similar PSNR as the baseline (33.19 dB) with a very high size reduction (from 67.69MB to 7.02MB). Focusing on all the layers of the model, as it would be done in a generic model pruning scheme~\cite{han2015learning, tartaglione2018learning, Frankle2019TheLT} leads to a very high drop in performance (26.72dB, namely -6.42dB when compared to the baseline). This shows how important it is to focus on voxels and designing specific solutions rather than relying on generic approaches.\\
\textbf{Re-including helps.} 
The proposed strategy needs a ``balancing'' for the voxel removal phase, which can be extreme. Towards this end, re-including come removed voxels slightly increases the size of the model, which however turns into performance recovery. In particular, by adding just 0.10MB we gain 0.07dB: please notice that the baseline PSNR is lower than the achieved performance with remove+re-include. This phenomenon is even more evident in the high compressibility regime, where we gain approximately 2dB with just 0.16MB added.\\
\textbf{Effect of quantization.} 
In traditional NeRF models quantizing is a delicate process, requiring non-uniform, custom quantization strategies~\cite{shi2022distilled}. In our case, however, quantizing on 8 bits maintains the performance to high PSNR values (losing 0.16 dB without additional fine-tuning) but significantly reduces the size of the compressed model (from 7.12MB to 1.48MB). This is very evident in the high compressibility result, where we move from 1.24MB to 0.34MB only.

\subsection{A deeper view on \method's effect} 
As a final analysis, we wish to test what happens in the voxel grid for a baseline model and for the same with \method~applied. Fig.~\ref{fig:ablation} visualizes the content of the \texttt{density.grid} layer for the baseline (up, in red) and for the compressed one (down, in blue). Looking at the spatial occupancy for the density grid, without \method~evidently we have a much higher than necessary voxel occupation (Fig.~\ref{fig:abloccbase}) which is trimmed to the real object shape by \method~(Fig.~\ref{fig:abloccmethod}). Looking at the effective value of each voxel (here normalized and modeled as transparency) we can easily guess the structure of the object in the \method~case (Fig.~\ref{fig:bbloccmethod}) while the density is so spread in the space for the baseline case that the object is almost impossible to distinguish (Fig.~\ref{fig:bbloccbase}). This has a clear effect on the distribution of the parameter's value for the layer: while in the baseline case we observe very different behavior for positive and negative values, making problems like compression and quantization harder (Fig.~\ref{fig:cbloccbase}), the distribution tends to be more specular when applying \method~(Fig.~\ref{fig:cbloccmethod}): this is due to both the suppression of irrelevant parameters in the model and to the exclusive re-inclusion of parameters having as neighbors others already included. 

\section{Conclusion \& future works}
\label{sec:conclusion}
In this work we have presented \method, an approach to compress NeRF models that utilizes explicit voxel grid representations. This approach removes parameters from the model, while at the same time ensures not to have a large drop in performance. This is achieved by a re-inclusion mechanism, which allows previously removed parameters that are neighbors of the remaining parameters to be re-included if they show high gradient loss. \method~is easily deployable for any model, having different architecture, training strategy, or objective function. For this reason, we have tested its effectiveness on three very different approaches: DVGO~\cite{sun2022direct}, where a part of the model learns the density and the other maps complex voxel dependencies with an MLP, TensoRF~\cite{chen2022tensorf} which learns a 4D grid and performs low-rank decomposition on the radiance fields, and Plenoxels~\cite{fridovich2022plenoxels} which optimized the voxel grid directly with no MLP supporting the learning. These approaches have been tested on four popular datasets, two synthetic and two from real images.\\
In all the cases, \method~is able to compress the approaches with compression rates scaling up to $80\times$. Reducing the storage memory required by these models, designed mainly to improve training and inference time but sacrificing storage memory when compared to the original NeRF~\cite{mildenhall2020nerf}, further emphasizes EVG-NeRF's benefits and pushes towards their large-scale deployability in memory-constrained or bandwidth-limited applications. Interestingly, in a low compressibility setup, the performance is essentially unharmed, while the model is effectively compressed. This opens the road towards the model's budget re-allocation, like efficient ensembling, towards further performance enhancement with specific memory constraints.

{\small
\bibliographystyle{ieee_fullname}
\bibliography{main}

\begin{thebibliography}{10}\itemsep=-1pt

\bibitem{boss2021nerd}
Mark Boss, Raphael Braun, Varun Jampani, Jonathan~T Barron, Ce Liu, and Hendrik
  Lensch.
\newblock Nerd: Neural reflectance decomposition from image collections.
\newblock In {\em Proceedings of the IEEE/CVF International Conference on
  Computer Vision}, pages 12684--12694, 2021.

\bibitem{chan2021pi}
Eric~R Chan, Marco Monteiro, Petr Kellnhofer, Jiajun Wu, and Gordon Wetzstein.
\newblock pi-gan: Periodic implicit generative adversarial networks for
  3d-aware image synthesis.
\newblock In {\em Proceedings of the IEEE/CVF conference on computer vision and
  pattern recognition}, pages 5799--5809, 2021.

\bibitem{chen2022tensorf}
Anpei Chen, Zexiang Xu, Andreas Geiger, Jingyi Yu, and Hao Su.
\newblock Tensorf: Tensorial radiance fields.
\newblock {\em ECCV}, 2022.

\bibitem{davis2012unstructured}
Abe Davis, Marc Levoy, and Fredo Durand.
\newblock Unstructured light fields.
\newblock In {\em Computer Graphics Forum}, volume~31, pages 305--314. Wiley
  Online Library, 2012.

\bibitem{debevec1996modeling}
Paul~E Debevec, Camillo~J Taylor, and Jitendra Malik.
\newblock Modeling and rendering architecture from photographs: A hybrid
  geometry-and image-based approach.
\newblock In {\em Proceedings of the 23rd annual conference on Computer
  graphics and interactive techniques}, pages 11--20, 1996.

\bibitem{flynn2019deepview}
John Flynn, Michael Broxton, Paul Debevec, Matthew DuVall, Graham Fyffe, Ryan
  Overbeck, Noah Snavely, and Richard Tucker.
\newblock Deepview: View synthesis with learned gradient descent.
\newblock In {\em Proceedings of the IEEE/CVF Conference on Computer Vision and
  Pattern Recognition}, pages 2367--2376, 2019.

\bibitem{Frankle2019TheLT}
Jonathan Frankle and Michael Carbin.
\newblock The lottery ticket hypothesis: Finding sparse, trainable neural
  networks.
\newblock {\em ICLR}, 2019.

\bibitem{fridovich2022plenoxels}
Sara Fridovich-Keil, Alex Yu, Matthew Tancik, Qinhong Chen, Benjamin Recht, and
  Angjoo Kanazawa.
\newblock Plenoxels: Radiance fields without neural networks.
\newblock In {\em Proceedings of the IEEE/CVF Conference on Computer Vision and
  Pattern Recognition}, pages 5501--5510, 2022.

\bibitem{gafni2021dynamic}
Guy Gafni, Justus Thies, Michael Zollhofer, and Matthias Nie{\ss}ner.
\newblock Dynamic neural radiance fields for monocular 4d facial avatar
  reconstruction.
\newblock In {\em Proceedings of the IEEE/CVF Conference on Computer Vision and
  Pattern Recognition}, pages 8649--8658, 2021.

\bibitem{gao2021dynamic}
Chen Gao, Ayush Saraf, Johannes Kopf, and Jia-Bin Huang.
\newblock Dynamic view synthesis from dynamic monocular video.
\newblock In {\em Proceedings of the IEEE/CVF International Conference on
  Computer Vision}, pages 5712--5721, 2021.

\bibitem{guo2022fast}
Pengsheng Guo, Miguel~Angel Bautista, Alex Colburn, Liang Yang, Daniel
  Ulbricht, Joshua~M Susskind, and Qi Shan.
\newblock Fast and explicit neural view synthesis.
\newblock In {\em Proceedings of the IEEE/CVF Winter Conference on Applications
  of Computer Vision}, pages 3791--3800, 2022.

\bibitem{han2015learning}
Song Han, Jeff Pool, John Tran, and William Dally.
\newblock Learning both weights and connections for efficient neural network.
\newblock {\em Advances in neural information processing systems}, 28, 2015.

\bibitem{huang2022stylizednerf}
Yi-Hua Huang, Yue He, Yu-Jie Yuan, Yu-Kun Lai, and Lin Gao.
\newblock Stylizednerf: consistent 3d scene stylization as stylized nerf via
  2d-3d mutual learning.
\newblock In {\em Proceedings of the IEEE/CVF Conference on Computer Vision and
  Pattern Recognition}, pages 18342--18352, 2022.

\bibitem{kaya2022neural}
Berk Kaya, Suryansh Kumar, Francesco Sarno, Vittorio Ferrari, and Luc Van~Gool.
\newblock Neural radiance fields approach to deep multi-view photometric
  stereo.
\newblock In {\em Proceedings of the IEEE/CVF Winter Conference on Applications
  of Computer Vision}, pages 1965--1977, 2022.

\bibitem{knapitsch2017tanks}
Arno Knapitsch, Jaesik Park, Qian-Yi Zhou, and Vladlen Koltun.
\newblock Tanks and temples: Benchmarking large-scale scene reconstruction.
\newblock {\em ACM Transactions on Graphics (ToG)}, 36(4):1--13, 2017.

\bibitem{kosiorek2021nerf}
Adam~R Kosiorek, Heiko Strathmann, Daniel Zoran, Pol Moreno, Rosalia Schneider,
  Sona Mokr{\'a}, and Danilo~Jimenez Rezende.
\newblock Nerf-vae: A geometry aware 3d scene generative model.
\newblock In {\em International Conference on Machine Learning}, pages
  5742--5752. PMLR, 2021.

\bibitem{lecun1989optimal}
Yann LeCun, John Denker, and Sara Solla.
\newblock Optimal brain damage.
\newblock {\em Advances in neural information processing systems}, 2, 1989.

\bibitem{lee2018snip}
Namhoon Lee, Thalaiyasingam Ajanthan, and Philip~HS Torr.
\newblock Snip: Single-shot network pruning based on connection sensitivity.
\newblock {\em ICLR}, 2019.

\bibitem{levin2010linear}
Anat Levin and Fredo Durand.
\newblock Linear view synthesis using a dimensionality gap light field prior.
\newblock In {\em 2010 IEEE Computer Society Conference on Computer Vision and
  Pattern Recognition}, pages 1831--1838. IEEE, 2010.

\bibitem{li2021neural}
Zhengqi Li, Simon Niklaus, Noah Snavely, and Oliver Wang.
\newblock Neural scene flow fields for space-time view synthesis of dynamic
  scenes.
\newblock In {\em Proceedings of the IEEE/CVF Conference on Computer Vision and
  Pattern Recognition}, pages 6498--6508, 2021.

\bibitem{li2020crowdsampling}
Zhengqi Li, Wenqi Xian, Abe Davis, and Noah Snavely.
\newblock Crowdsampling the plenoptic function.
\newblock In {\em European Conference on Computer Vision}, pages 178--196.
  Springer, 2020.

\bibitem{liu2020neural}
Lingjie Liu, Jiatao Gu, Kyaw Zaw~Lin, Tat-Seng Chua, and Christian Theobalt.
\newblock Neural sparse voxel fields.
\newblock {\em Advances in Neural Information Processing Systems},
  33:15651--15663, 2020.

\bibitem{martin2021nerf}
Ricardo Martin-Brualla, Noha Radwan, Mehdi~SM Sajjadi, Jonathan~T Barron,
  Alexey Dosovitskiy, and Daniel Duckworth.
\newblock Nerf in the wild: Neural radiance fields for unconstrained photo
  collections.
\newblock In {\em Proceedings of the IEEE/CVF Conference on Computer Vision and
  Pattern Recognition}, pages 7210--7219, 2021.

\bibitem{mildenhall2019local}
Ben Mildenhall, Pratul~P Srinivasan, Rodrigo Ortiz-Cayon, Nima~Khademi
  Kalantari, Ravi Ramamoorthi, Ren Ng, and Abhishek Kar.
\newblock Local light field fusion: Practical view synthesis with prescriptive
  sampling guidelines.
\newblock {\em ACM Transactions on Graphics (TOG)}, 38(4):1--14, 2019.

\bibitem{mildenhall2020nerf}
Ben Mildenhall, Pratul~P Srinivasan, Matthew Tancik, Jonathan~T Barron, Ravi
  Ramamoorthi, and Ren Ng.
\newblock Nerf: Representing scenes as neural radiance fields for view
  synthesis.
\newblock In {\em European conference on computer vision}, pages 405--421.
  Springer, 2020.

\bibitem{mueller2022instant}
Thomas M\"uller, Alex Evans, Christoph Schied, and Alexander Keller.
\newblock Instant neural graphics primitives with a multiresolution hash
  encoding.
\newblock {\em ACM Trans. Graph.}, 41(4):102:1--102:15, July 2022.

\bibitem{noguchi2021neural}
Atsuhiro Noguchi, Xiao Sun, Stephen Lin, and Tatsuya Harada.
\newblock Neural articulated radiance field.
\newblock In {\em Proceedings of the IEEE/CVF International Conference on
  Computer Vision}, pages 5762--5772, 2021.

\bibitem{reiser2021kilonerf}
Christian Reiser, Songyou Peng, Yiyi Liao, and Andreas Geiger.
\newblock Kilonerf: Speeding up neural radiance fields with thousands of tiny
  mlps.
\newblock In {\em Proceedings of the IEEE/CVF International Conference on
  Computer Vision}, pages 14335--14345, 2021.

\bibitem{schwarz2020graf}
Katja Schwarz, Yiyi Liao, Michael Niemeyer, and Andreas Geiger.
\newblock Graf: Generative radiance fields for 3d-aware image synthesis.
\newblock {\em Advances in Neural Information Processing Systems},
  33:20154--20166, 2020.

\bibitem{shi2022distilled}
Jinglei Shi and Christine Guillemot.
\newblock Distilled low rank neural radiance field with quantization for light
  field compression.
\newblock {\em arXiv preprint arXiv:2208.00164}, 2022.

\bibitem{shi2014light}
Lixin Shi, Haitham Hassanieh, Abe Davis, Dina Katabi, and Fredo Durand.
\newblock Light field reconstruction using sparsity in the continuous fourier
  domain.
\newblock {\em ACM Transactions on Graphics (TOG)}, 34(1):1--13, 2014.

\bibitem{srinivasan2021nerv}
Pratul~P Srinivasan, Boyang Deng, Xiuming Zhang, Matthew Tancik, Ben
  Mildenhall, and Jonathan~T Barron.
\newblock Nerv: Neural reflectance and visibility fields for relighting and
  view synthesis.
\newblock In {\em Proceedings of the IEEE/CVF Conference on Computer Vision and
  Pattern Recognition}, pages 7495--7504, 2021.

\bibitem{srinivasan2019pushing}
Pratul~P Srinivasan, Richard Tucker, Jonathan~T Barron, Ravi Ramamoorthi, Ren
  Ng, and Noah Snavely.
\newblock Pushing the boundaries of view extrapolation with multiplane images.
\newblock In {\em Proceedings of the IEEE/CVF Conference on Computer Vision and
  Pattern Recognition}, pages 175--184, 2019.

\bibitem{sun2022direct}
Cheng Sun, Min Sun, and Hwann-Tzong Chen.
\newblock Direct voxel grid optimization: Super-fast convergence for radiance
  fields reconstruction.
\newblock In {\em Proceedings of the IEEE/CVF Conference on Computer Vision and
  Pattern Recognition}, pages 5459--5469, 2022.

\bibitem{tancik2021learned}
Matthew Tancik, Ben Mildenhall, Terrance Wang, Divi Schmidt, Pratul~P
  Srinivasan, Jonathan~T Barron, and Ren Ng.
\newblock Learned initializations for optimizing coordinate-based neural
  representations.
\newblock In {\em Proceedings of the IEEE/CVF Conference on Computer Vision and
  Pattern Recognition}, pages 2846--2855, 2021.

\bibitem{tartaglione2022loss}
Enzo Tartaglione, Andrea Bragagnolo, Attilio Fiandrotti, and Marco Grangetto.
\newblock Loss-based sensitivity regularization: towards deep sparse neural
  networks.
\newblock {\em Neural Networks}, 146:230--237, 2022.

\bibitem{tartaglione2021serene}
Enzo Tartaglione, Andrea Bragagnolo, Francesco Odierna, Attilio Fiandrotti, and
  Marco Grangetto.
\newblock Serene: Sensitivity-based regularization of neurons for structured
  sparsity in neural networks.
\newblock {\em IEEE Transactions on Neural Networks and Learning Systems},
  2021.

\bibitem{tartaglione2018learning}
Enzo Tartaglione, Skjalg Leps{\o}y, Attilio Fiandrotti, and Gianluca Francini.
\newblock Learning sparse neural networks via sensitivity-driven
  regularization.
\newblock {\em Advances in neural information processing systems}, 31, 2018.

\bibitem{thies2019deferred}
Justus Thies, Michael Zollh{\"o}fer, and Matthias Nie{\ss}ner.
\newblock Deferred neural rendering: Image synthesis using neural textures.
\newblock {\em ACM Transactions on Graphics (TOG)}, 38(4):1--12, 2019.

\bibitem{tretschk2021non}
Edgar Tretschk, Ayush Tewari, Vladislav Golyanik, Michael Zollh{\"o}fer,
  Christoph Lassner, and Christian Theobalt.
\newblock Non-rigid neural radiance fields: Reconstruction and novel view
  synthesis of a dynamic scene from monocular video.
\newblock In {\em Proceedings of the IEEE/CVF International Conference on
  Computer Vision}, pages 12959--12970, 2021.

\bibitem{waechter2014let}
Michael Waechter, Nils Moehrle, and Michael Goesele.
\newblock Let there be color! large-scale texturing of 3d reconstructions.
\newblock In {\em European conference on computer vision}, pages 836--850.
  Springer, 2014.

\bibitem{xian2021space}
Wenqi Xian, Jia-Bin Huang, Johannes Kopf, and Changil Kim.
\newblock Space-time neural irradiance fields for free-viewpoint video.
\newblock In {\em Proceedings of the IEEE/CVF Conference on Computer Vision and
  Pattern Recognition}, pages 9421--9431, 2021.

\bibitem{yu2021pixelnerf}
Alex Yu, Vickie Ye, Matthew Tancik, and Angjoo Kanazawa.
\newblock pixelnerf: Neural radiance fields from one or few images.
\newblock In {\em Proceedings of the IEEE/CVF Conference on Computer Vision and
  Pattern Recognition}, pages 4578--4587, 2021.

\end{thebibliography}
}
\appendix
\section{Detailed results}
Here follow the detailed tables for Synthetic-NeRF (Table~\ref{tab:syntheticnerf}), Synthetic-NSVF (Table~\ref{tab:syntheticnsvf}) and Tanks \& Temples (Table~\ref{tab:tandt}). For these, we also include, as output quality metric, the LPIPS score evaluated on the VGG backbone.\\
Besides, we also provide evaluations on the BlendedMVS dataset for DVGO (Table~\ref{tab:blendedmvf}), whose configuration follows.

\subsection{Configuration for BlendedMVS}
BlendedMVS, although being a synthetic dataset, differently from Synthetic-NeRF and Synthetic-NSVF, has more realistic ambient lighting, which is taken from real image blending. In this case, following the same approach as \cite{sun2022direct}, we have used a subset of 4 objects (namely: \textit{jade, fountain, character} and \textit{statue}). We have here used as image resolution 768 × 576 pixels; $10\%$ of the images are used for validation and $10\%$ for testing. For Re:NeRF, we have used $\gamma=0.5$ and $\delta=0.5$, with $\Delta T=1$dB.

\begin{table*}
    \caption{Results for Synthetic-NeRF.}
    \label{tab:syntheticnerf}
	\renewcommand{\arraystretch}{1.2}
	\centering
    \resizebox{\textwidth}{!}{
		\begin{tabular}{c c c c c c c c c c c c c c c c c c c }
        \toprule
        &\multicolumn{10}{c}{\bf Synthetic-NeRF}\\
        &Architecture & Pruning & Chair & Drums & Ficus & Hotdog & Lego & Materials & Mic & Ship &\bf Avg\\
        \hline
         \multirow{9}{*}{PSNR(dB) ($\uparrow$)}                  
         & & - &34.11 & 25.48 & 32.59& 36.77 & 34.69 & 29.52 & 33.16 &29.04& \bf 31.92\\
         &DVGO~\cite{sun2022direct} & LOW & 33.75 & 25.34 &32.36 &36.00 &34.30&29.26&33.16 &28.69& \bf 31.61\\
         && HIGH & 33.45 & 24.98& 32.11 & 35.44 & 33.90 &28.38&32.77&28.60& \bf 31.20\\
         \cline{2-12}
         & & - &33.98 & 25.35 & 31.83& 36.43 & 34.10 & 29.14 & 33.26 &27.78 &\bf 31.48\\
         &Plenoxels~\cite{fridovich2022plenoxels} & LOW & 34.35 & 25.09 &31.69 &36.33 &34.40&28.73&33.92 &27.71& \bf 31.52\\
         & & HIGH & 33.65 & 24.96& 31.21 & 35.44 & 33.91 & 28.15&33.14&27.31& \bf 30.97\\
         \cline{2-12}
        & & - &35.76& 26.01& 33.99& 37.41& 36.46& 30.12 &34.61 &30.77& \bf 33.14 \\
        &TensoRF~\cite{chen2022tensorf} & LOW &36.00&26.01&34.12&37.58&36.73&30.01&34.67&30.92&\bf33.26\\
        && HIGH& 35.66&25.59&33.57&37.35&36.38&29.64&34.00&30.31&\bf32.81 \\
        \hline
         \multirow{9}{*}{Size(MB)($\downarrow $)} 
         & & - & 103.994 & 92.06 &108.71 &130.01&124.201& 171.36& 49.41 &100.98&\bf 110.09\\
         &DVGO~\cite{sun2022direct} & LOW & 4.44 & 2.62 & 2.67 &5.19& 5.38&7.47&1.50 & 5.43 &\bf 4.33\\
         && HIGH &2.53 & 1.21 & 1.82 &2.41&2.85 &3.06&0.89 & 2.85&\bf 2.20\\
        \cline{2-12}
        & & - & 187.04 &160.83 &108.39 &290.64&291.65& 196.15& 80.83 &197.08&\bf 189.08\\
         &Plenoxels~\cite{fridovich2022plenoxels} & LOW & 85.59&56.34&38.51&179.26&183.48&91.25&37.69&62.07&\bf 91.77\\
         & & HIGH &47.26&42.34&28.8&96.81&97.78&66.4&21.07&36.94&\bf 54.68\\
         \cline{2-12}
        & & - &65.46&65.62&67.87&77.61&65.75&80.06&64.45&67.12&\bf69.24\\
        &TensoRF~\cite{chen2022tensorf} & LOW &12.78&8.47&13.30&14.77&12.82&15.65&7.67&8.77&\bf11.78\\
        && HIGH& 8.51&6.11&8.88&9.47&8.41&10.31&5.58&6.26&\bf7.94 \\
         \hline
         \multirow{9}{*}{SSIM($\uparrow$)}
        & & - &0.976 & 0.930 &0.977& 0.986& 0.976 & 0.950 & 0.983 &0.878& \bf 0.957 \\
         &DVGO~\cite{sun2022direct} & LOW & 0.974 & 0.924 &0.975 &0.973&0.973 &0.943&0.982&0.871 & \bf 0.952 \\
         & & HIGH & 0.971& 0.916& 0.962 & 0.969 & 0.967 &0.915&0.981&0.867& \bf 0.943 \\
         \cline{2-12}
        & & - &0.977&0.933&0.976&0.98&0.975&0.949&0.985&0.869& \bf 0.956\\
         &Plenoxels~\cite{fridovich2022plenoxels} & LOW &0.978&0.922&0.97&0.979&0.976&0.939&0.985&0.867&\bf0.952\\
         && HIGH &0.971&0.915&0.97&0.972&0.972&0.923&0.977&0.854&\bf0.944\\
         \cline{2-12}
        && - &0.985 &0.937& 0.982& 0.982& 0.983& 0.952& 0.988 &0.895&\bf0.963\\
        &TensoRF~\cite{chen2022tensorf} & LOW &0.985&0.931&0.982&0.982&0.983&0.947&0.987&0.894&\bf0.962\\
        & & HIGH& 0.983&0.917&0.979&0.980&0.981&0.939&0.984&0.882&\bf0.956 \\
        \hline
         \multirow{9}{*}{LPIPS$_{VGG}$($\downarrow$)}
        && - &0.027 & 0.079 &0.025& 0.034& 0.027 & 0.059 & 0.018 &0.161& \bf 0.054\\
         &DVGO~\cite{sun2022direct} & LOW & 0.035 & 0.090 &0.033 &0.060 &0.032&0.070&0.022 &0.167& \bf 0.064\\
         & & HIGH & 0.038& 0.103& 0.037 & 0.067 & 0.039 &0.102&0.026&0.171& \bf 0.073\\
         \cline{2-12}
        && - &0.031&0.067&0.026&0.037&0.028&0.057&0.015&0.178&\bf0.055\\
         &Plenoxels~\cite{fridovich2022plenoxels} & LOW & 0.026&0.081&0.038&0.043&0.027&0.074&0.019&0.18&\bf0.061\\
         && HIGH & 0.033&0.088&0.044&0.062&0.033&0.090&0.031&0.193&\bf0.072\\
         \cline{2-12}
        & & - &0.022& 0.073 &0.022 &0.032 &0.018 &0.058& 0.015& 0.138& 0.047\\
        &TensoRF~\cite{chen2022tensorf} & LOW &0.022&0.103&0.026&0.035&0.018&0.067&0.022&0.138&\bf0.054\\
        & & HIGH& 0.028&0.157&0.042&0.045&0.022&0.081&0.042&0.159&\bf0.072 \\
         \bottomrule
	\end{tabular}
    }
\end{table*}
\begin{table*}
    \caption{Results for Synthetic-NSVF.}
    \label{tab:syntheticnsvf}
	\renewcommand{\arraystretch}{1.2}
	\centering
    \resizebox{\textwidth}{!}{
		\begin{tabular}{c c c c c c c c c c c c c c c c c c c }
        \toprule
         &\multicolumn{11}{c}{\bf Synthetic-NSVF}\\
         &Architecture & Pruning &Bike &Lifestyle &Palace &Robot&Spaceship & Steamtrain &Toad &Wineholder &\bf Avg\\
         \hline
        \multirow{6}{*}{PSNR(dB)($\uparrow$)} 
        &  & - &38.13 & 33.64 & 34.32& 36.23 &37.56 & 36.47 & 33.02 &30.21& \bf 34.95 \\
         & DVGO~\cite{sun2022direct} & LOW & 38.16 & 33.68& 34.47 &36.29 &37.26&36.10&32.98 &30.11& \bf 34.88\\
         &  & HIGH & 37.97 & 33.16& 33.88 & 36.00 & 36.82 &35.79&32.39&29.75& \bf 34.47\\
         \cline{2-12}
        & & -&39.23&34.51&37.56&38.26&38.6&37.87&31.32&34.85&\bf36.53 \\
         &TensoRF~\cite{chen2022tensorf}& LOW &39.38&34.68&37.92&38.72&38.58&38.06&34.85&31.77&\bf36.75\\
         & & HIGH & 38.90&34.33&37.53&38.40&38.10&37.40&33.20&31.23& \bf36.14 \\
         \hline
        \multirow{6}{*}{Size(MB)($\downarrow$)}  
        & & - &104.10 & 97.12 & 105.00& 97.17 &128.31 &144.64 & 128.30 &97.71& \bf 112.79 \\
         &DVGO~\cite{sun2022direct} & LOW & 3.55 &3.53& 4.84 &3.76&4.95&5.41&5.67 &3.30& \bf 4.38 \\
         & & HIGH & 2.52 & 2.38& 2.63 & 2.55 &2.78 &2.77&1.87&1.65& \bf2.39 \\
         \cline{2-12}
         & & - &70.92&65.46&64.94&68.63&68.01&80.02&68.71&65.74&\bf69.05 \\
         &TensoRF~\cite{chen2022tensorf}& LOW &13.74&12.84&12.67&13.15&13.38&15.39&14.24&12.77&\bf13.52\\
         & & HIGH & 9.11&8.47&8.36&8.80&8.83&7.12&9.12&8.35&\bf8.52 \\
        \hline
         \multirow{6}{*}{SSIM($\uparrow$)}  
        & & - &0.991 & 0.964 &0.992& 0.992 &0.987 &0.989 & 0.965 &0.949& \bf 0.979\\
         &DVGO~\cite{sun2022direct} & LOW &0.991 & 0.963 &0.961&0.991&0.985&0.986 &0.966& 0.950&\bf 0.974\\
         & & HIGH &0.990 & 0.958& 0.953 & 0.991 &0.982 &0.982&0.957&0.942& \bf0.969\\
         \cline{2-12}
        & & -  &0.993&0.968&0.979&0.994&0.989&0.991&0.961&0.978&\bf0.982\\
         &TensoRF~\cite{chen2022tensorf}& LOW &0.993&0.968&0.980&0.995&0.988&0.991&0.978&0.963&\bf0.982\\
         & & HIGH & 0.992&0.963&0.978&0.994&0.985&0.988&0.966&0.957&\bf0.978 \\
         \hline
         \multirow{6}{*}{LPIPS$_{VGG}$($\downarrow$)} 
           & & - &0.011 & 0.055 &0.045& 0.013&0.020 &0.019& 0.047 &0.059& \bf 0.034\\
         &DVGO~\cite{sun2022direct} & LOW &0.015 & 0.056&0.043&0.013&0.024&0.027 &0.045& 0.055&\bf 0.035\\
         & & HIGH &0.015 & 0.063& 0.050 & 0.013 &0.028 &0.036&0.054&0.067& \bf 0.041\\
         \cline{2-12}
         & & - & 0.003&0.021&0.011&0.003&0.009&0.006&0.024&0.016&\bf0.012\\
         &TensoRF~\cite{chen2022tensorf}& LOW &0.011&0.049&0.020&0.010&0.022&0.017&0.035&0.054&\bf0.027\\
         & & HIGH &0.016&0.061&0.022&0.011&0.027&0.029&0.059&0.077&\bf0.038 \\
         \bottomrule
	\end{tabular}
    }
\end{table*}
\begin{table*}
    \caption{Results for BlendedMVS}
    \label{tab:blendedmvf}
	\renewcommand{\arraystretch}{1.2}
	\centering
		\begin{tabular}{c c c c c c c c c c c c c c c c c c c }
        \toprule
         &\multicolumn{7}{c}{\bf BlendedMVS}\\
         &Architecture & Pruning& Character & Fountain & Jade & Statue&\bf Avg\\
         \hline
        \multirow{3}{*}{PSNR(dB)($\uparrow$)} 
            && - &30.26 & 28.27 &27.75& 26.41  &\bf28.17  \\
            &DVGO~\cite{sun2022direct} & LOW &30.05 & 28.21&27.48& 26.08  &\bf 27.86  \\
            & & HIGH &29.78 &27.90&27.08& 25.97  &\bf  27.68\\

        \hline
        \multirow{3}{*}{Size(MB)($\downarrow$)}
            & & - &131.73 & 72.33 &158.03& 97.08 & \bf114.79  \\
            &DVGO~\cite{sun2022direct} & LOW &5.43 & 3.54&5.24& 2.81  &\bf 4.26 \\
            & & HIGH &2.92 &1.70&1.85& 1.85  &\bf 2.08\\
        \hline
         \multirow{3}{*}{SSIM($\uparrow$)}
            & & - & 0.963 & 0.923 & 0.916& 0.887 &\bf 0.922 \\
            &DVGO~\cite{sun2022direct} & LOW & 0.960 & 0.921& 0.909& 0.876 & \bf 0.917 \\
            & & HIGH & 0.957 & 0.910& 0.887& 0.867 & \bf  0.908\\
         \hline
         \multirow{3}{*}{LPIPS$_{VGG}$($\downarrow$)}
            && - & 0.046 & 0.116 & 0.106& 0.137 &\bf  0.101\\
            &DVGO~\cite{sun2022direct} & LOW & 0.048 & 0.114& 0.107& 0.142 & \bf 0.103 \\
            & & HIGH & 0.052 & 0.126& 0.129& 0.151 & \bf 0.115 \\
         \bottomrule
	\end{tabular}
\end{table*}
\begin{table*}
    \caption{Results for Tanks \& Temples.}
    \label{tab:tandt}
    \small
	\renewcommand{\arraystretch}{1.2}
	\centering
		\begin{tabular}{c c c c c c c c c c c c c c c c c c c }
        \toprule
         &\multicolumn{8}{c}{\bf Tanks \& Temples}\\
         &Architecture & Pruning &Barn&Caterpillar&Family& Ignatius & Truck  & \bf Avg\\
         \hline
         \multirow{6}{*}{PSNR(dB)($\uparrow$)}  
            & & - &26.84 & 25.70 &33.68& 28.00 &27.09 &  \bf 28.26\\
            &DVGO~\cite{sun2022direct} & LOW &26.76 & 25.67&33.60& 28.06&27.04&  \bf  28.23\\
            & & HIGH &26.32 &25.22&33.36& 27.86&26.78  &\bf 27.91 \\
        \cline{2-9}    
        & & - &25.95&24.63&32.25&27.49&26.52&\bf27.37\\
         &Plenoxels~\cite{fridovich2022plenoxels} & LOW & 26.58&24.78&32.86&27.22&26.87&\bf27.66\\
         & & HIGH &26.31&24.40&32.29&27.00&26.69 &\bf27.34\\ 
          \cline{2-9}    
             & & - &28.34&27.14&27.22&26.19&33.92&  \bf 28.56\\
            &TensoRF & LOW &27.28&26.09&33.75&28.06& 27.32&\bf28.50\\
            & & HIGH &26.99&25.77&33.36&27.86 &27.20 &\bf 28.40\\
        \hline
         \multirow{6}{*}{Size(MB)($\downarrow$)}
            & & - & 128.21 & 109.94 &92.72&95.10&106.43 &\bf  106.48\\
            &DVGO~\cite{sun2022direct} & LOW &5.52 & 5.23&3.85& 3.51&5.35&  \bf4.69  \\
            & & HIGH &1.75 &1.89&2.37& 1.08&1.87 &\bf 1.79 \\
        \cline{2-9}
         & & - &282.85&133.43&103.64&115.81&104.08&\bf147.96\\
         &Plenoxels~\cite{fridovich2022plenoxels} & LOW &213.45&87.08&73.16&93.94&43.65&\bf102.26\\
         & & HIGH &181.09&66.67&60.53&84.06&35.01&\bf85.47\\ 
                   \cline{2-9}    
             & & - &73.95&64.56&60.06&61.25&65.36&\bf65.04\\
            &TensoRF & LOW &9.39&8.16&7.40&12.13& 12.87 & \bf9.99\\
            & & HIGH &6.62&5.69&5.17&7.66& 8.34 & \bf6.70\\
        \hline
         \multirow{6}{*}{SSIM($\uparrow$)}
          & & - & 0.836 & 0.904 & 0.961& 0.941 & 0.905 &\bf0.909  \\
            &DVGO~\cite{sun2022direct} & LOW & 0.838 & 0.904& 0.962& 0.941& 0.904&  \bf0.910  \\
            & & HIGH & 0.826 & 0.859& 0.958& 0.931 & 0.895 &\bf0.894  \\
        \cline{2-9}
        & & - &0.828&0.899&0.954&0.942&0.899&\bf0.904\\
         &Plenoxels~\cite{fridovich2022plenoxels} & LOW &0.856&0.894&0.959&0.935&0.902&\bf0.909\\
         & & HIGH &0.844&0.871&0.95&0.923&0.892&\bf0.896\\ 
                            \cline{2-9}    
             & & - &0.948&0.914&0.864&0.912&0.965&\bf0.920\\
            &TensoRF & LOW &0.862&0.901&0.961&0.941&0.913 & \bf0.916\\
            & & HIGH &0.852&0.888&0.956&0.934& 0.907 &\bf0.907\\
         \hline
         \multirow{6}{*}{LPIPS$_{VGG}$($\downarrow$)} 
            & & - & 0.297 & 0.171 & 0.071& 0.089 & 0.162& \bf  0.158\\
            &DVGO~\cite{sun2022direct} & LOW & 0.291 & 0.172& 0.079& 0.091& 0.161&  \bf 0.159 \\
            & & HIGH & 0.312 & 0.194& 0.073& 0.107 & 0.174 &\bf 0.172  \\
        \cline{2-9}
        & & - &0.306&0.169&0.081&0.102&0.167&\bf0.165\\
         &Plenoxels~\cite{fridovich2022plenoxels} & LOW &0.263&0.174&0.071&0.112&0.155&\bf0.155\\
         & & HIGH &0.285&0.200&0.081&0.126&0.165&\bf0.171\\ 
                                     \cline{2-9}    
             & & - &0.078&0.145&0.252&0.159&0.064 &\bf0.140\\
            &TensoRF & LOW &0.258&0.187&0.067&0.087& 0.149 & \bf0.149\\
            & & HIGH &0.277&0.211&0.077&0.096& 0.169 & \bf0.166\\
        \bottomrule
	\end{tabular}
\end{table*}

\begin{table*}
    \caption{Results for forward-facing scenes from NeRF}
    \label{tab:llff_nerf}
	\renewcommand{\arraystretch}{1.2}
	\centering
    \resizebox{\textwidth}{!}{
		\begin{tabular}{c c c c c c c c c c c c c c c c c c c }
        \toprule
         &\multicolumn{11}{c}{\bf Forward-facing-NeRF}\\
         &Architecture & Pruning &Fern&Flower&Fortress&Horns&Leaves&Orchids&Room&Trex&\bf Avg\\
         \hline
        \multirow{6}{*}{PSNR(dB)($\uparrow$)} 
        &  & - &24.57&27.64&30.17&27.10&21.56&20.54&29.20&26.41& \bf 25.90 \\
         & Plenoxel~\cite{fridovich2022plenoxels}& LOW & 25.45&27.82&30.57&27.51&21.38 &20.37&30.33&26.50 & \bf 26.24\\
         &  & HIGH & 25.23&27.65&30.18&26.98&21.12&20.32&29.76&26.27& \bf 25.94\\
         \cline{2-12}
        & & -&25.27&28.60&31.36&28.14&21.30&19.87&32.35&26.97&\bf26.73 \\
         &TensoRF~\cite{chen2022tensorf}& LOW &24.50&28.64&31.30&28.87&21.22&19.31&33.33&27.26&\bf26.80\\
         & & HIGH & 24.40&28.29&31.09&28.49&20.78&19.08&33.10&27.16&\bf26.55 \\
         \hline
        \multirow{6}{*}{Size(MB)($\downarrow$)}  
        & & - &1658.40&1471.78&1407.66&1726.17&1851.71&720.01&1421.66&1622.31&\bf1484.96 \\
          & Plenoxel~\cite{fridovich2022plenoxels} & LOW & 407.02&726.00&383.61&432.01&438.90 &221.51&495.84&552.94& \bf 457.23\\
         & & HIGH & 305.97&523.17&291.59&321.62&327.51&163.36&366.04&404.93& \bf338.02 \\
         \cline{2-12}
         & & - &148.49&152.44&149.99&152.51&151.72&159.80&151.14&148.20&\bf151.79\\
         &TensoRF~\cite{chen2022tensorf}& LOW &19.26&30.72&19.69&31.29&19.66&21.96&86.23&29.93&\bf32.34\\
         & & HIGH & 12.97&19.44&13.36&20.17&13.38&15.29&48.47&19.09&\bf20.27\\
        \hline
         \multirow{6}{*}{SSIM($\uparrow$)}  
        & & - &0.830&0.863&0.884&0.857&0.763&0.681&0.937&0.890&\bf0.838\\
          & Plenoxel~\cite{fridovich2022plenoxels} & LOW &0.831&0.862&0.880&0.859& 0.758&0.684&0.938&0.895&\bf 0.838\\
         & & HIGH &0.821&0.858&0.873&0.840&0.734&0.681&0.927&0.889& \bf0.828\\
         \cline{2-12}
        & & -  &0.814&0.871&0.897&0.877&0.752&0.649&0.952&0.900&\bf0.839\\
         &TensoRF~\cite{chen2022tensorf}& LOW &0.764&0.864&0.891&0.892&0.725&0.570&0.955&0.900&\bf0.820\\
         & & HIGH & 0.744&0.842&0.880&0.873&0.677&0.520&0.950&0.889&\bf0.797 \\
         \hline
         \multirow{6}{*}{LPIPS$_{VGG}$($\downarrow$)} 
           & & - &0.225&0.177&0.180&0.230&0.194&0.271&0.194&0.237&\bf0.213\\
          & Plenoxel~\cite{fridovich2022plenoxels}& LOW &0.225&0.177&0.183&0.228&0.197 &0.264&0.199&0.234&\bf 0.213\\
         & & HIGH &0.241&0.178&0.188&0.254& 0.226&0.265&0.234&0.250& \bf 0.229\\
         \cline{2-12}
         & & - & 0.237&0.169&0.148&0.196&0.217&0.278&0.167&0.221&\bf0.204\\
         &TensoRF~\cite{chen2022tensorf}& LOW &0.299&0.158&0.144&0.158&0.299&0.383&0.149&0.203&\bf0.224\\
         & & HIGH &0.337&0.200&0.172&0.193&0.364&0.449&0.168&0.227&\bf0.264 \\
         \bottomrule
	\end{tabular}
    }
\end{table*}

 \begin{table*}
     \caption{Examples generated from the Synthetic-NeRF dataset with TensoRF.}
     \centering
     \resizebox{\textwidth}{!}{
     \begin{tabular}{c c c c}
     \toprule
     \bf Ground Truth & \bf Baseline & \bf LOW compression & \bf HIGH compression\\
     \midrule
      \includegraphics[width=0.24\textwidth]{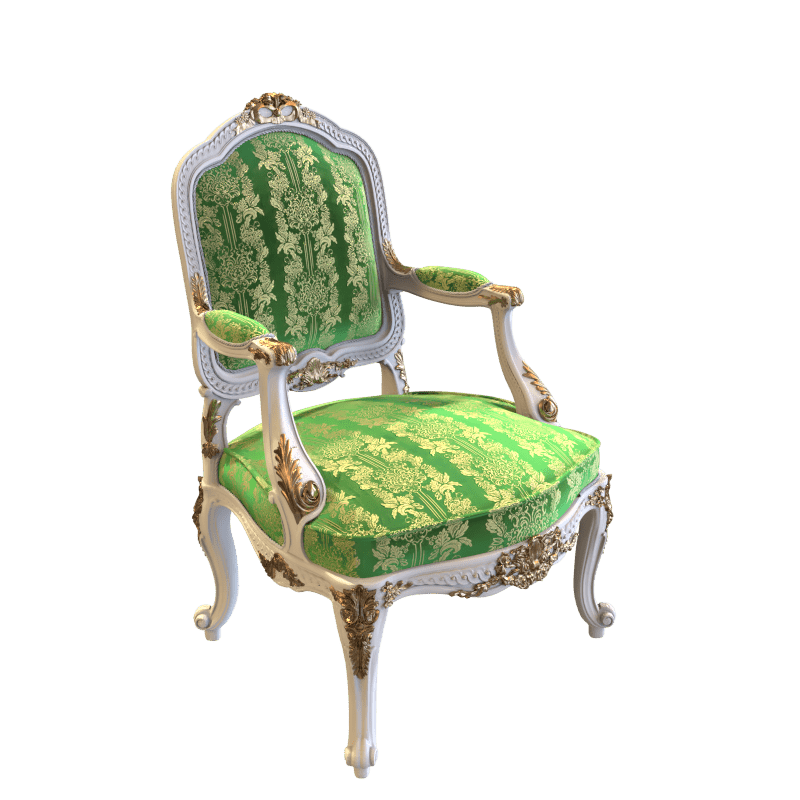}&
      \includegraphics[width=0.24\textwidth]{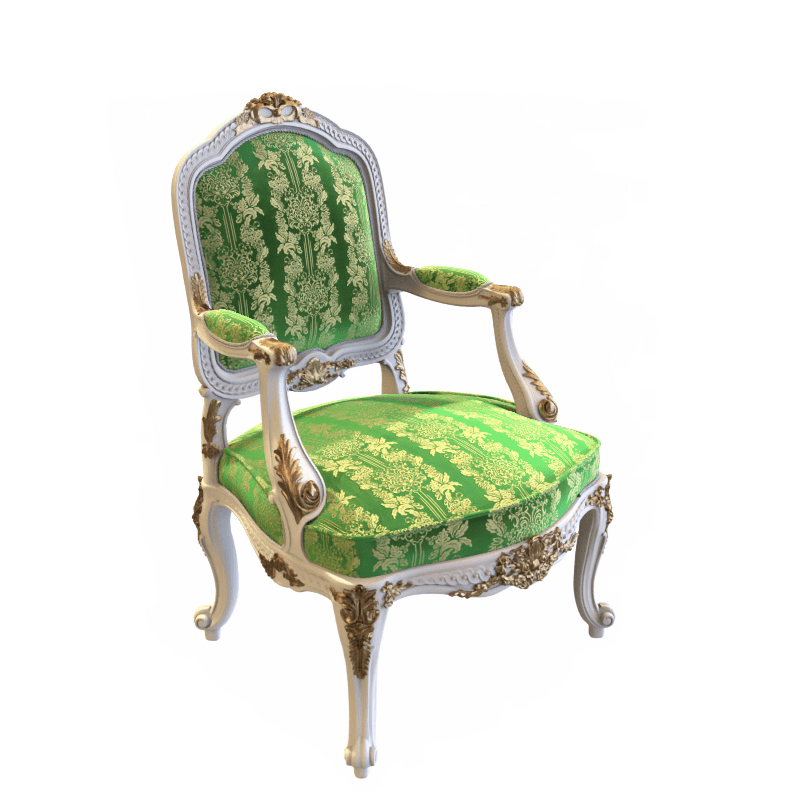}&
      \includegraphics[width=0.24\textwidth]{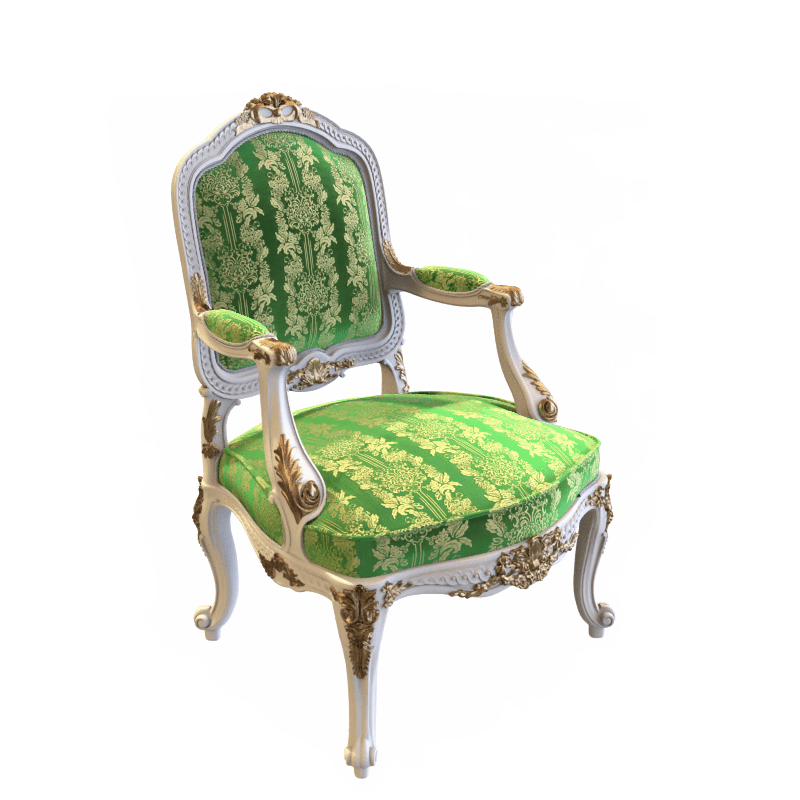}&
      \includegraphics[width=0.24\textwidth]{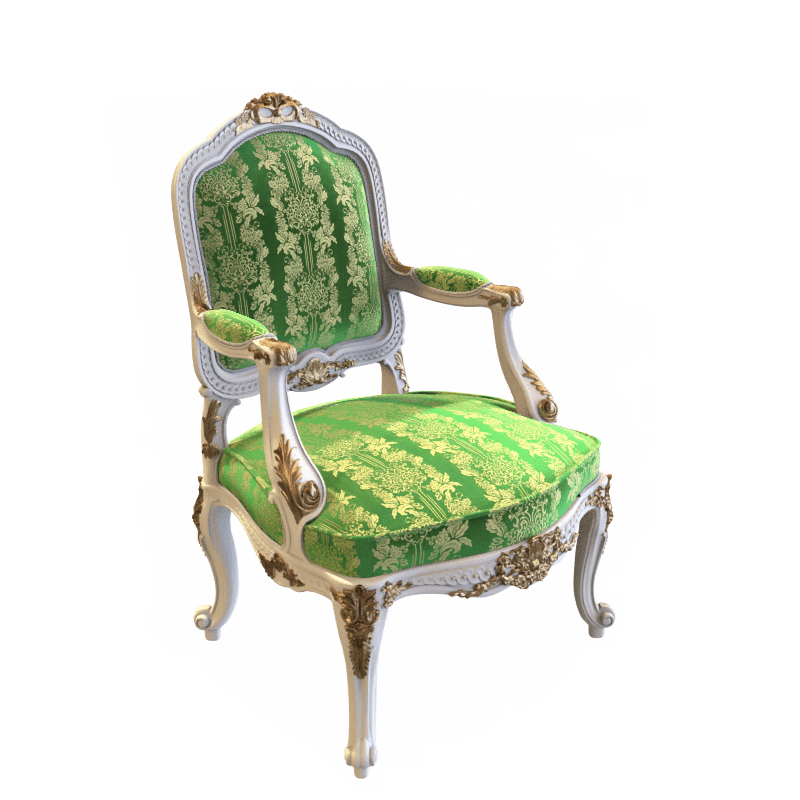}\\
      \midrule
      \includegraphics[width=0.24\textwidth]{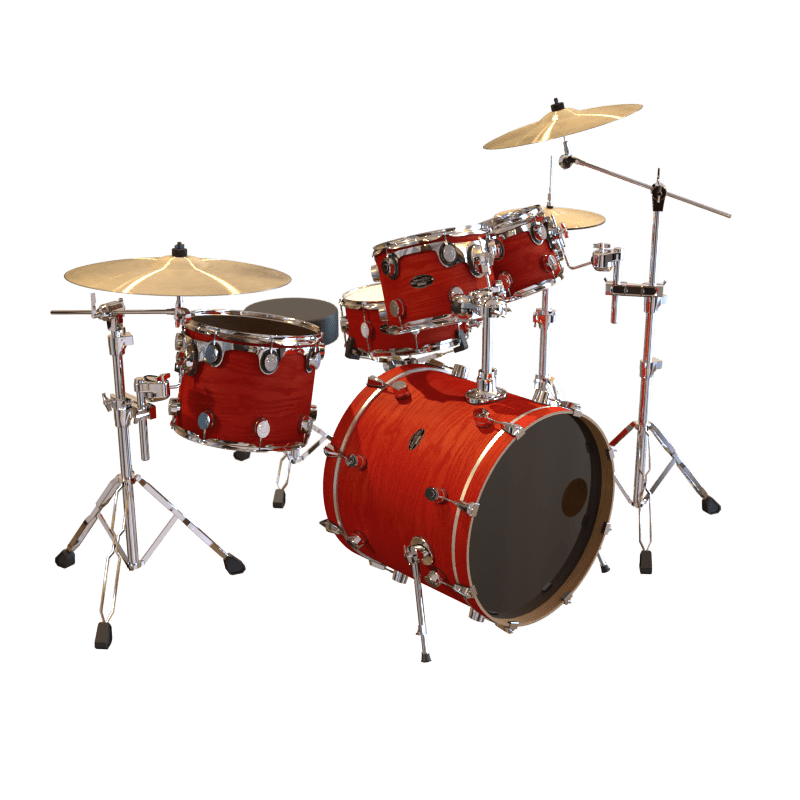}&
      \includegraphics[width=0.24\textwidth]{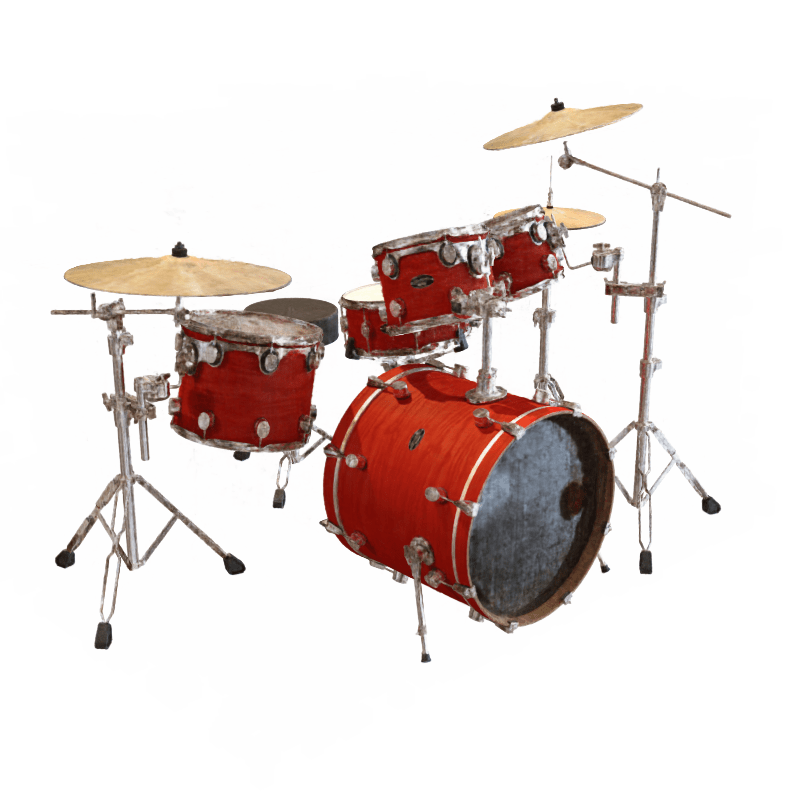}&
      \includegraphics[width=0.24\textwidth]{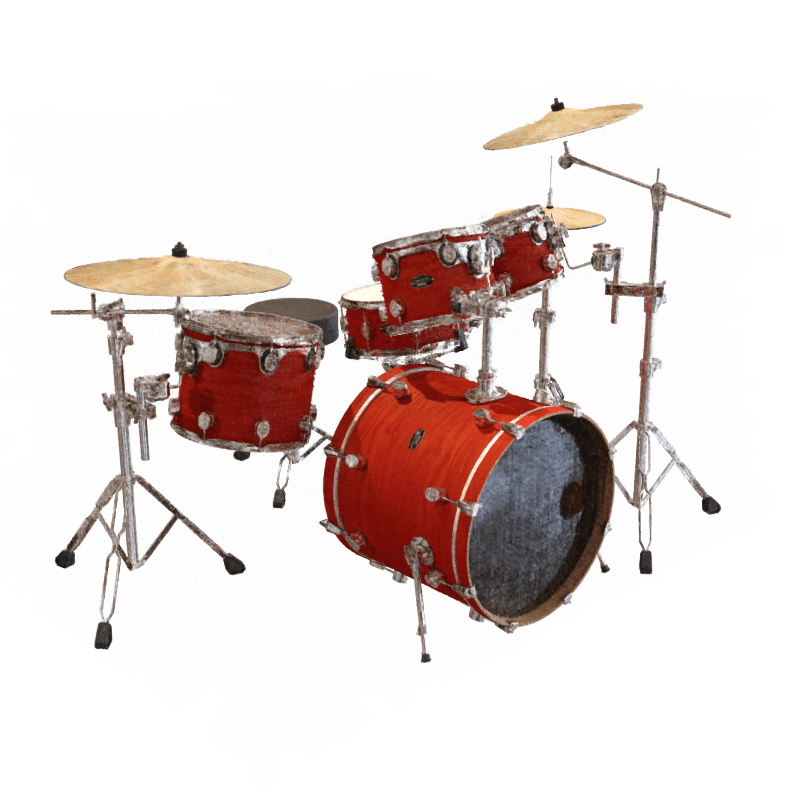}&
      \includegraphics[width=0.24\textwidth]{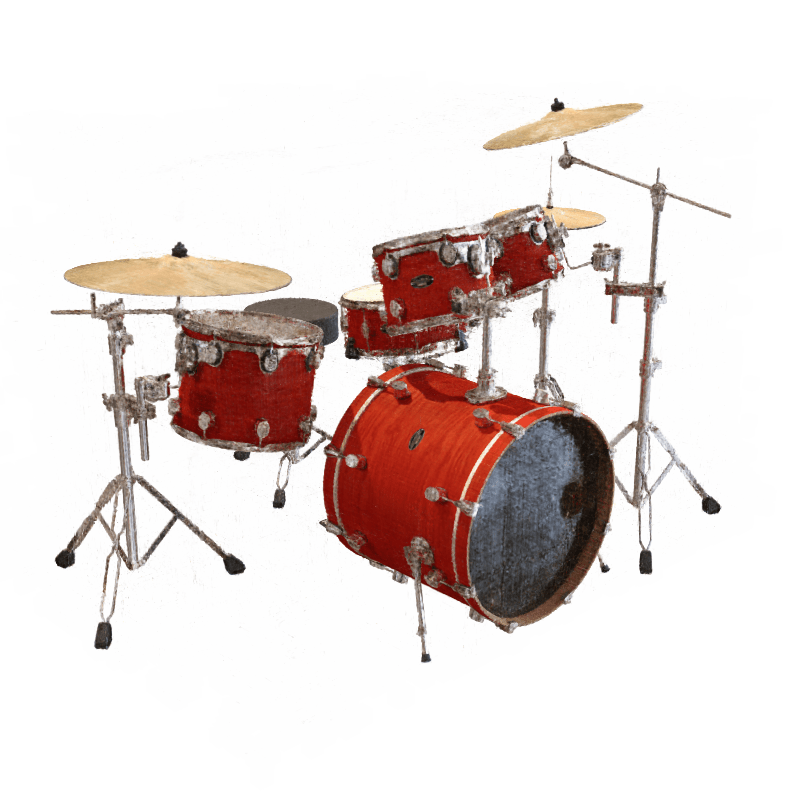}\\
      \midrule 
      \includegraphics[width=0.24\textwidth]{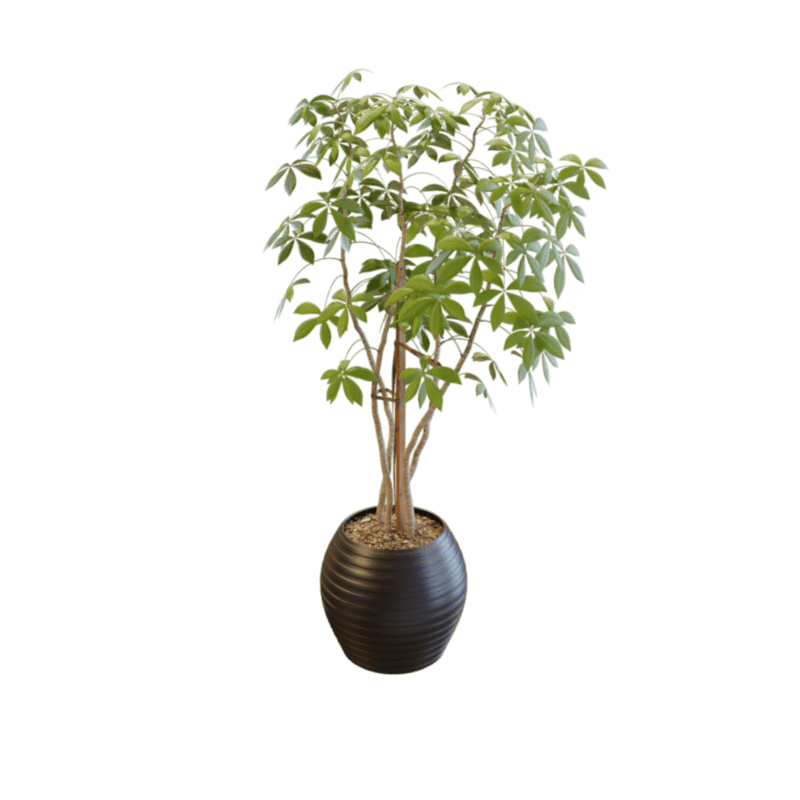}&
      \includegraphics[width=0.24\textwidth]{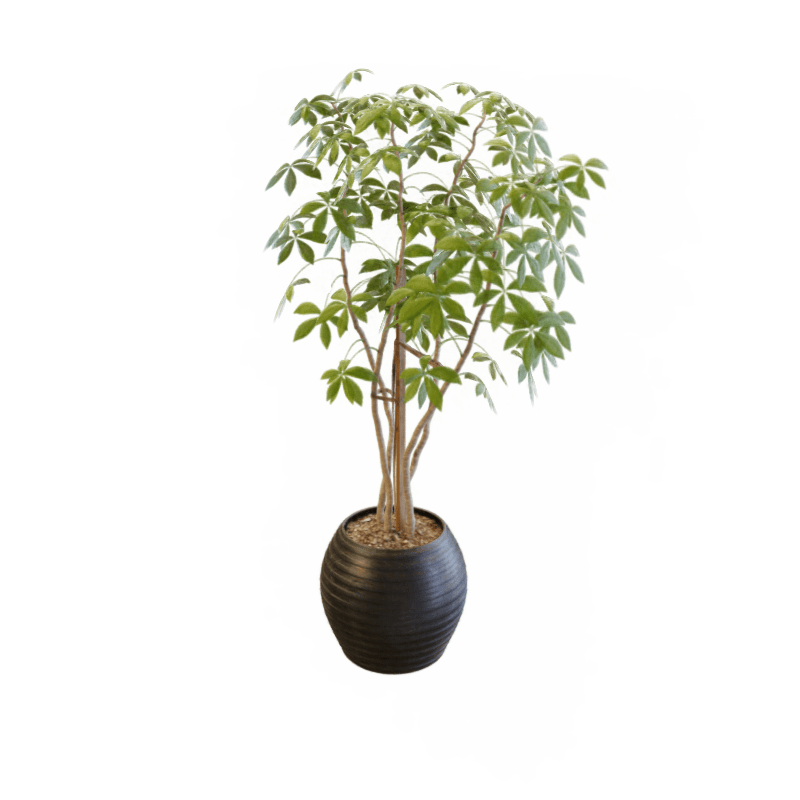}&
      \includegraphics[width=0.24\textwidth]{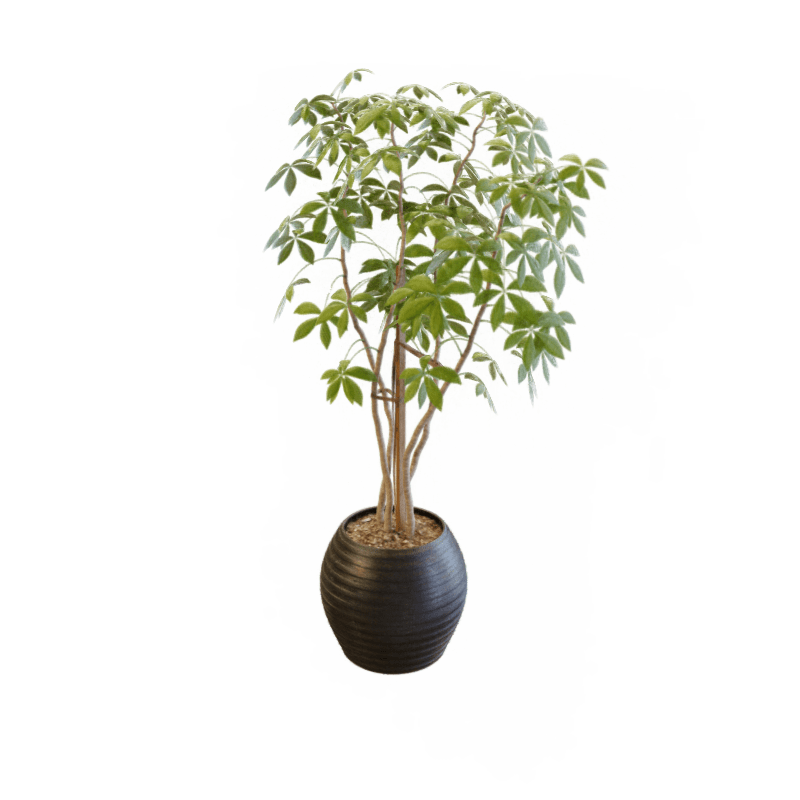}&
      \includegraphics[width=0.24\textwidth]{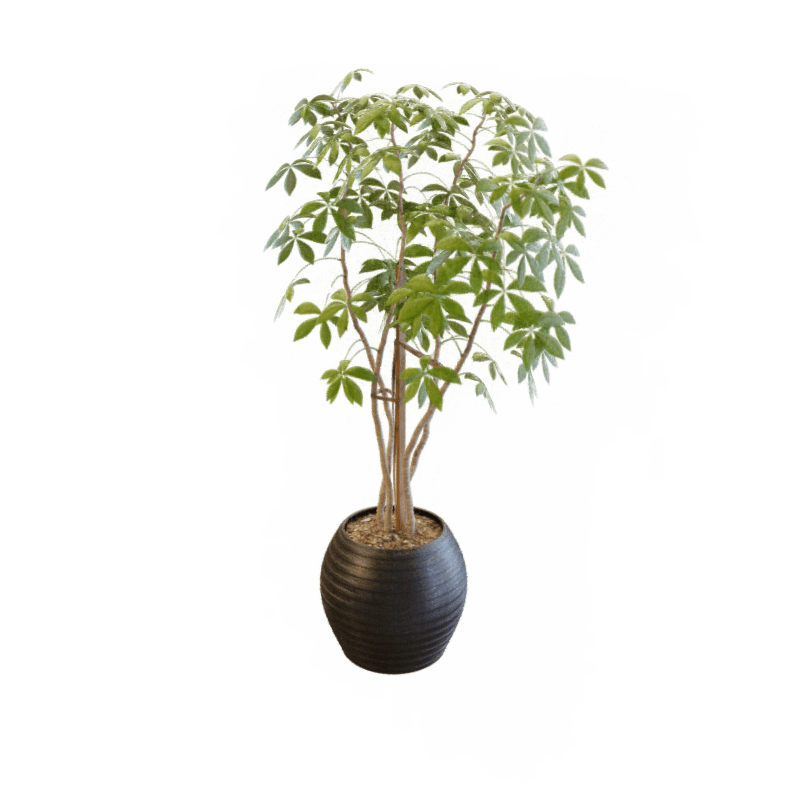}\\
      \midrule
      \includegraphics[width=0.24\textwidth]{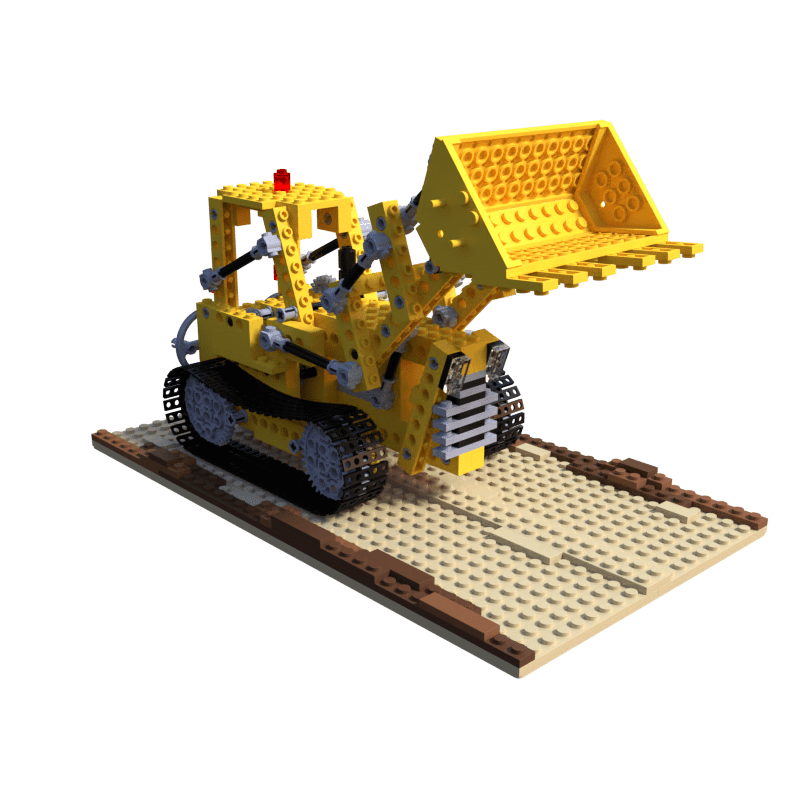}&
      \includegraphics[width=0.24\textwidth]{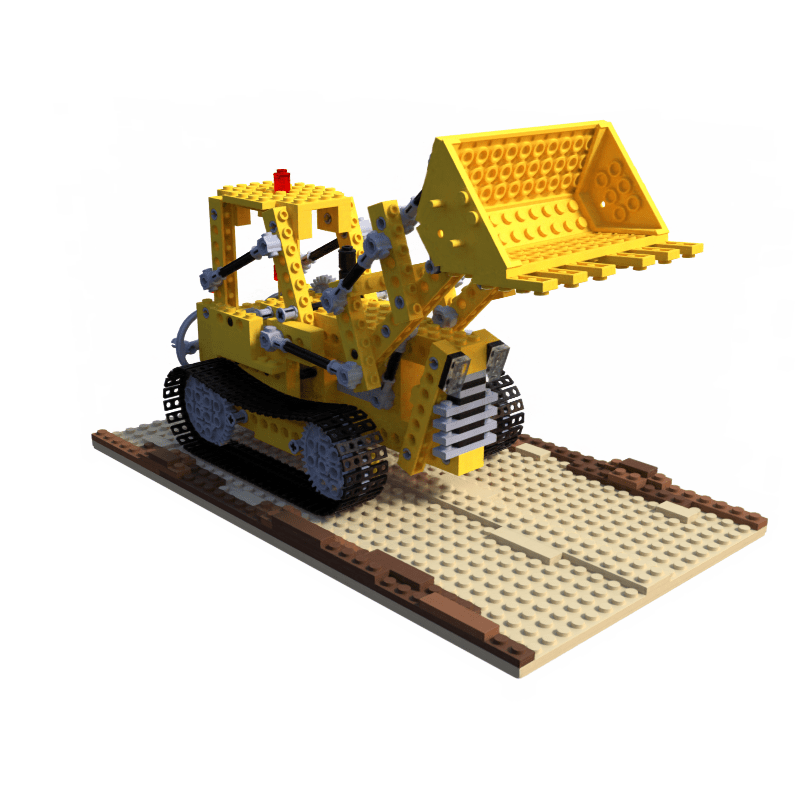}&
      \includegraphics[width=0.24\textwidth]{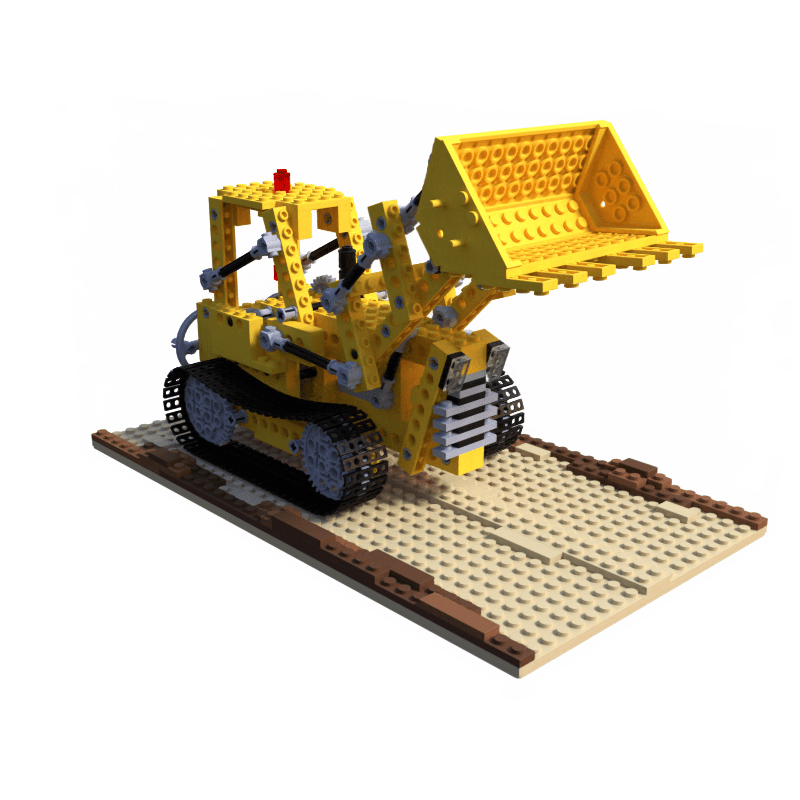}&
      \includegraphics[width=0.24\textwidth]{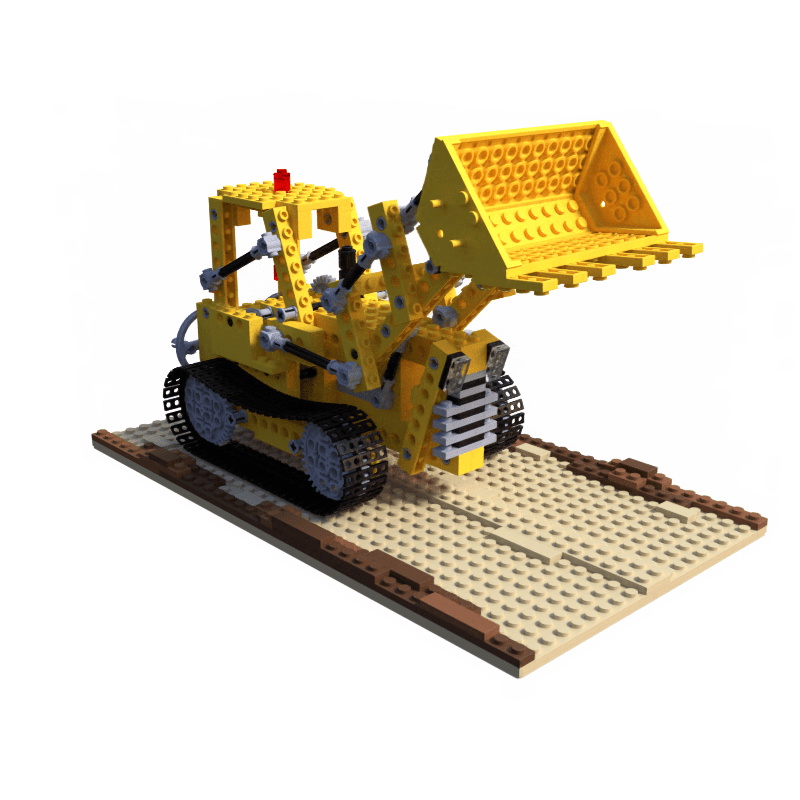}\\
    \bottomrule
     \end{tabular}
     }
 \end{table*}
 \begin{table*}
     \caption{Examples generated from the Synthetic-NSVF dataset with TensoRF.}
     \centering
     \resizebox{\textwidth}{!}{
     \begin{tabular}{c c c c}
     \toprule
     \bf Ground Truth & \bf Baseline & \bf  LOW compression & \bf HIGH compression\\
     \midrule
      \includegraphics[width=0.24\textwidth]{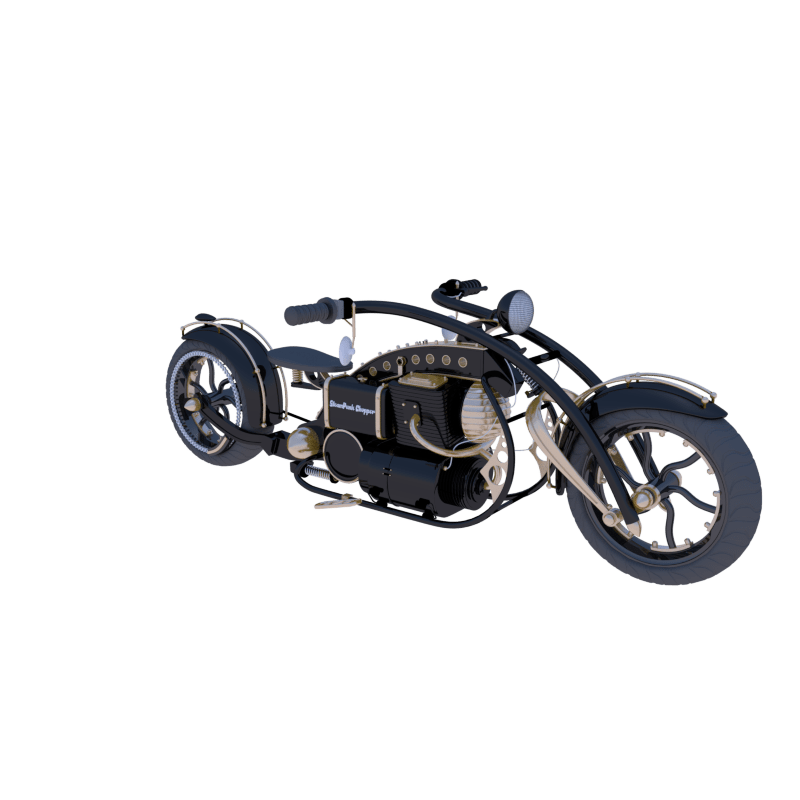}&
      \includegraphics[width=0.24\textwidth]{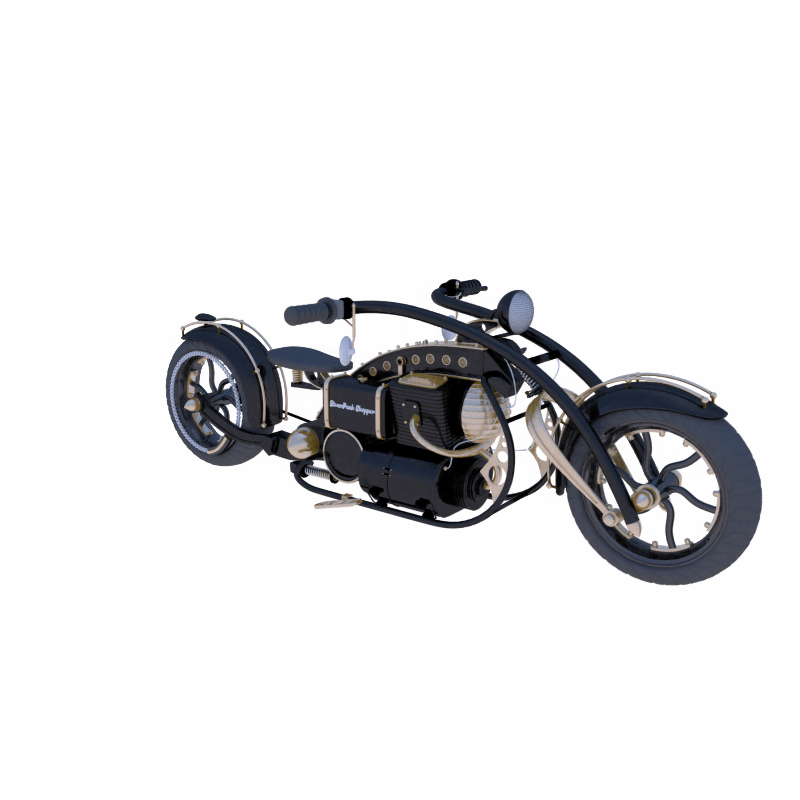} & 
      \includegraphics[width=0.24\textwidth]{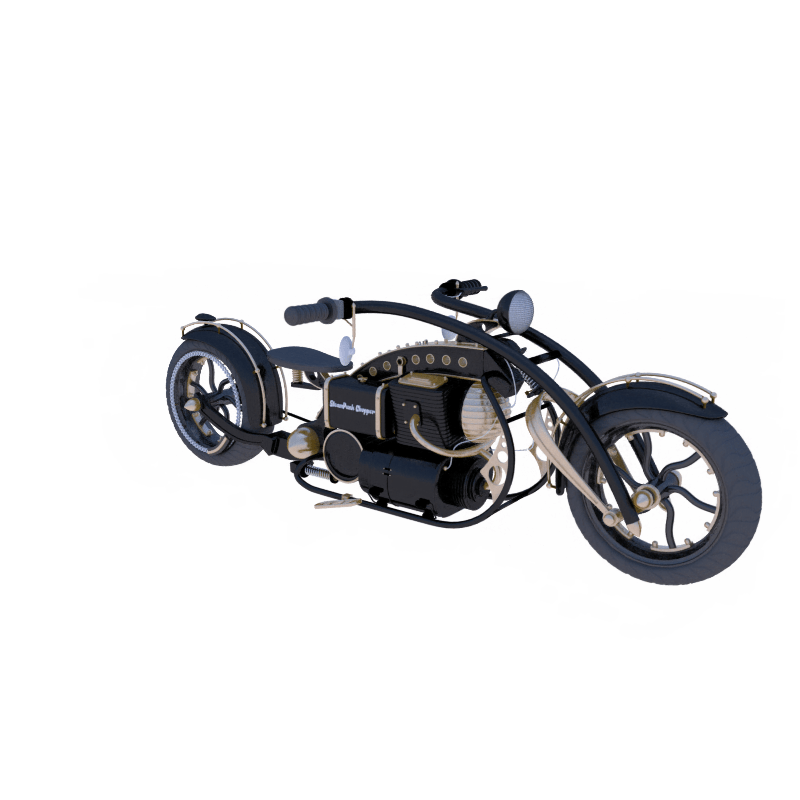}&
      \includegraphics[width=0.24\textwidth]{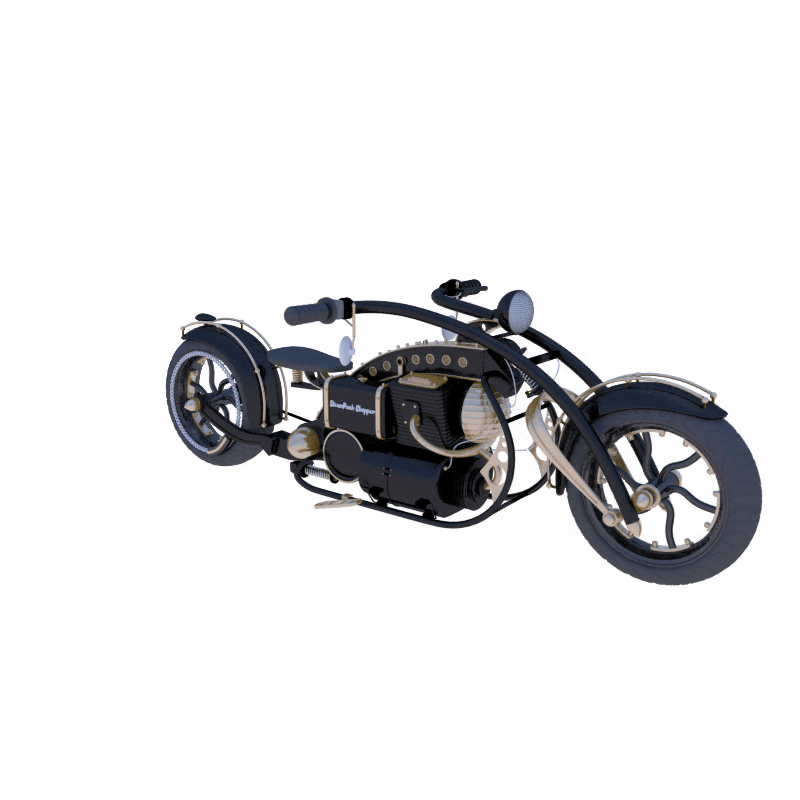}\\
    \midrule
      \includegraphics[width=0.24\textwidth]{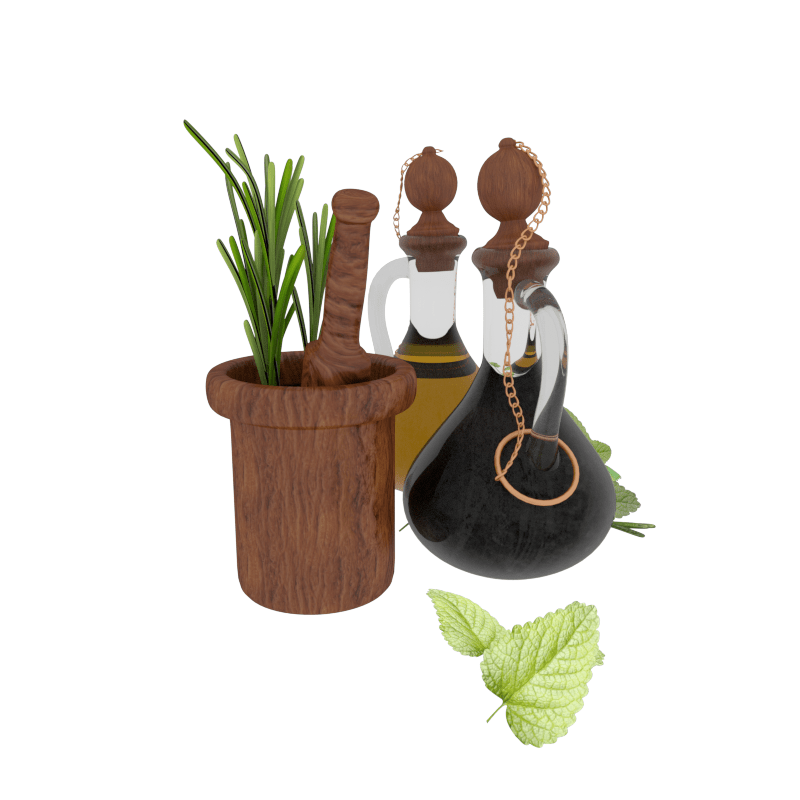}&
      \includegraphics[width=0.24\textwidth]{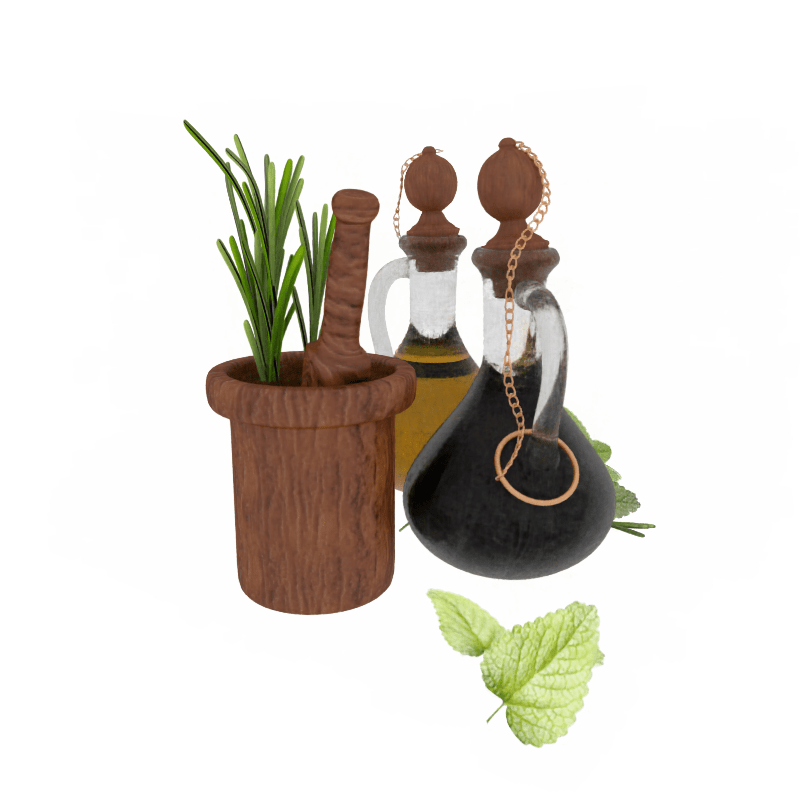} & 
      \includegraphics[width=0.24\textwidth]{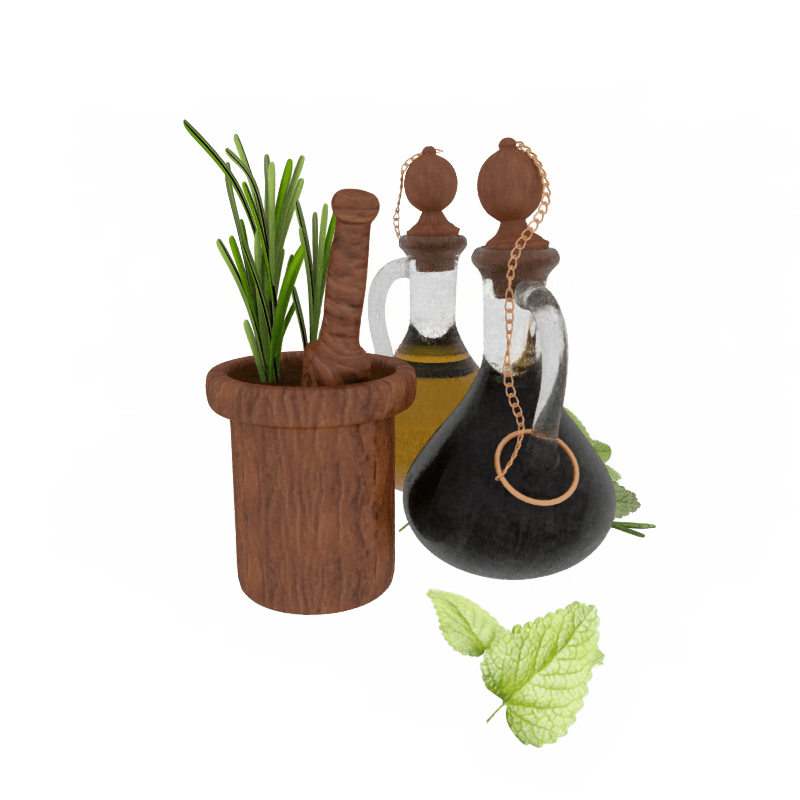}&
      \includegraphics[width=0.24\textwidth]{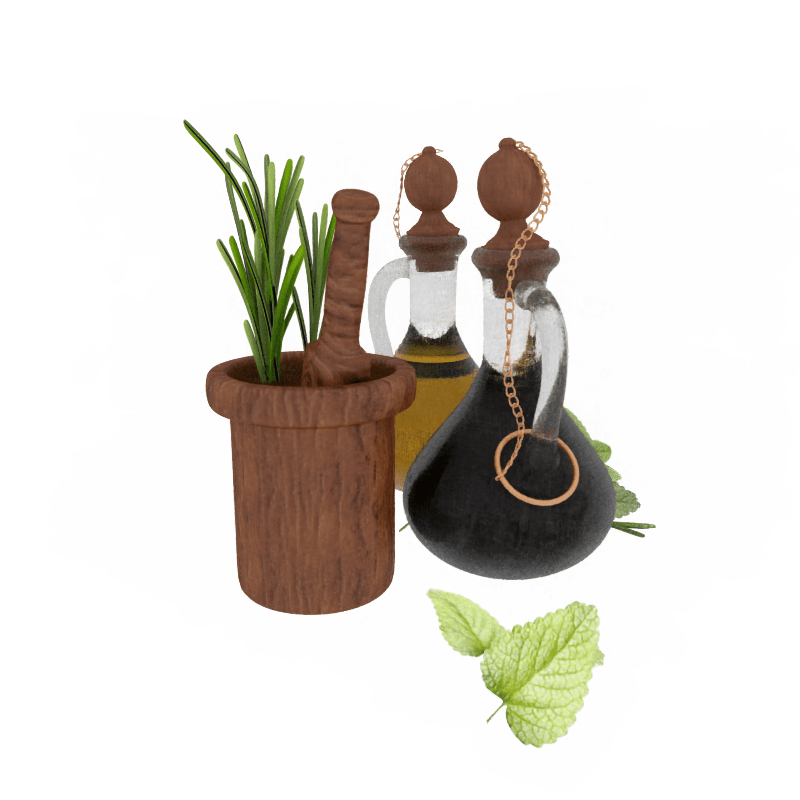}\\
\midrule
      \includegraphics[width=0.24\textwidth]{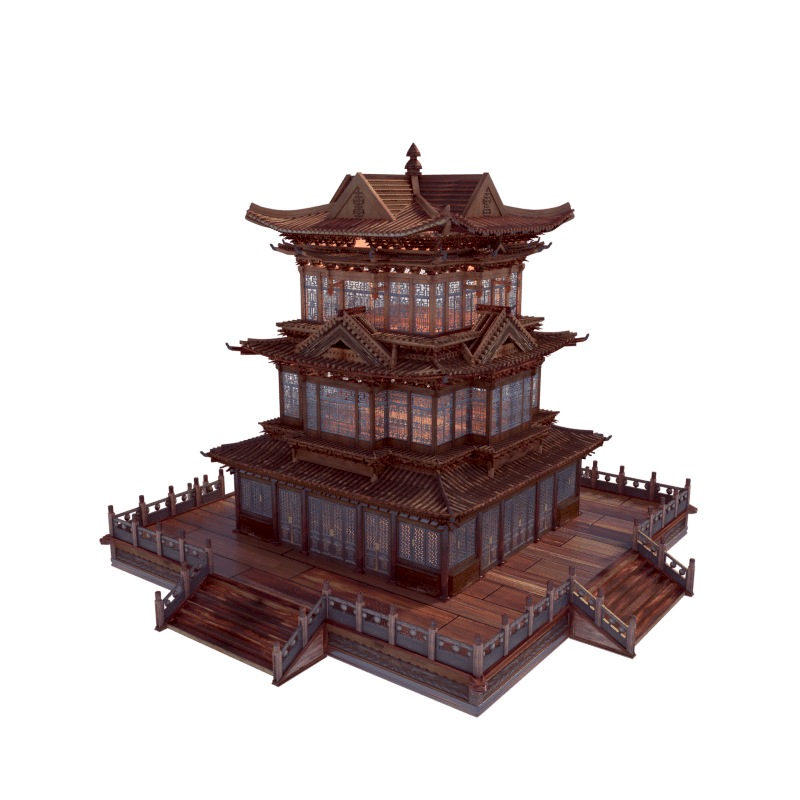}&
      \includegraphics[width=0.24\textwidth]{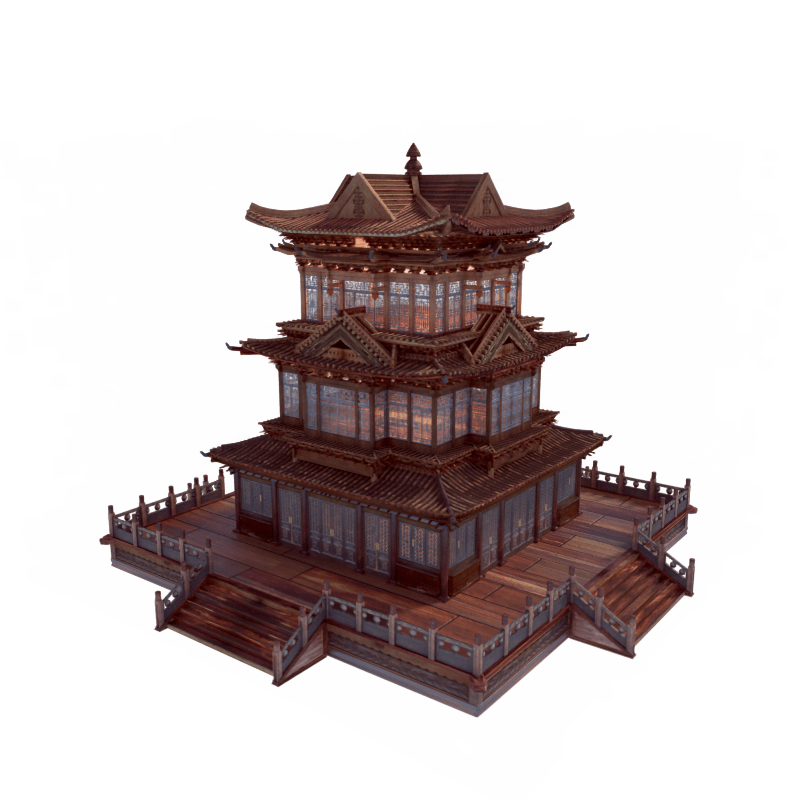} & 
      \includegraphics[width=0.24\textwidth]{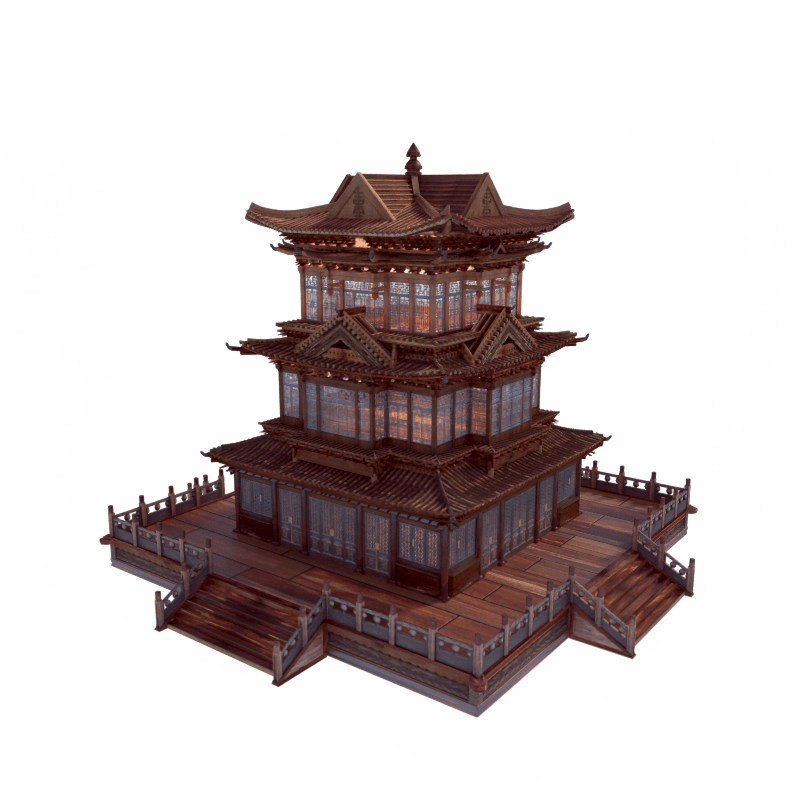}&
      \includegraphics[width=0.24\textwidth]{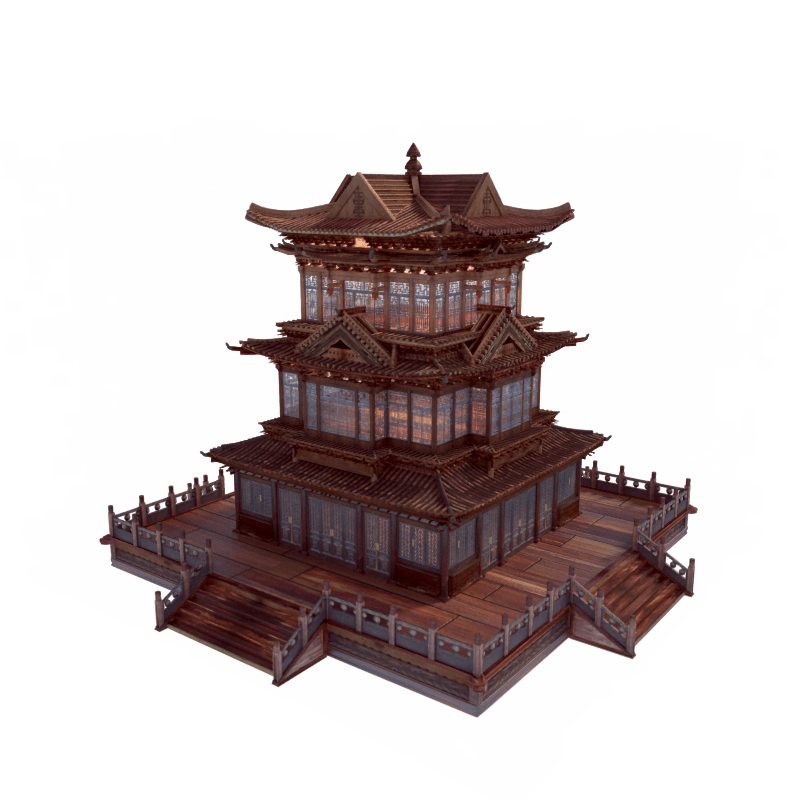}\\
      \midrule
      \includegraphics[width=0.24\textwidth]{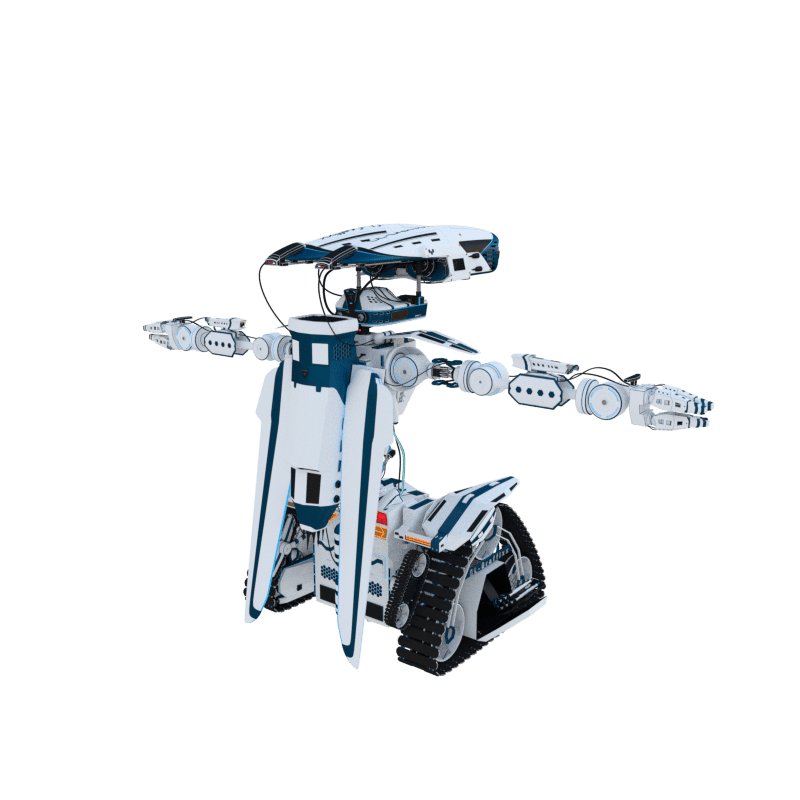}&
      \includegraphics[width=0.24\textwidth]{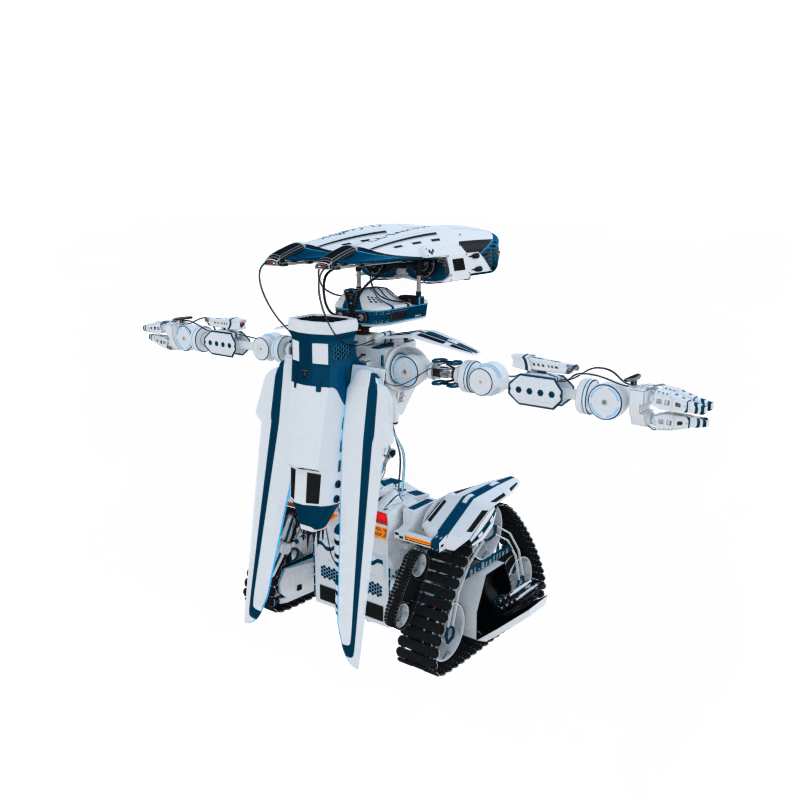} & 
      \includegraphics[width=0.24\textwidth]{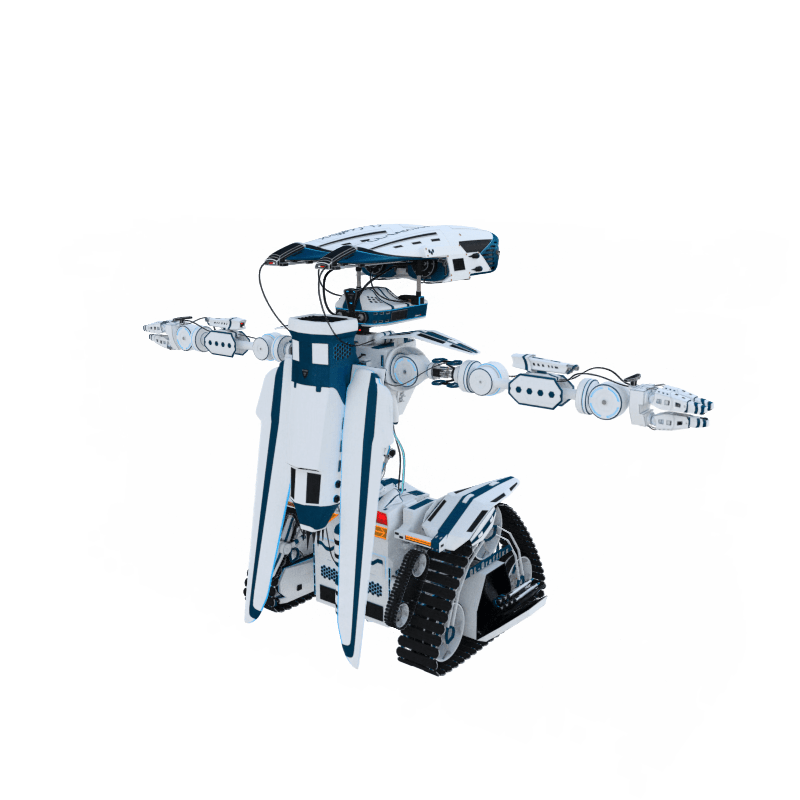}&
      \includegraphics[width=0.24\textwidth]{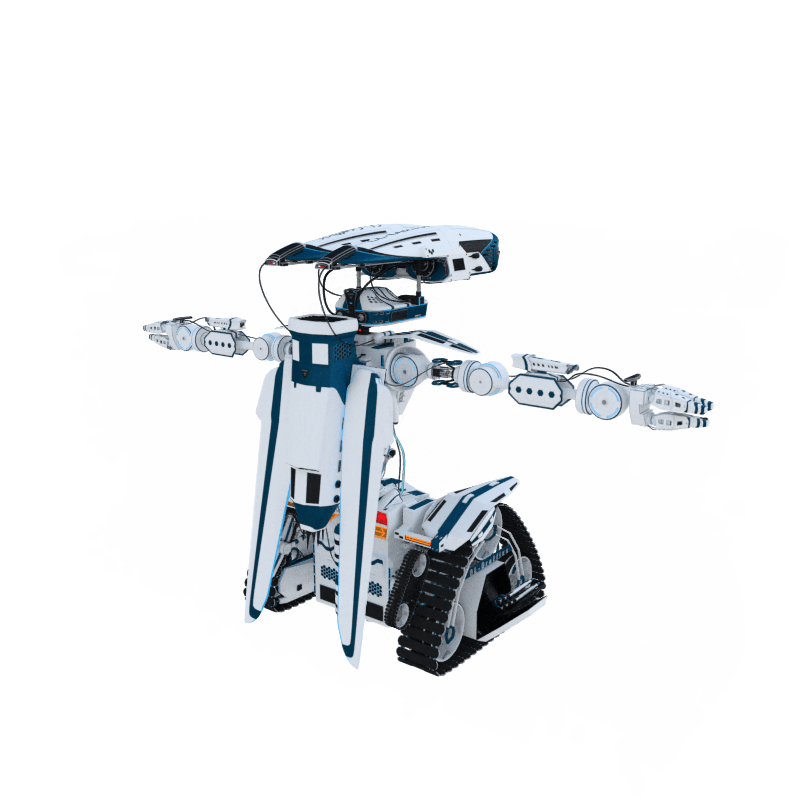}\\
\midrule
      \includegraphics[width=0.24\textwidth]{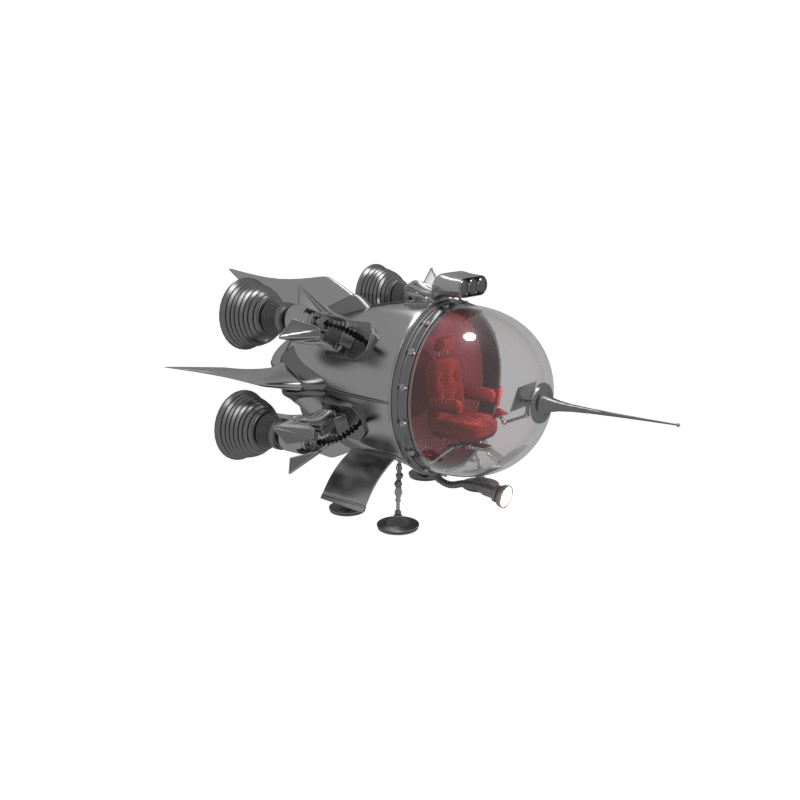}&
      \includegraphics[width=0.24\textwidth]{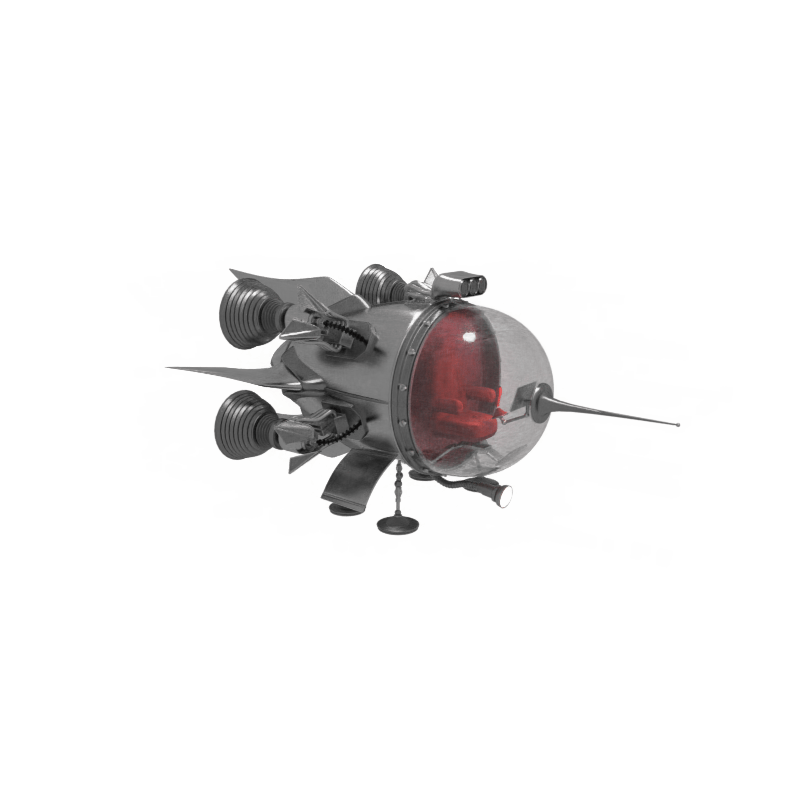} & 
      \includegraphics[width=0.24\textwidth]{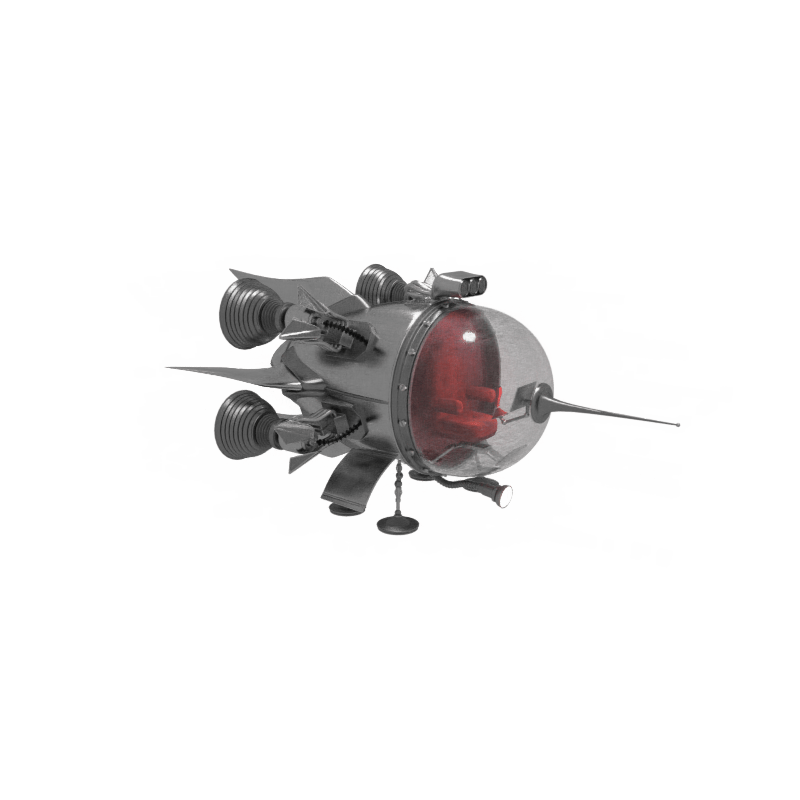}&
      \includegraphics[width=0.24\textwidth]{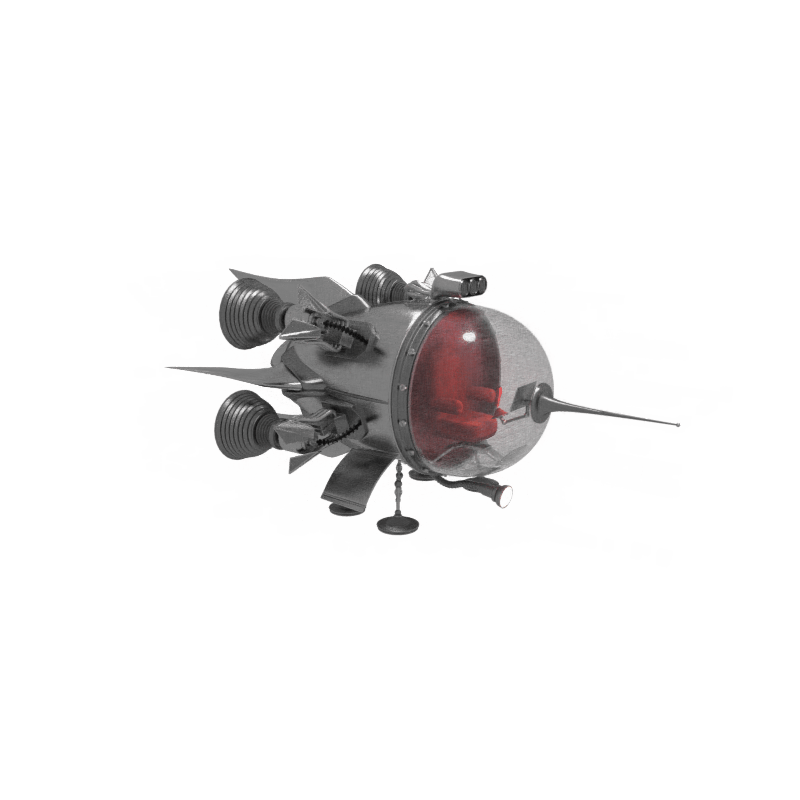}\\
\bottomrule
     \end{tabular}
     }
 \end{table*}
 \begin{table*}
     \caption{Examples generated from the Tanks\&Temples dataset with TensoRF.}
     \centering
     \resizebox{\textwidth}{!}{
     \begin{tabular}{c c c c}
     \toprule
     \bf Ground Truth &\bf  Baseline &\bf LOW compression &\bf HIGH compression\\
     \midrule
               \includegraphics[width=0.24\textwidth]{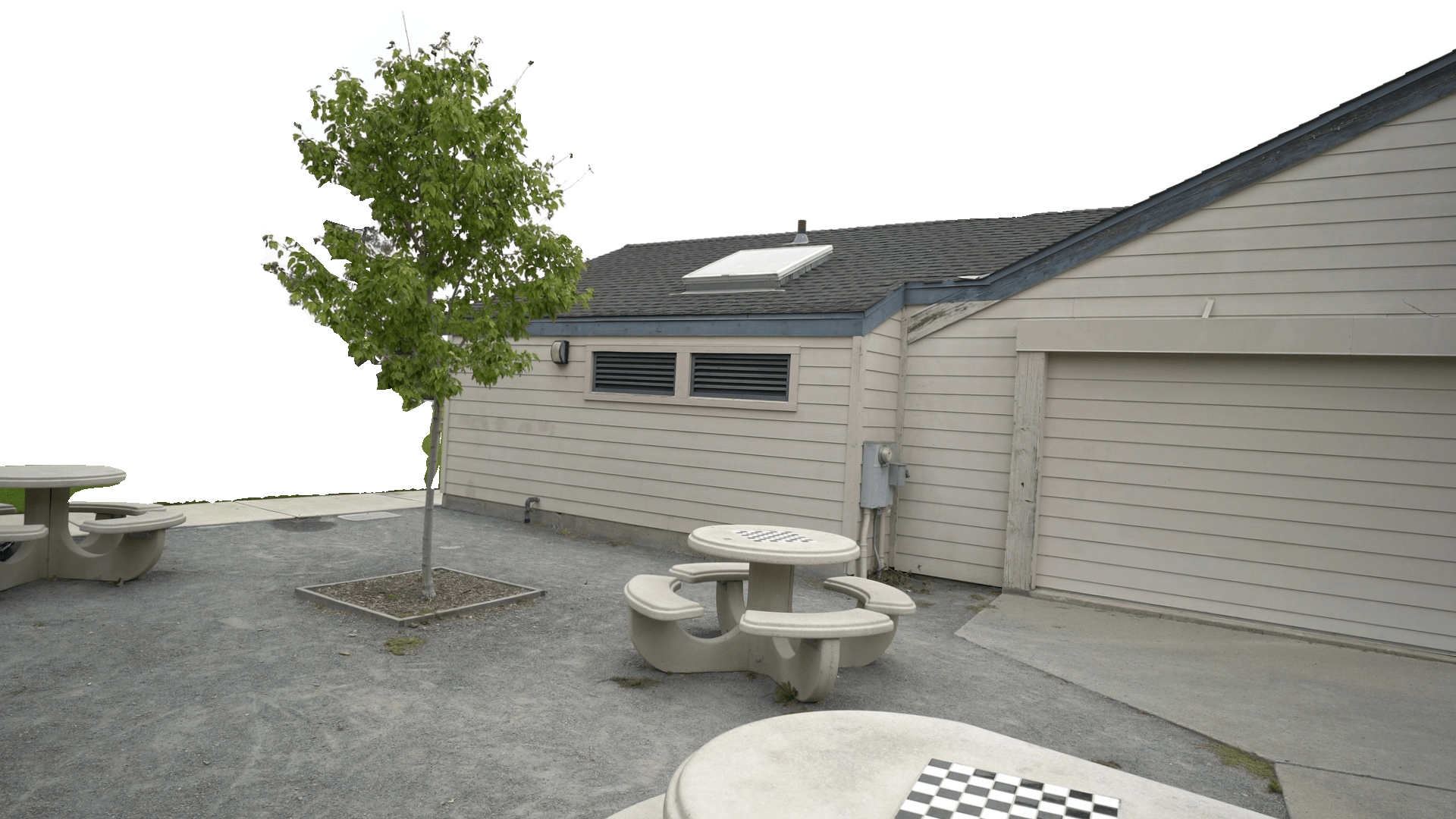}&
      \includegraphics[width=0.24\textwidth]{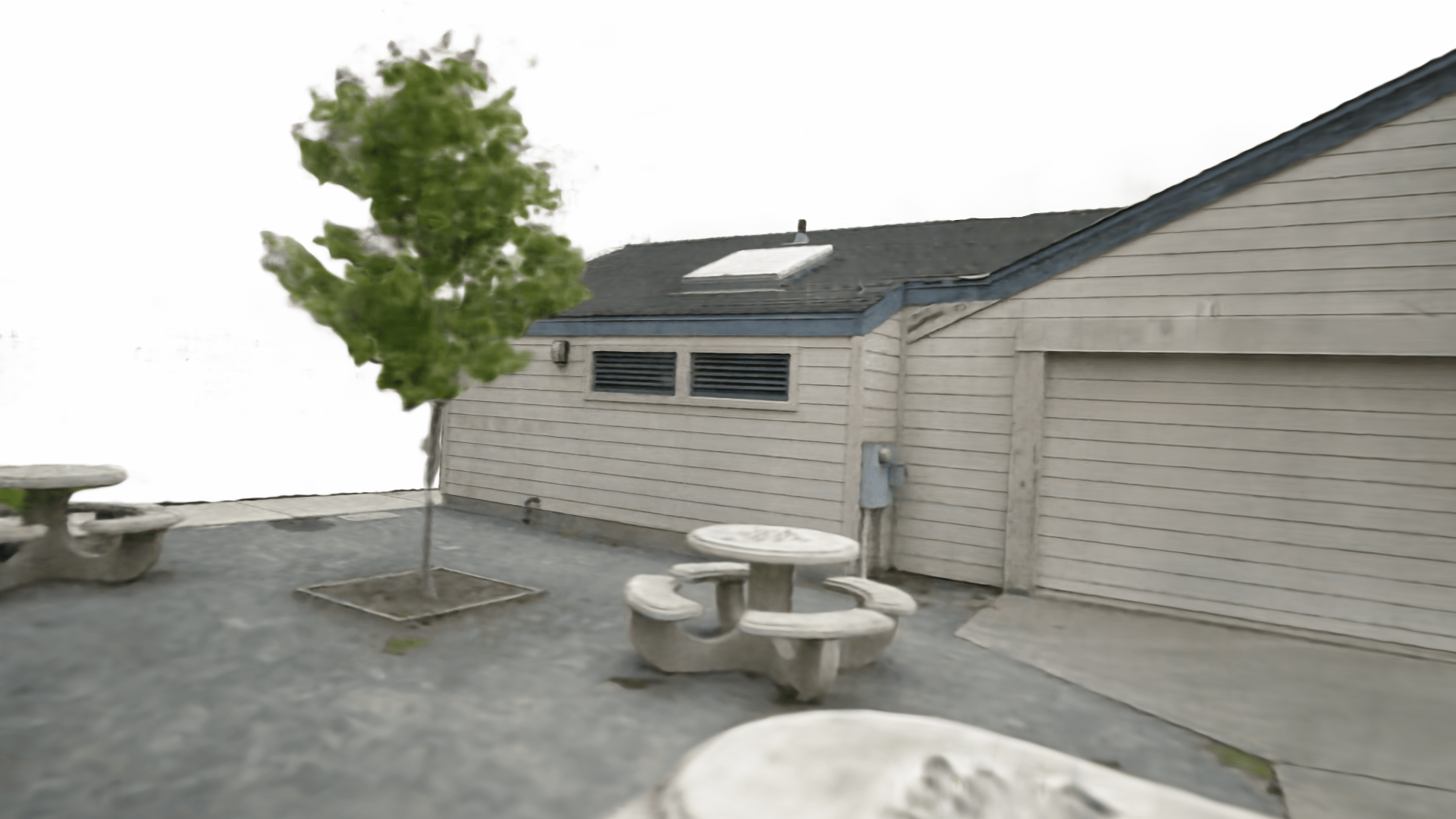} & 
      \includegraphics[width=0.24\textwidth]{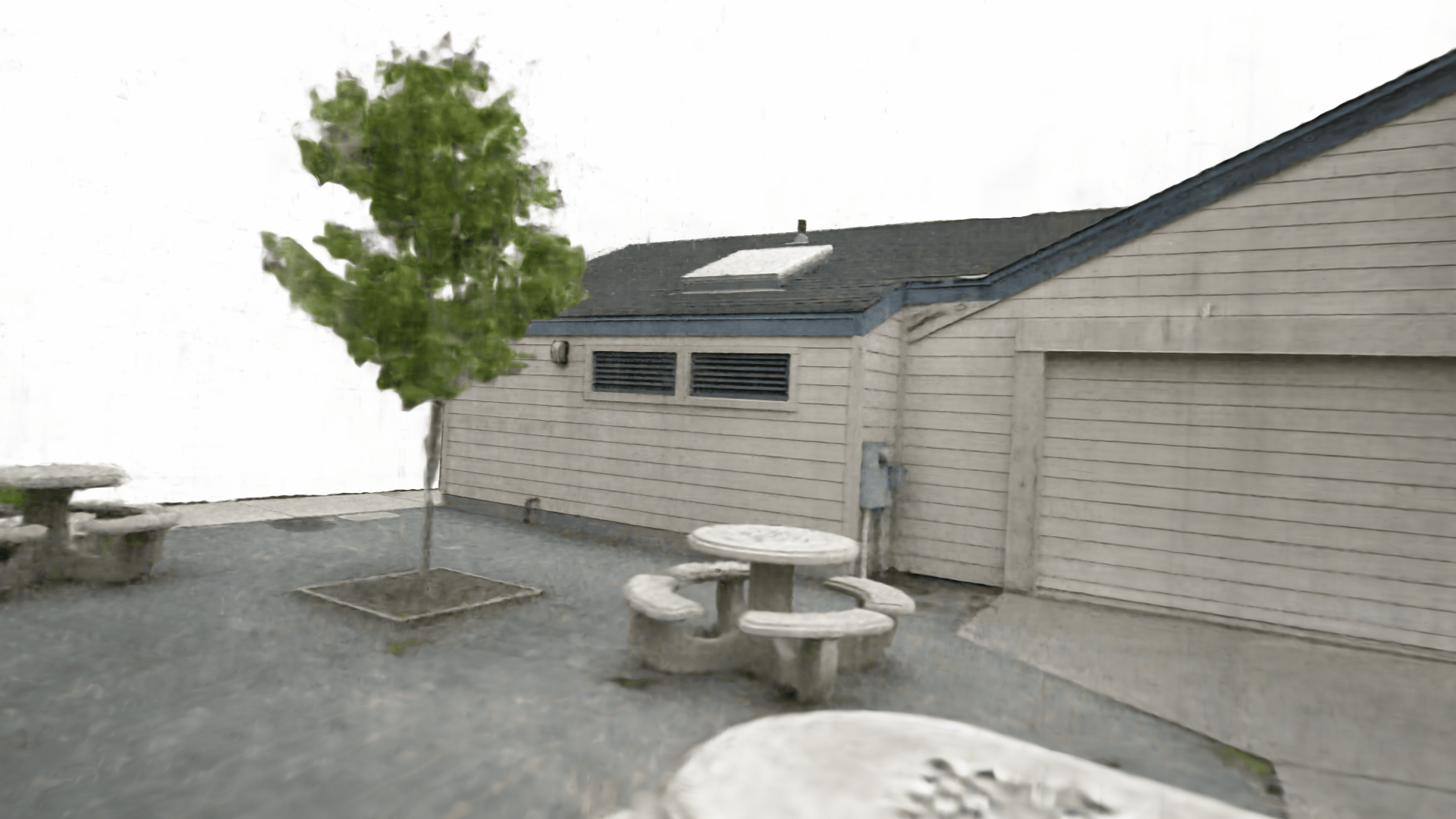} & 
      \includegraphics[width=0.24\textwidth]{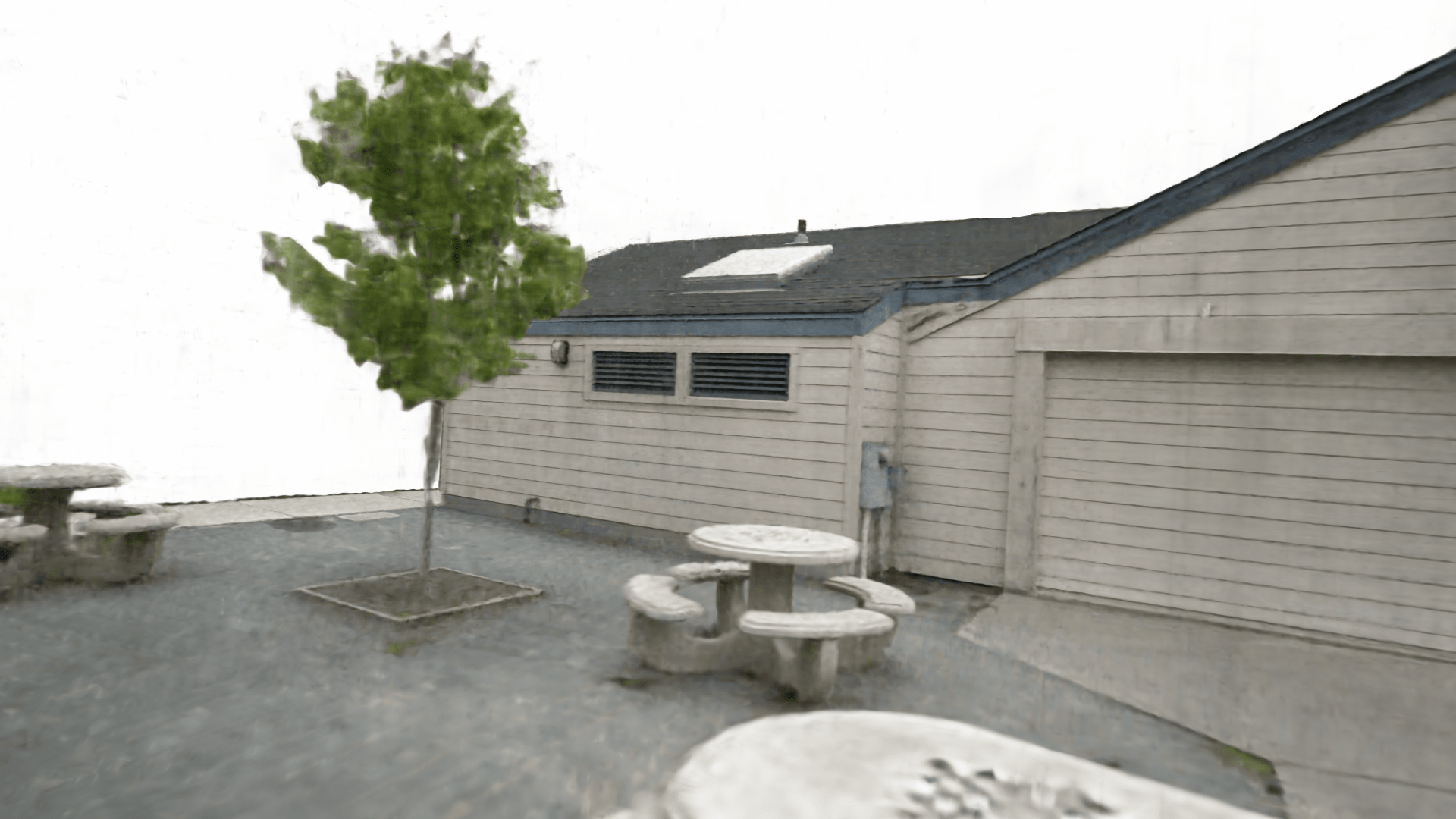} \\
     \midrule
      \includegraphics[width=0.24\textwidth]{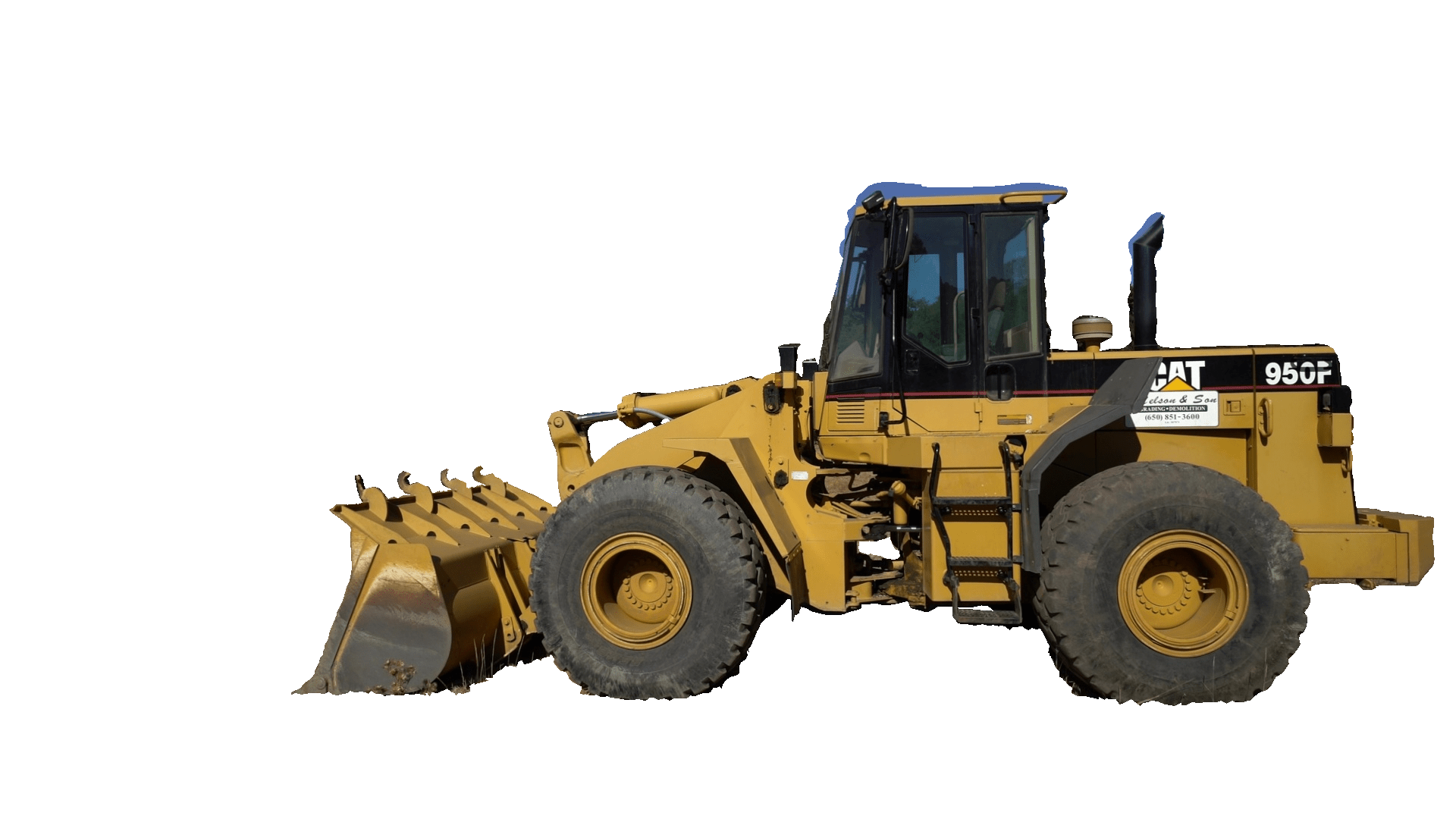}&
      \includegraphics[width=0.24\textwidth]{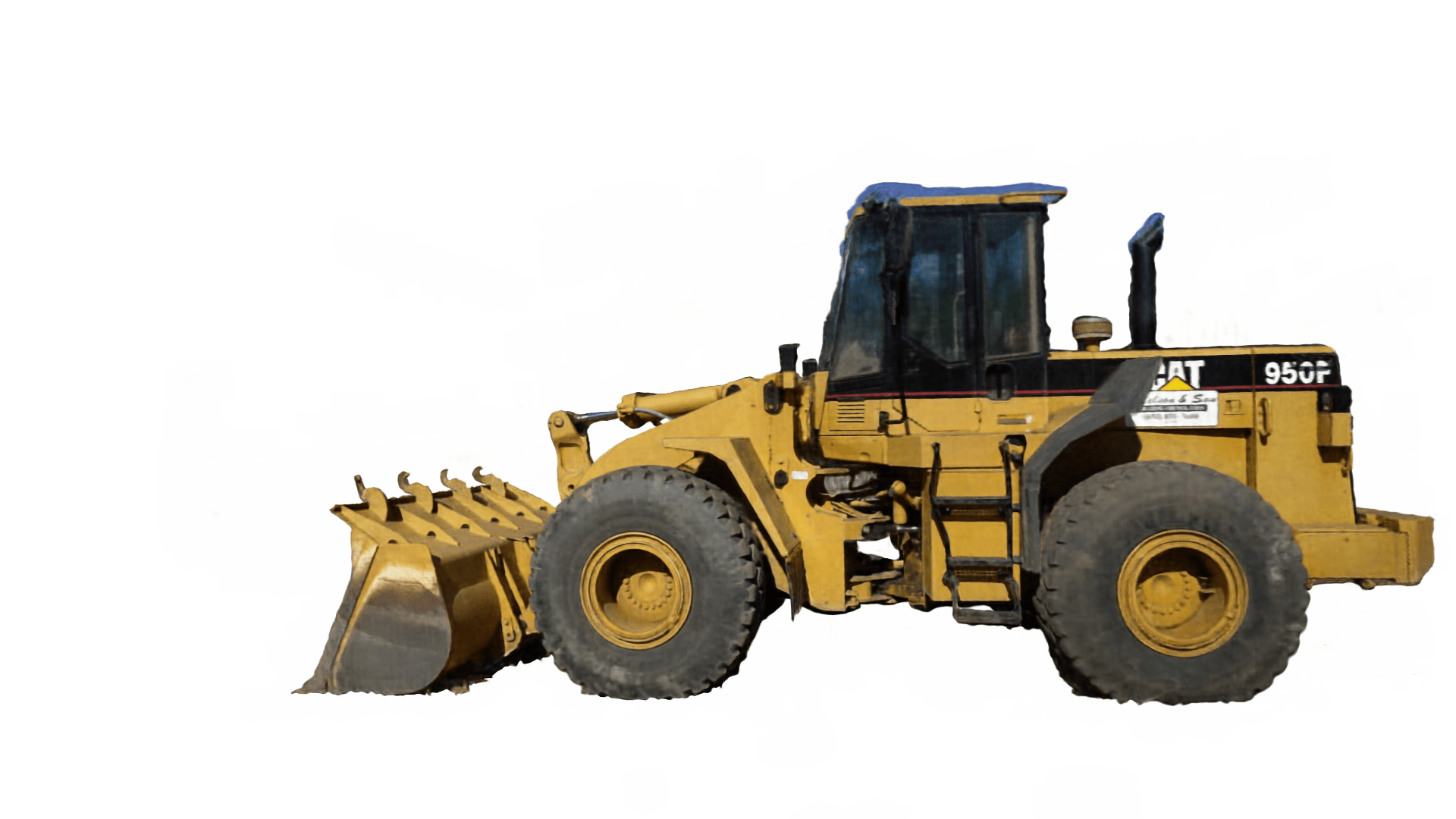} & 
      \includegraphics[width=0.24\textwidth]{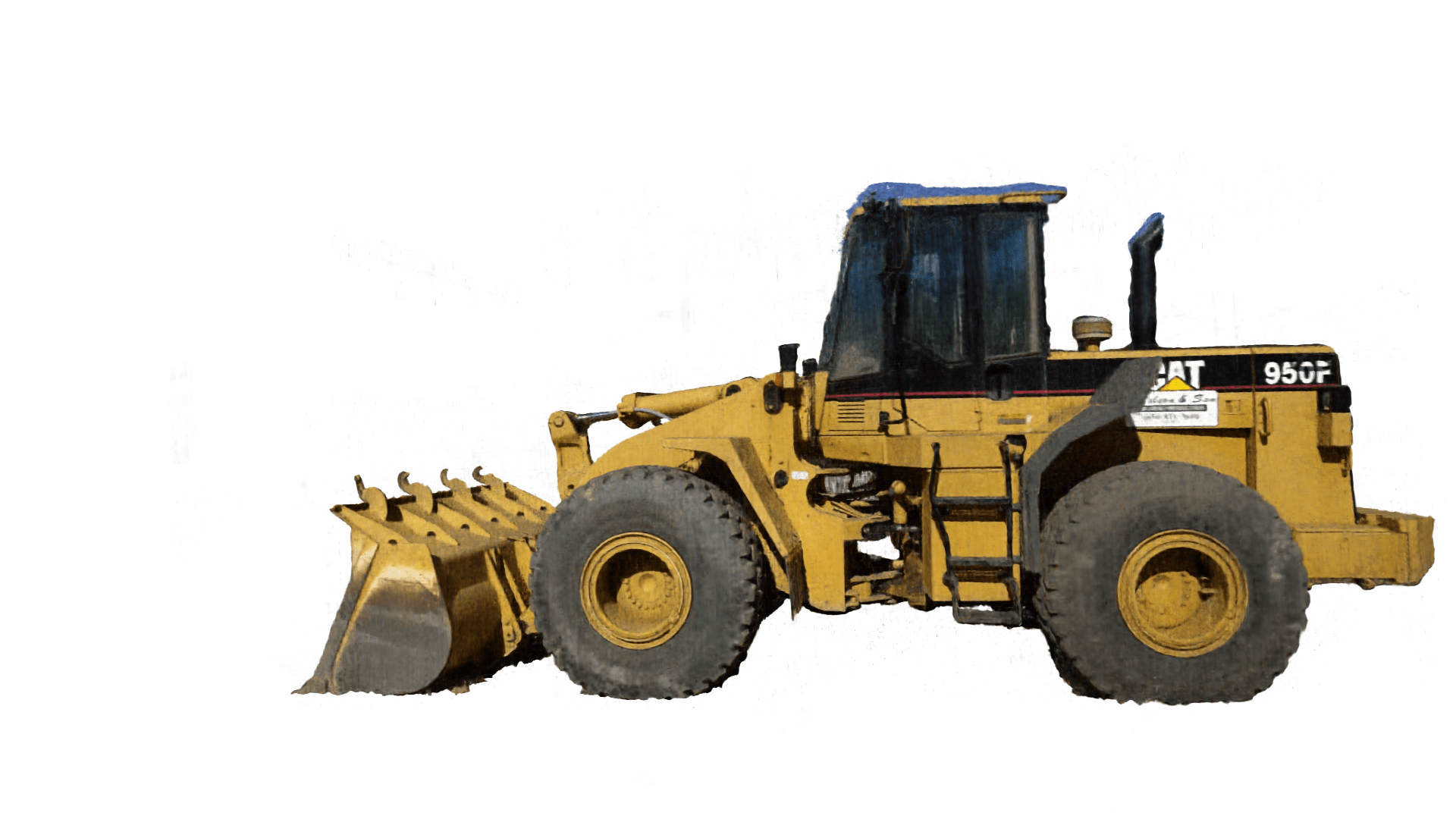} & 
      \includegraphics[width=0.24\textwidth]{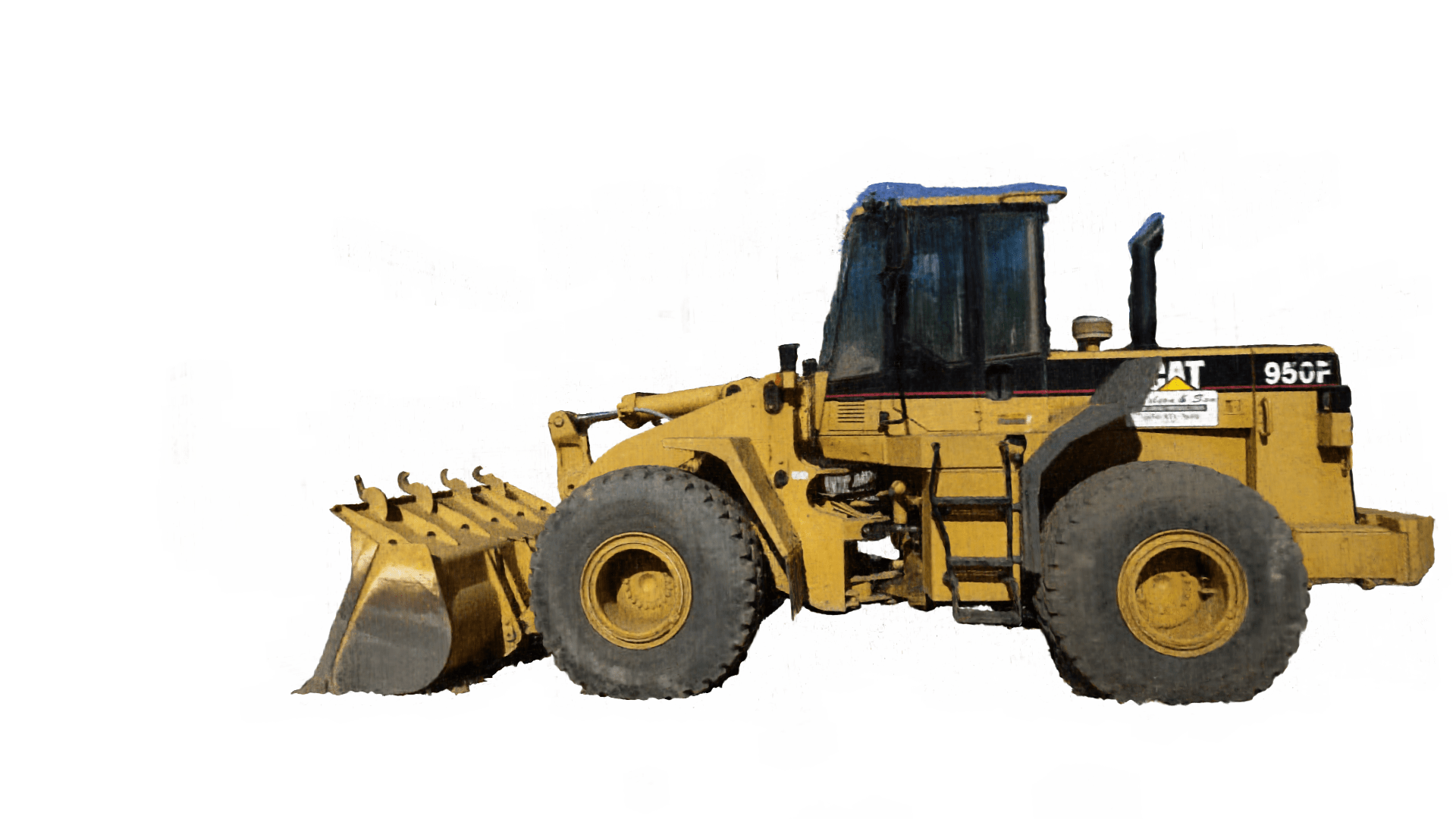} \\
\midrule
           \includegraphics[width=0.24\textwidth]{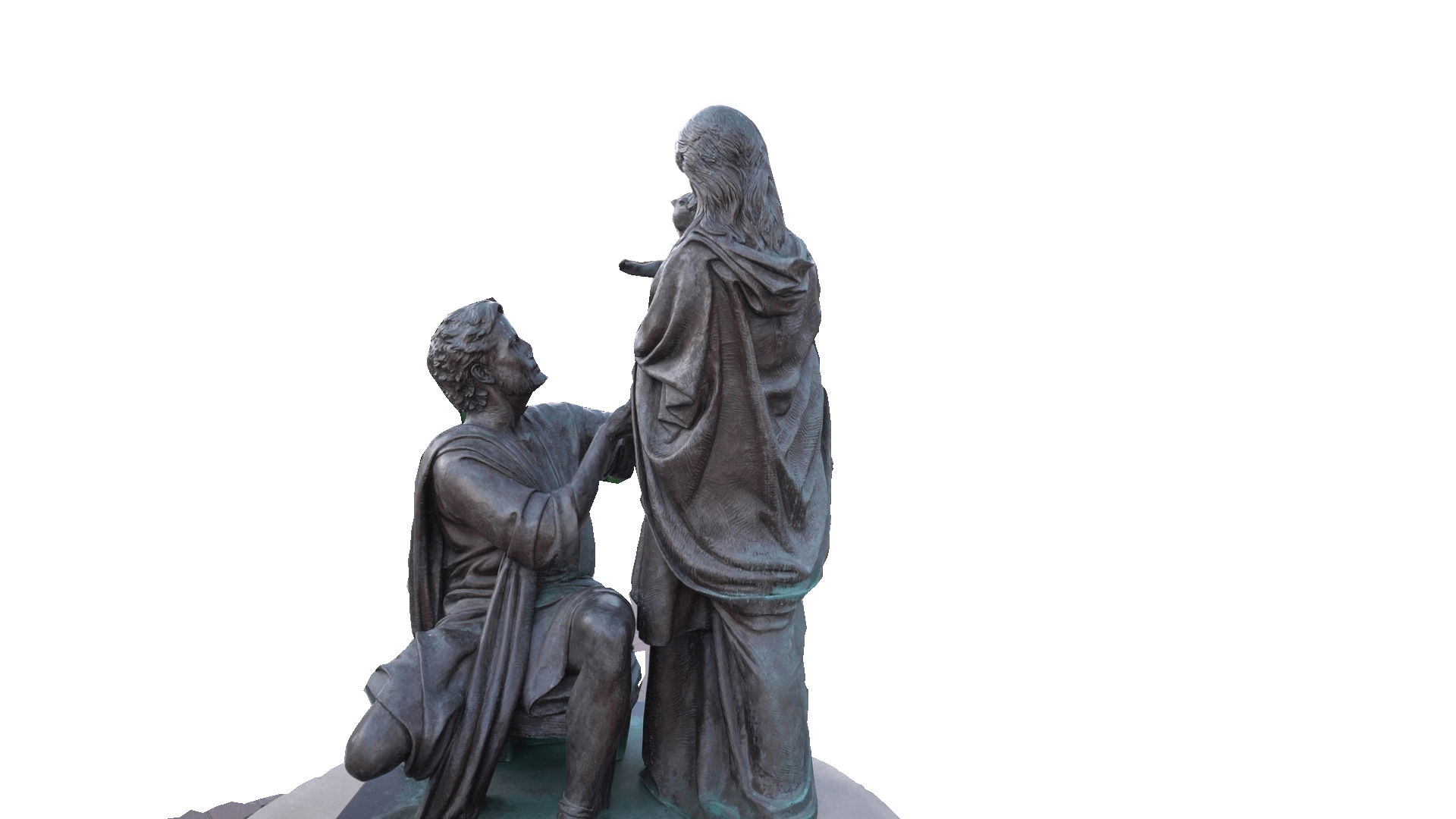}&
      \includegraphics[width=0.24\textwidth]{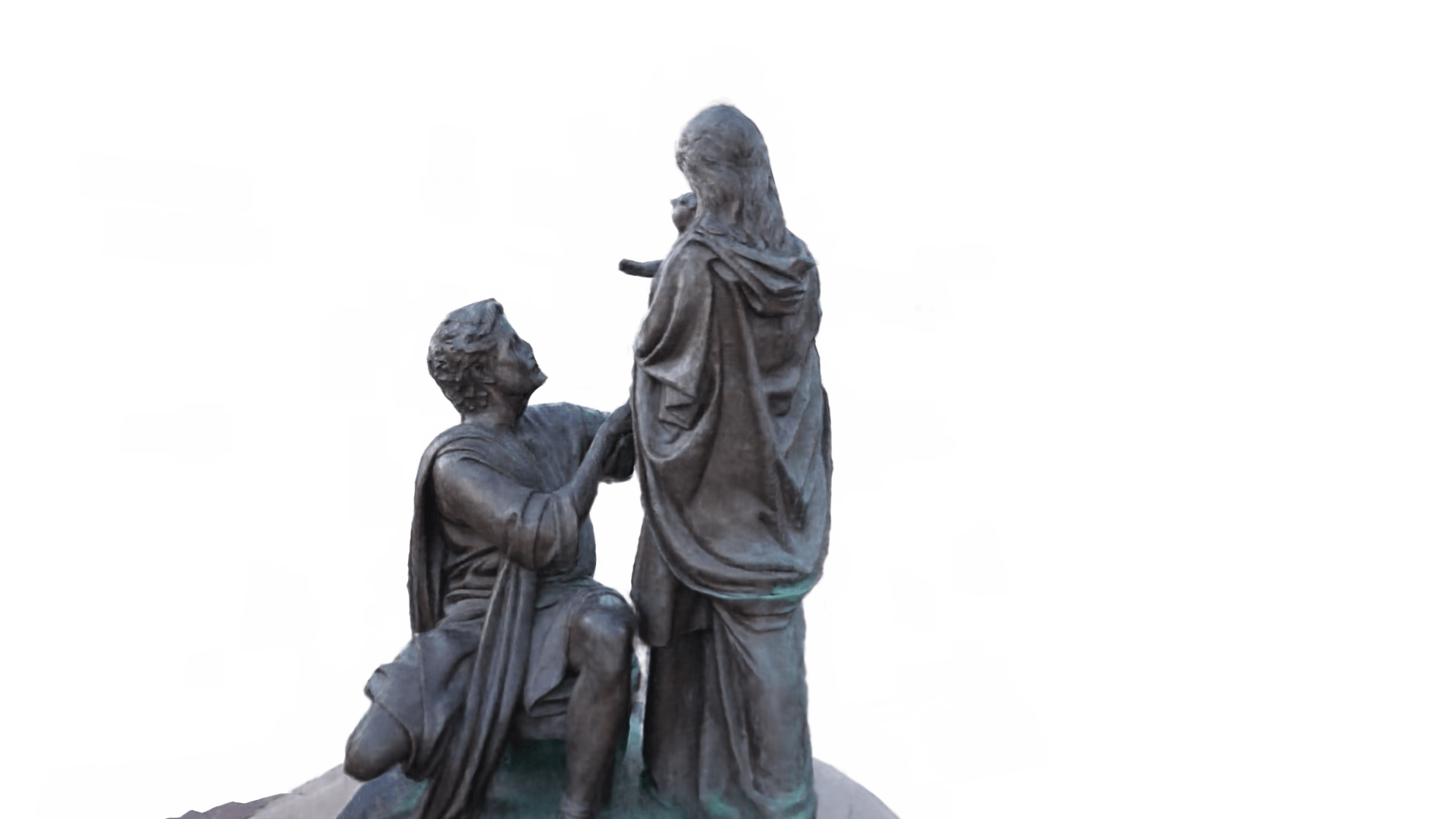} & 
      \includegraphics[width=0.24\textwidth]{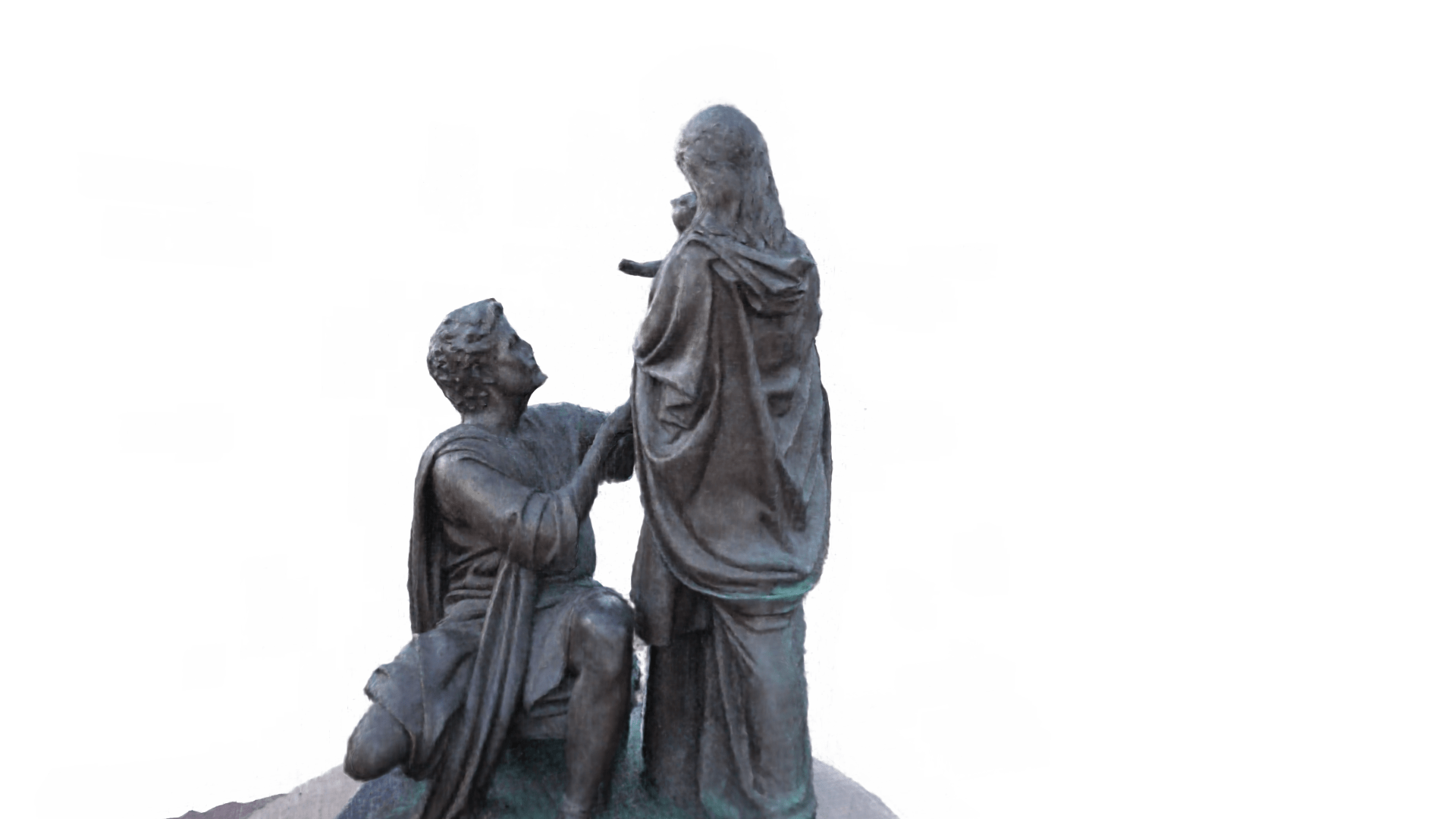} & 
      \includegraphics[width=0.24\textwidth]{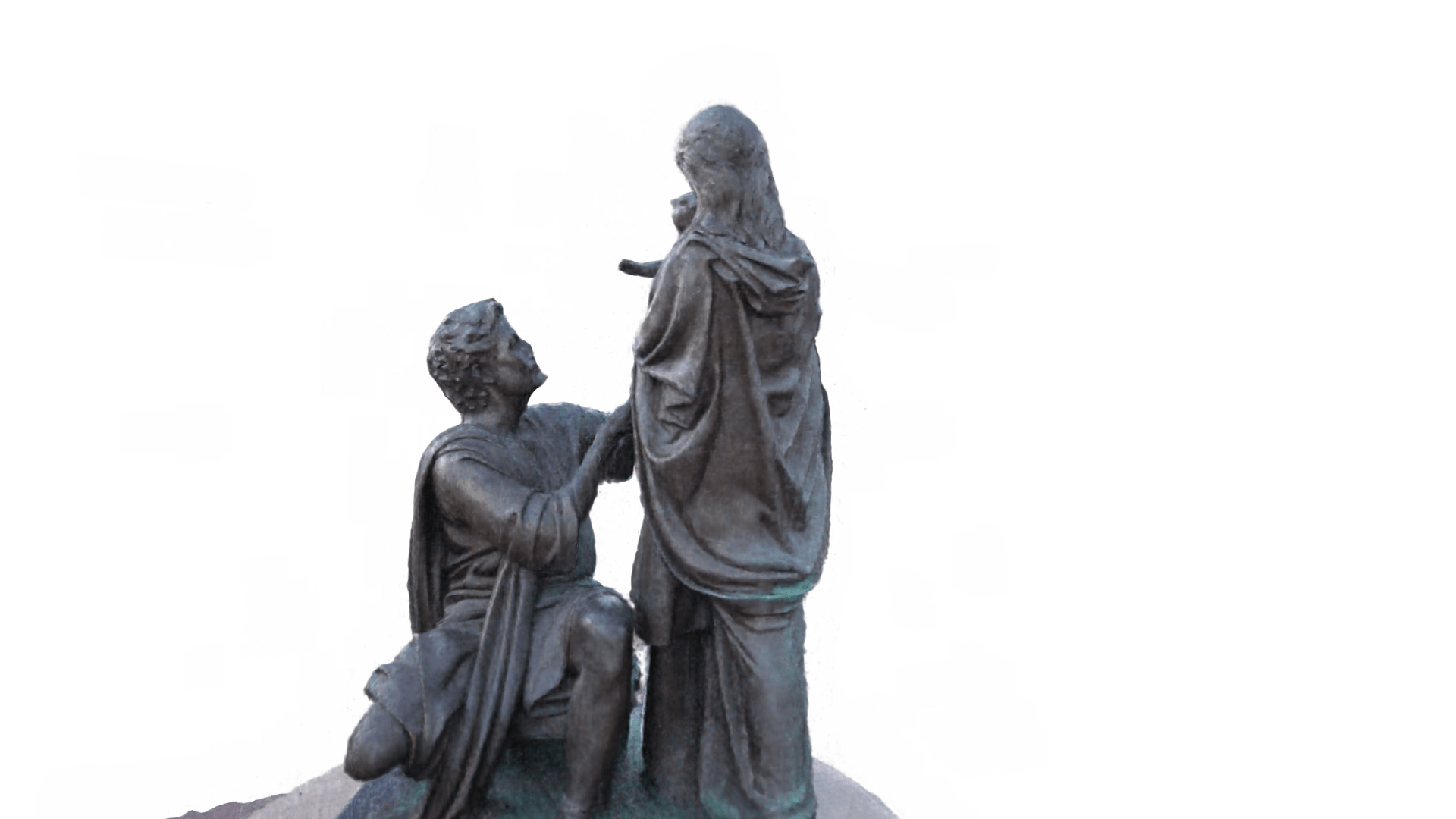} \\
\midrule
           \includegraphics[width=0.24\textwidth]{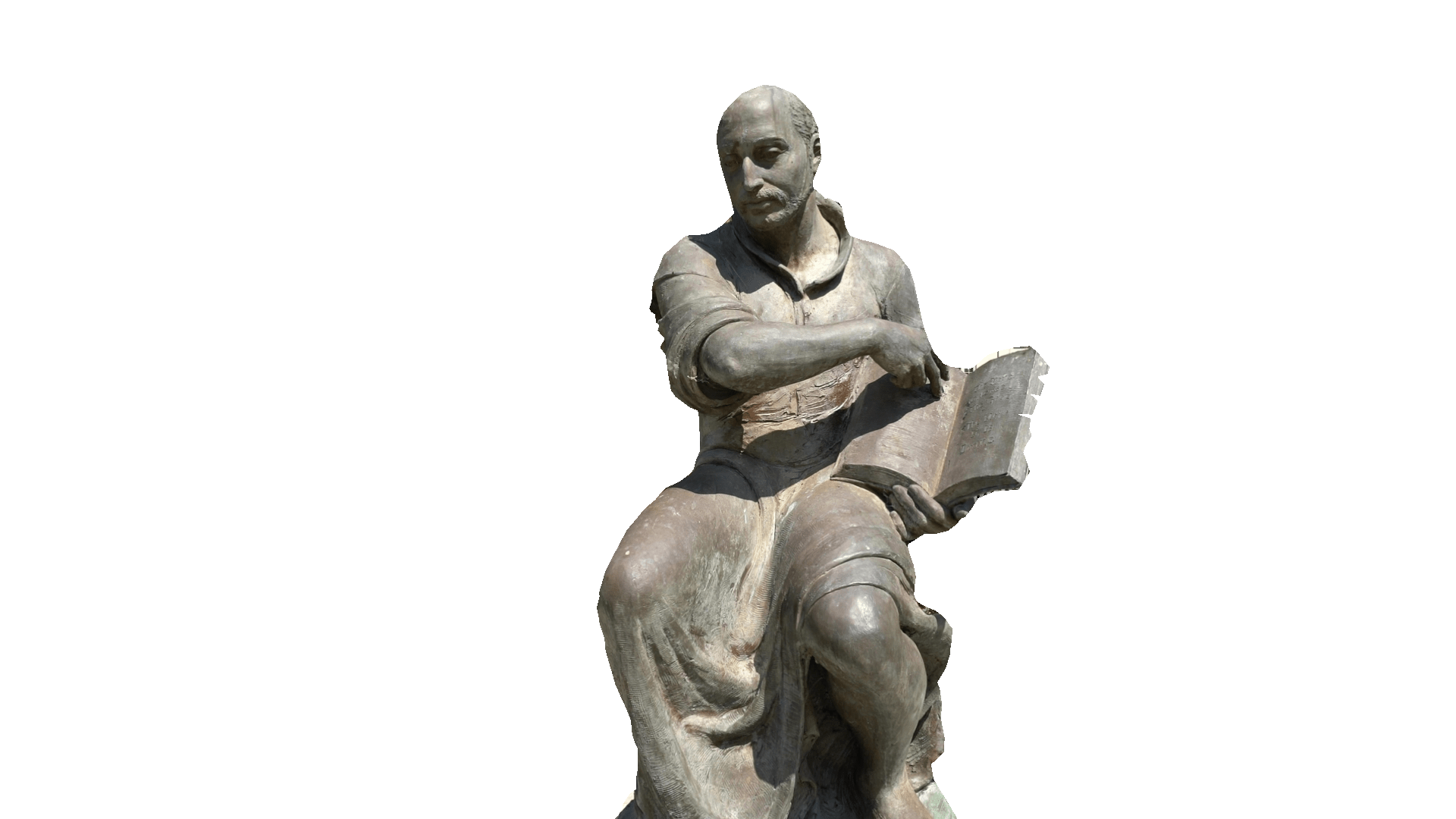}&
      \includegraphics[width=0.24\textwidth]{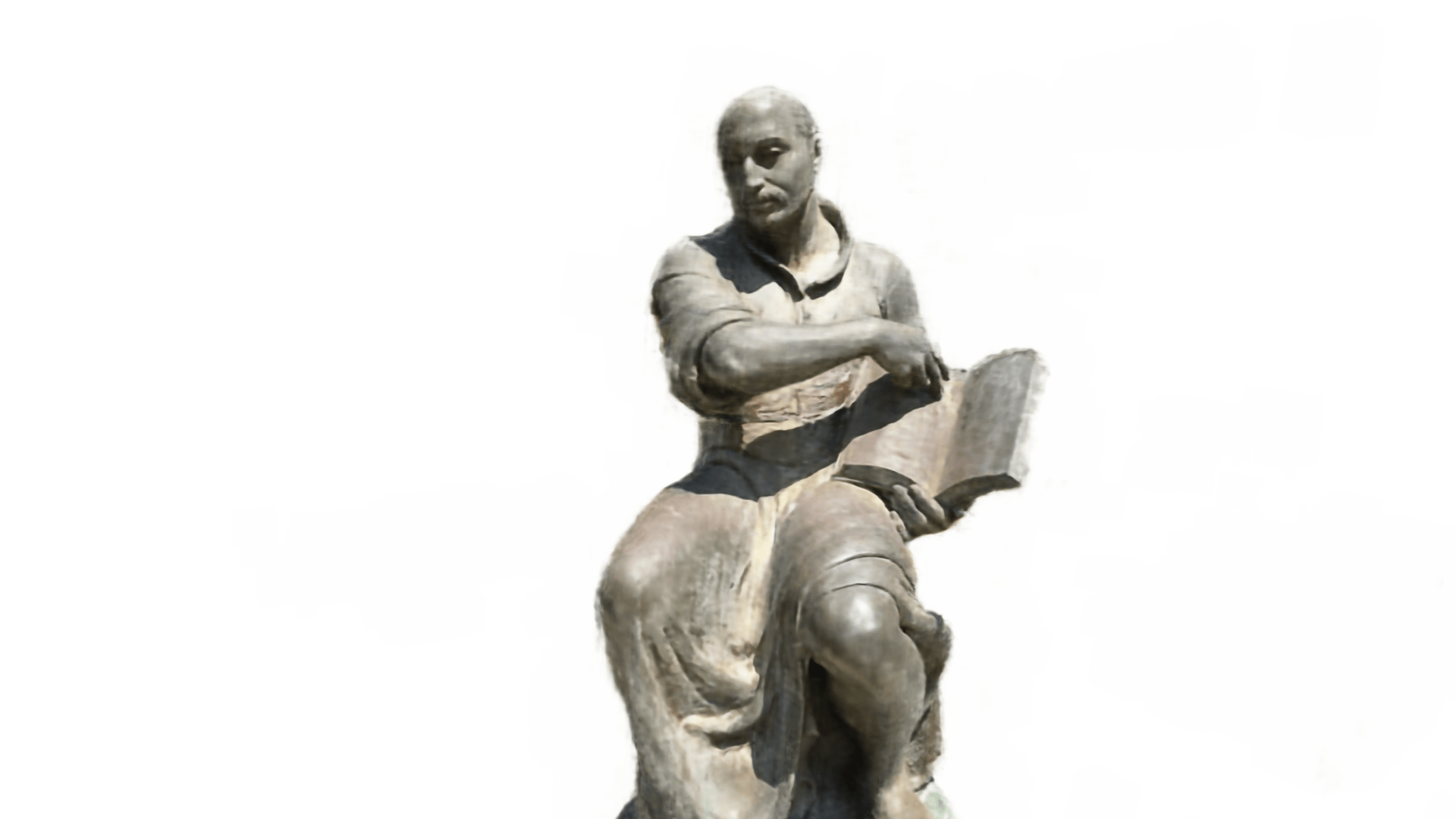} & 
      \includegraphics[width=0.24\textwidth]{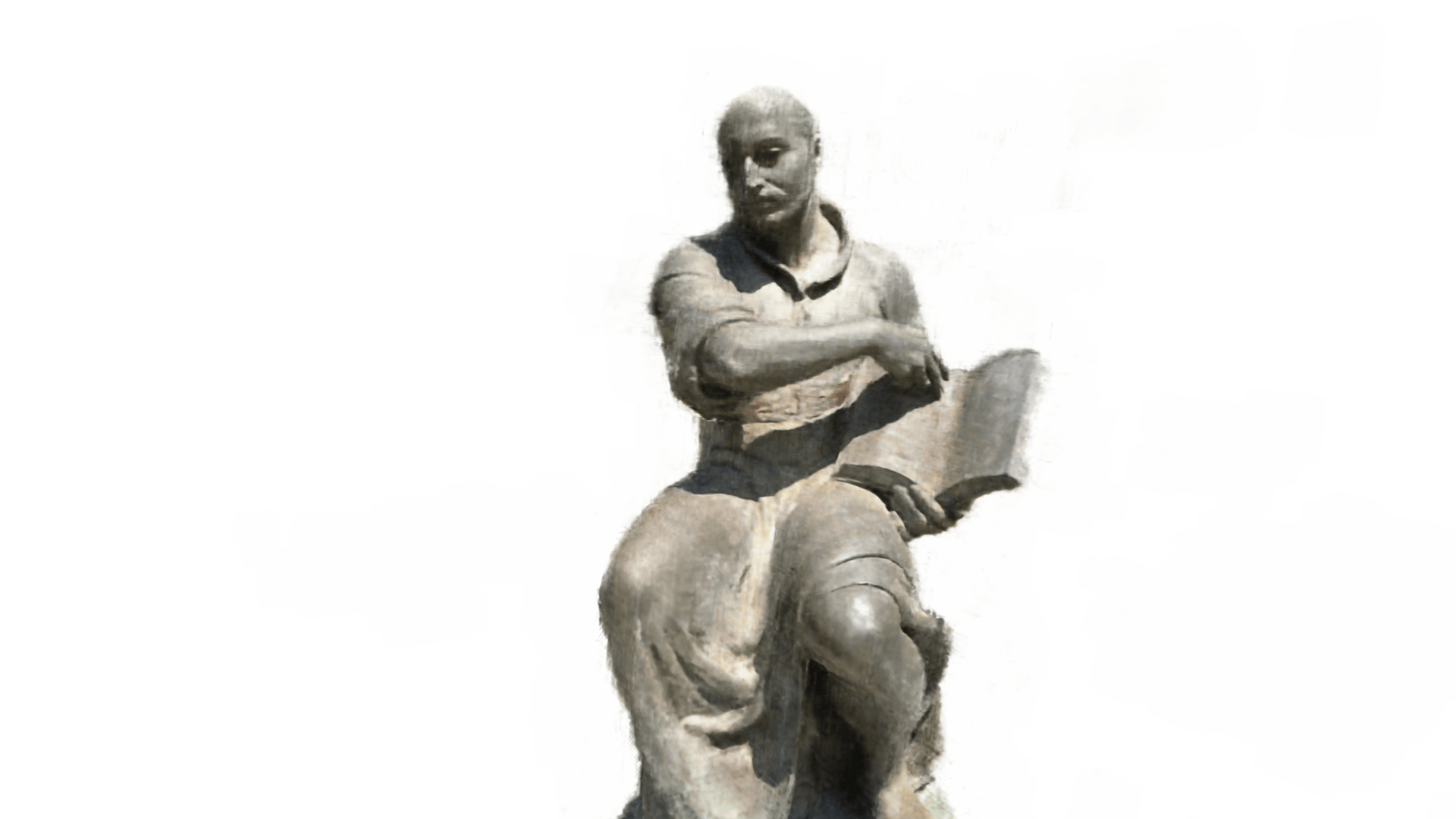} & 
      \includegraphics[width=0.24\textwidth]{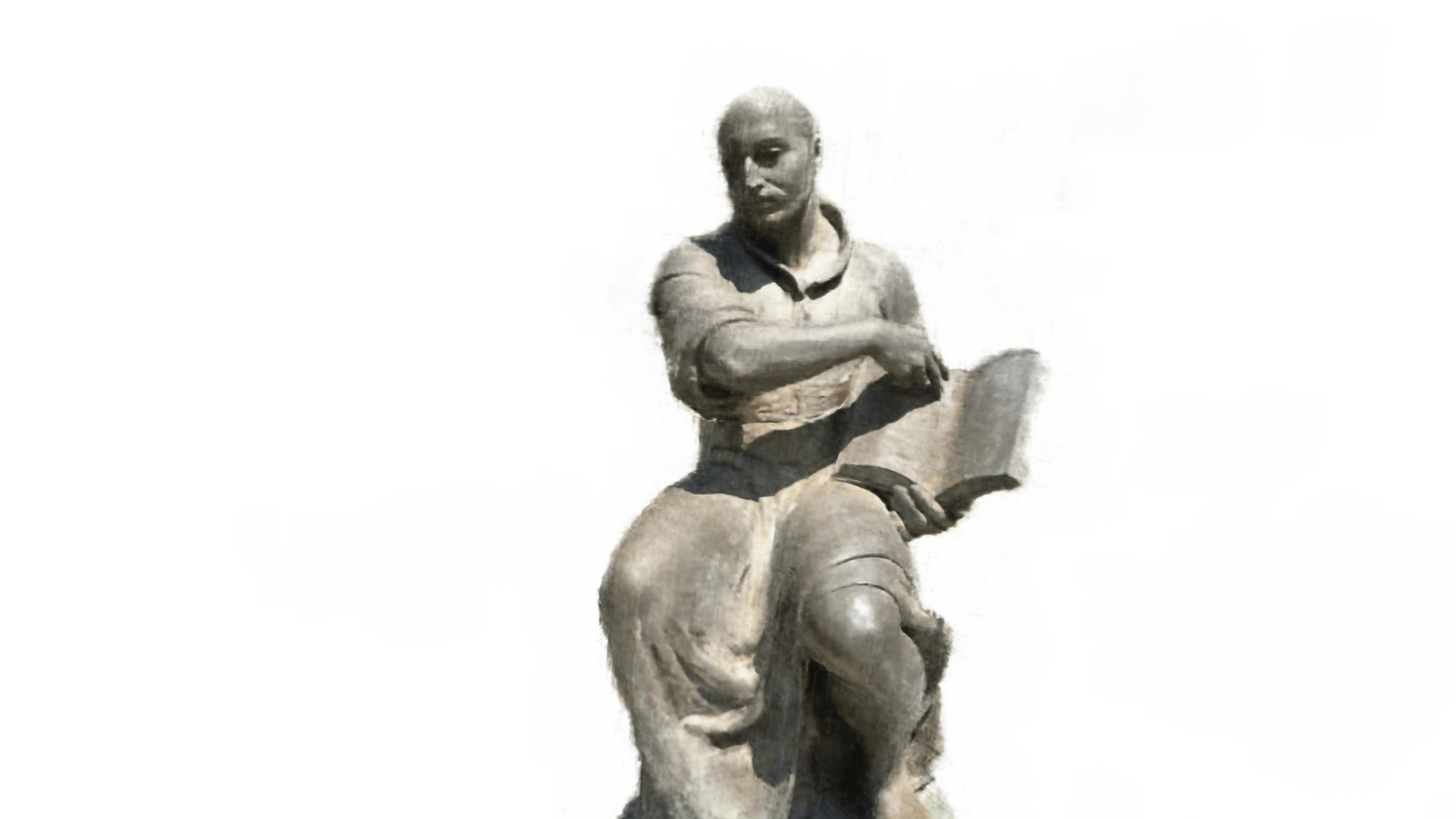} \\
      \midrule
           \includegraphics[width=0.24\textwidth]{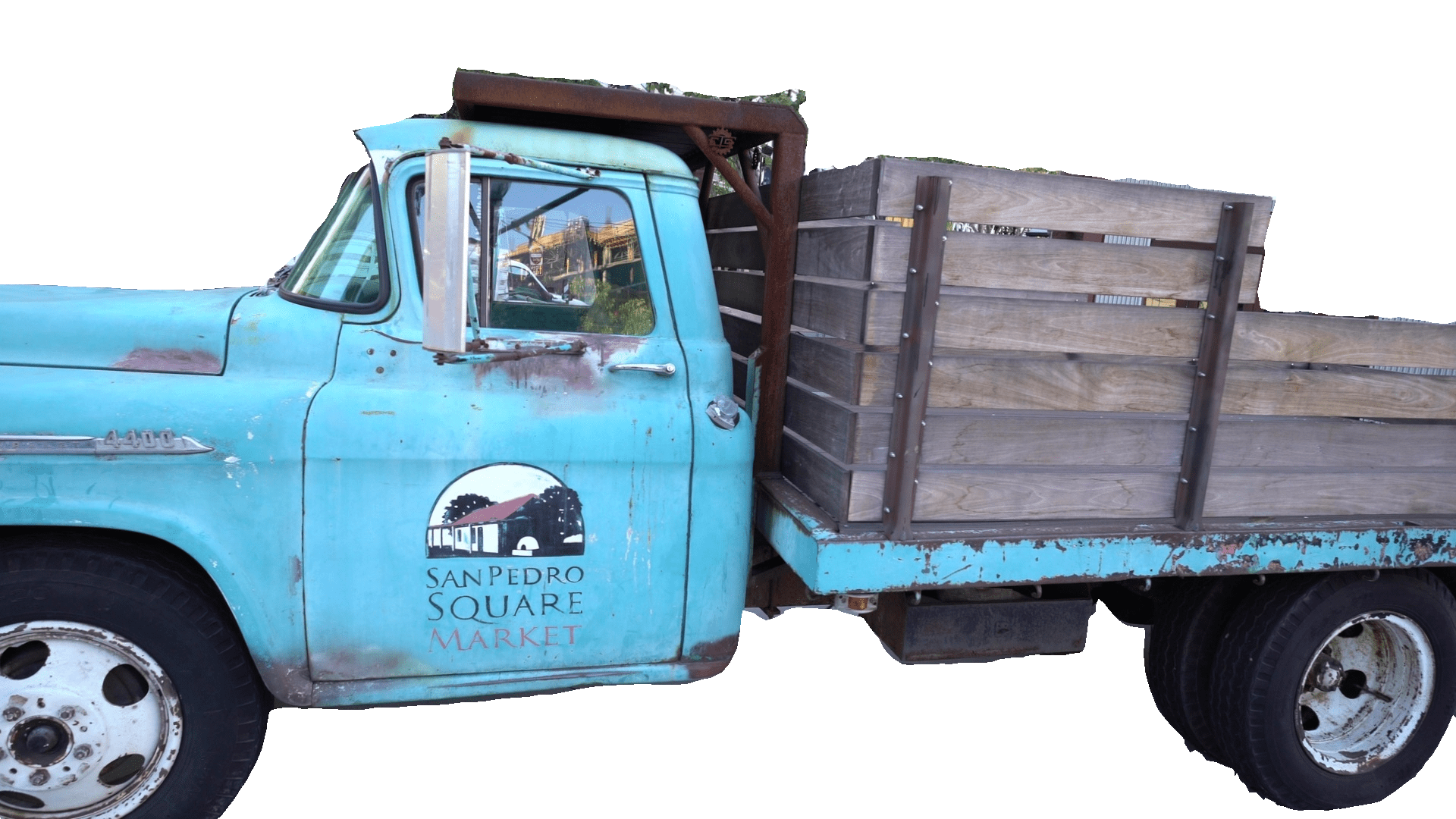}&
      \includegraphics[width=0.24\textwidth]{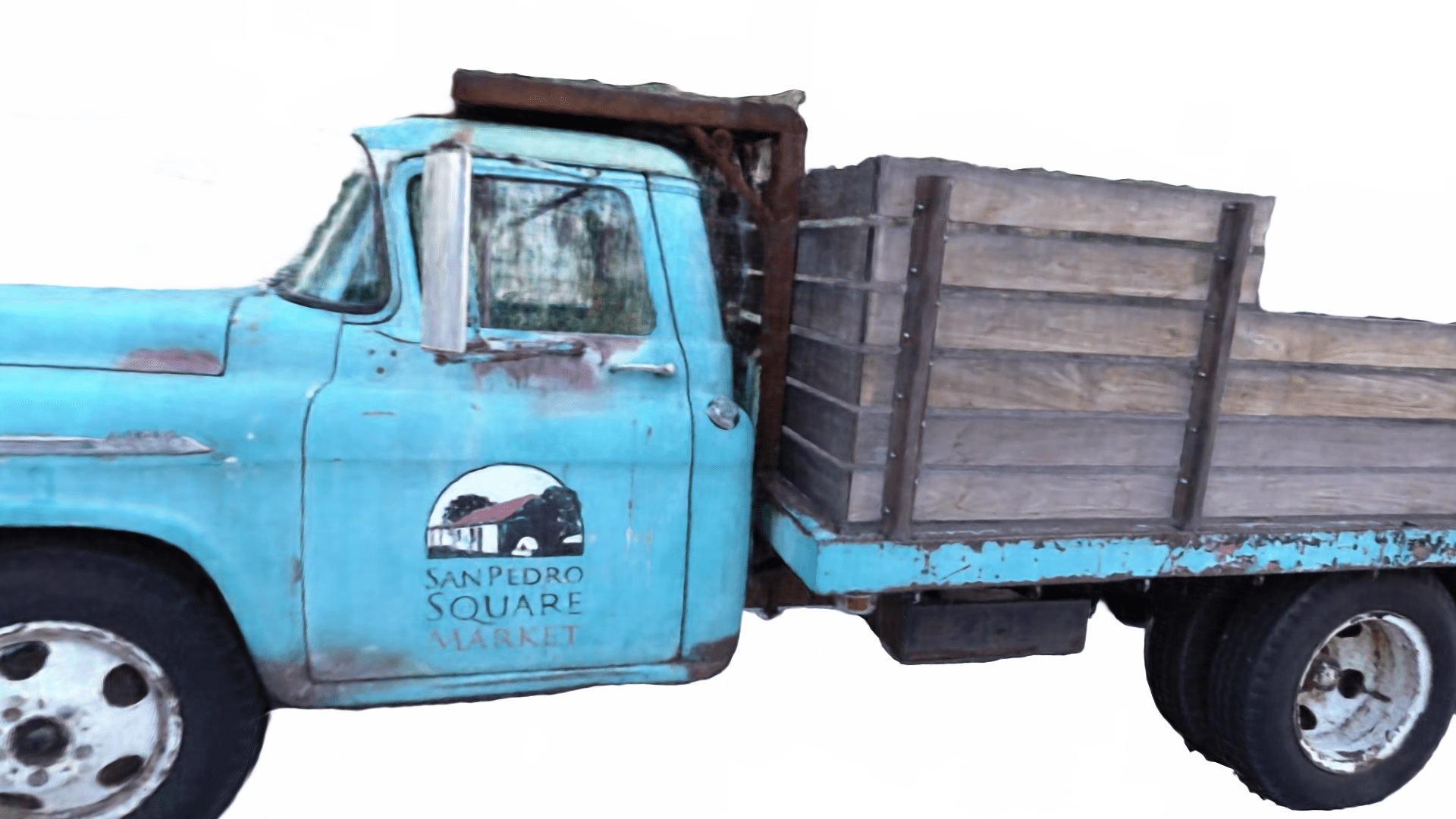} & 
      \includegraphics[width=0.24\textwidth]{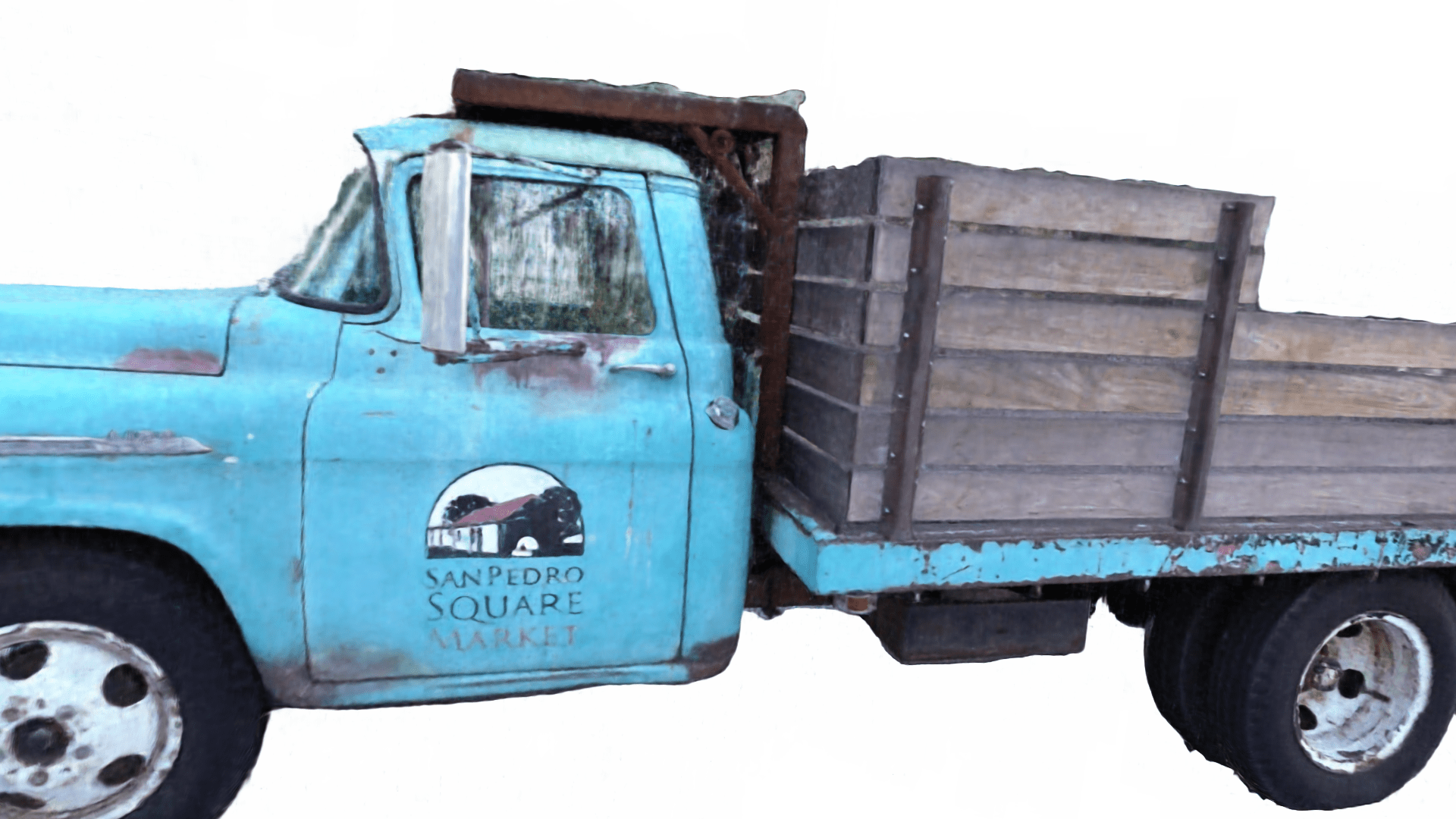} & 
      \includegraphics[width=0.24\textwidth]{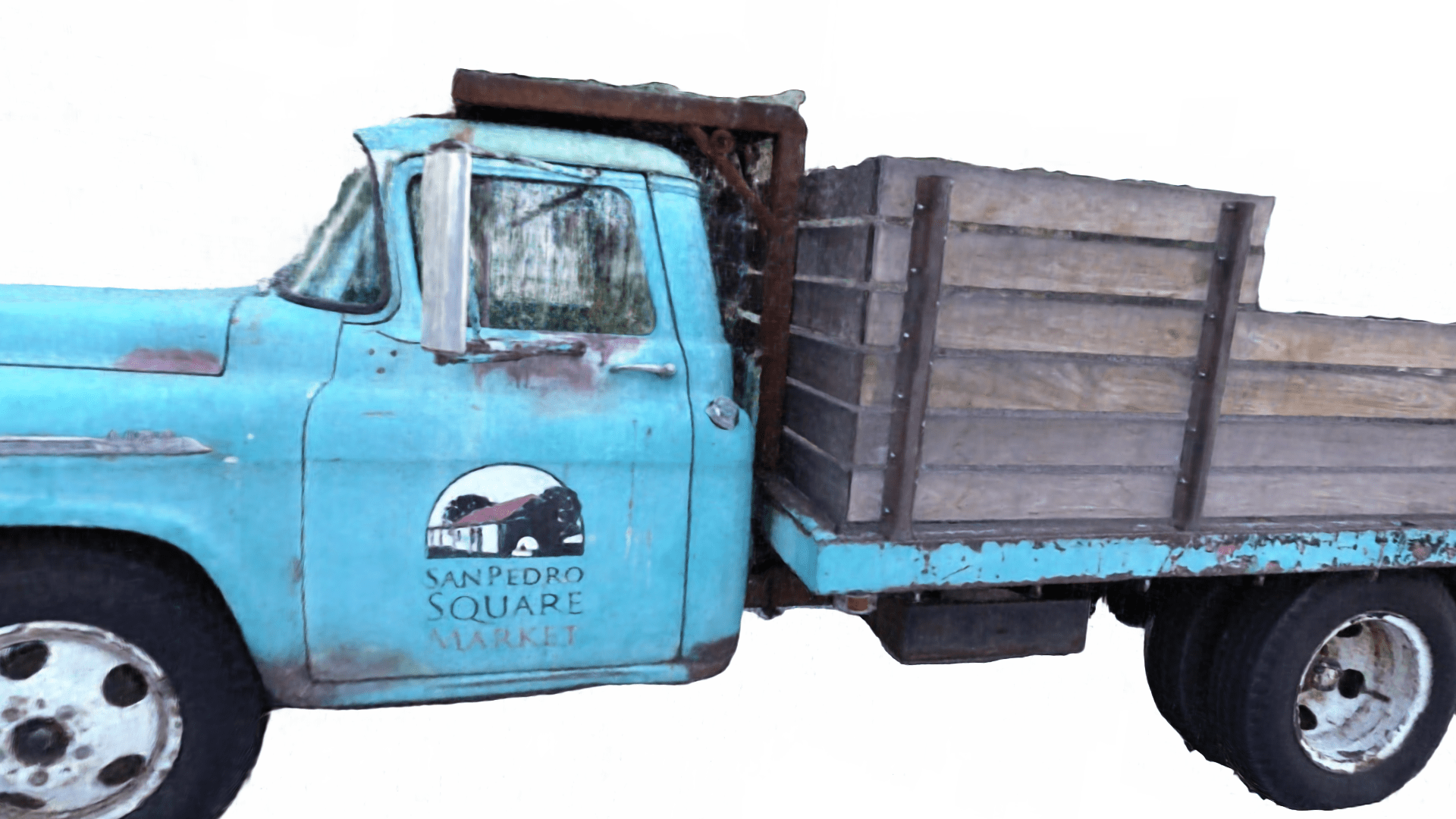} \\
\bottomrule
     \end{tabular}
     }
 \end{table*}

 \begin{table*}
     \caption{Examples generated from the LLFF dataset with TensoRF.}
     \centering
     \resizebox{\textwidth}{!}{
     \begin{tabular}{c c c c}
     \toprule
     \bf Ground Truth & \bf Baseline & \bf LOW compression &\bf HIGH compression\\
     \midrule

               \includegraphics[width=0.24\textwidth]{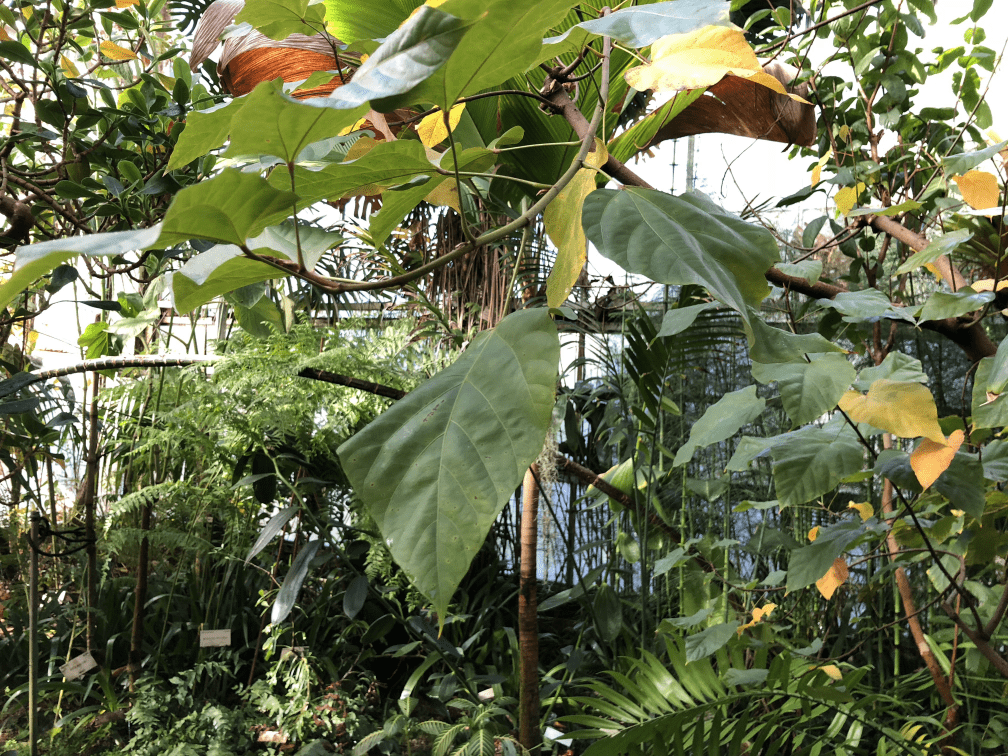}&
      \includegraphics[width=0.24\textwidth]{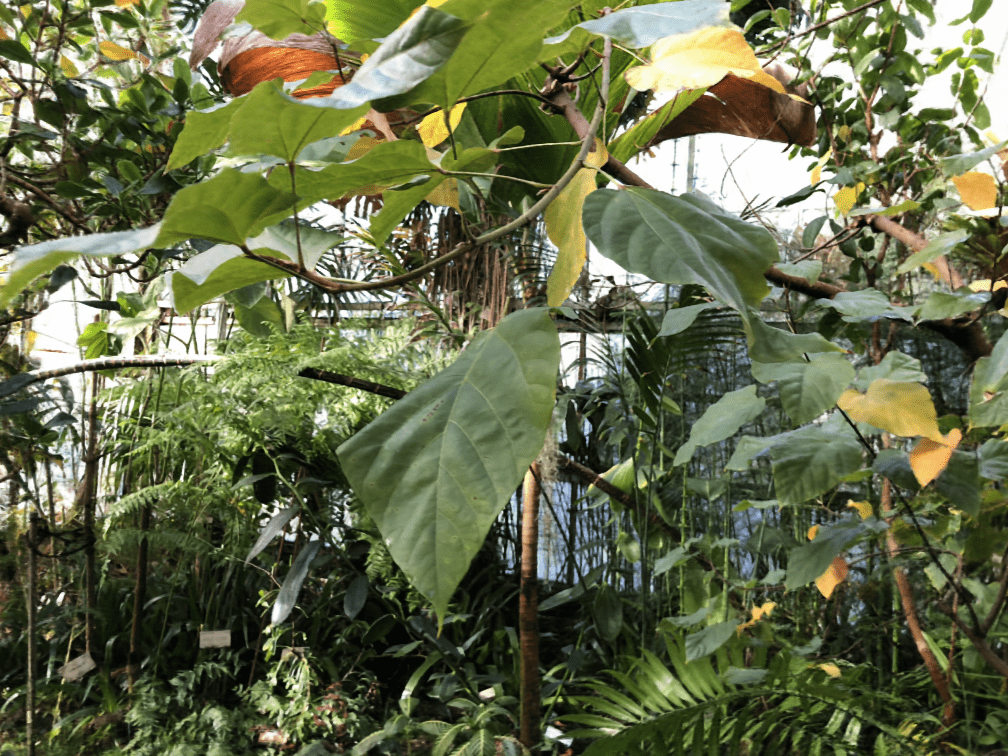} & 
      \includegraphics[width=0.24\textwidth]{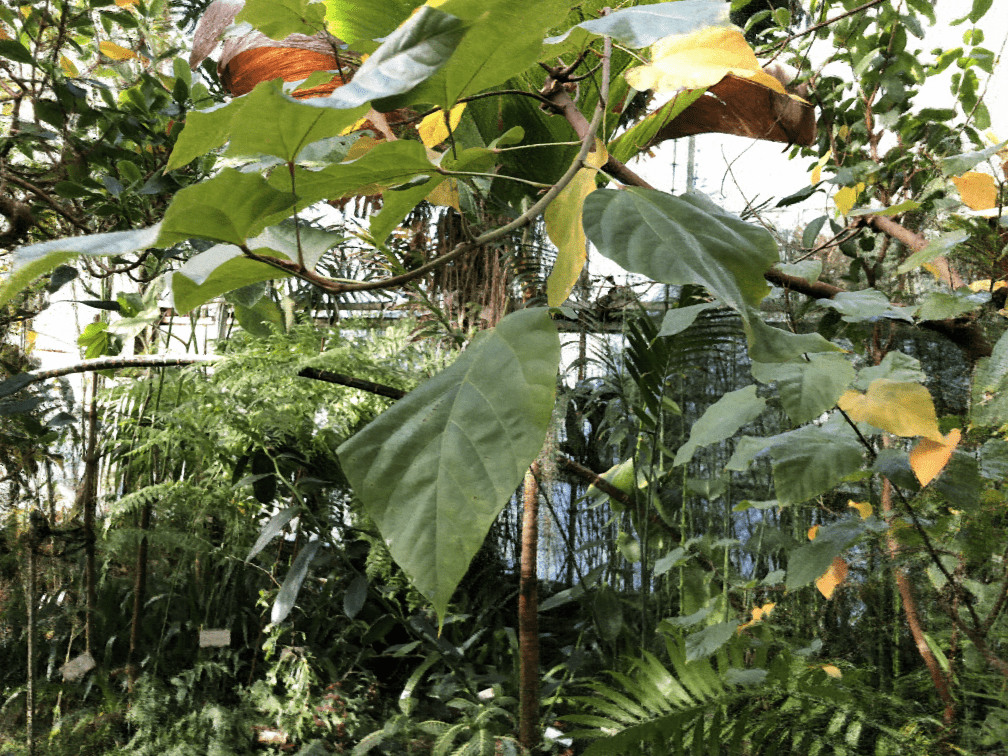} & 
      \includegraphics[width=0.24\textwidth]{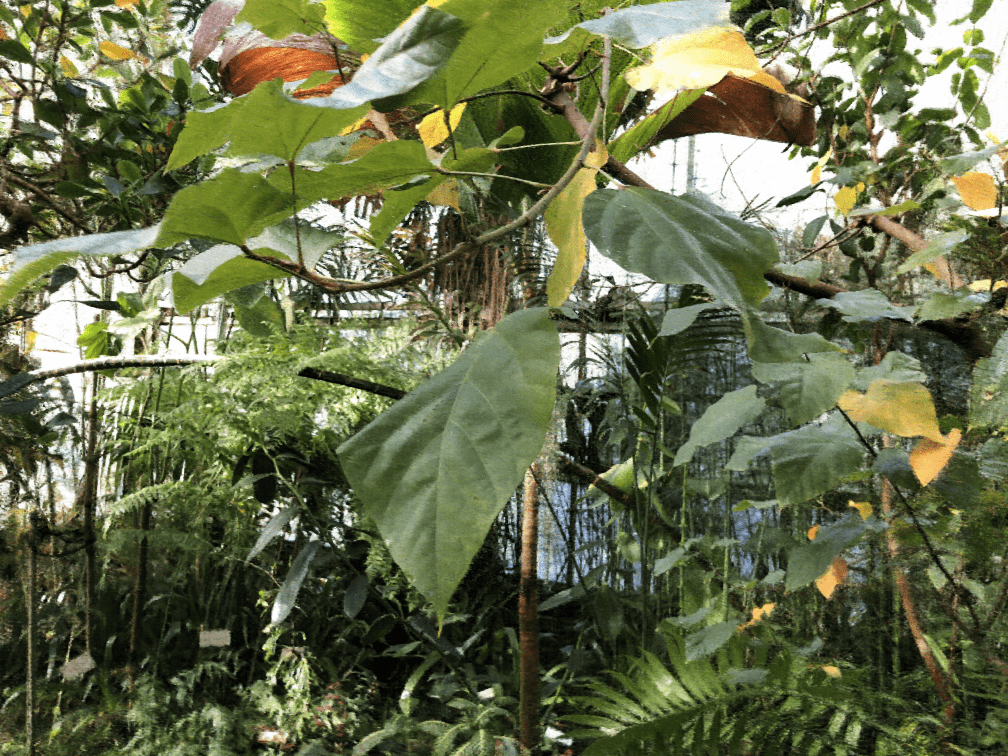} \\
      \midrule
               \includegraphics[width=0.24\textwidth]{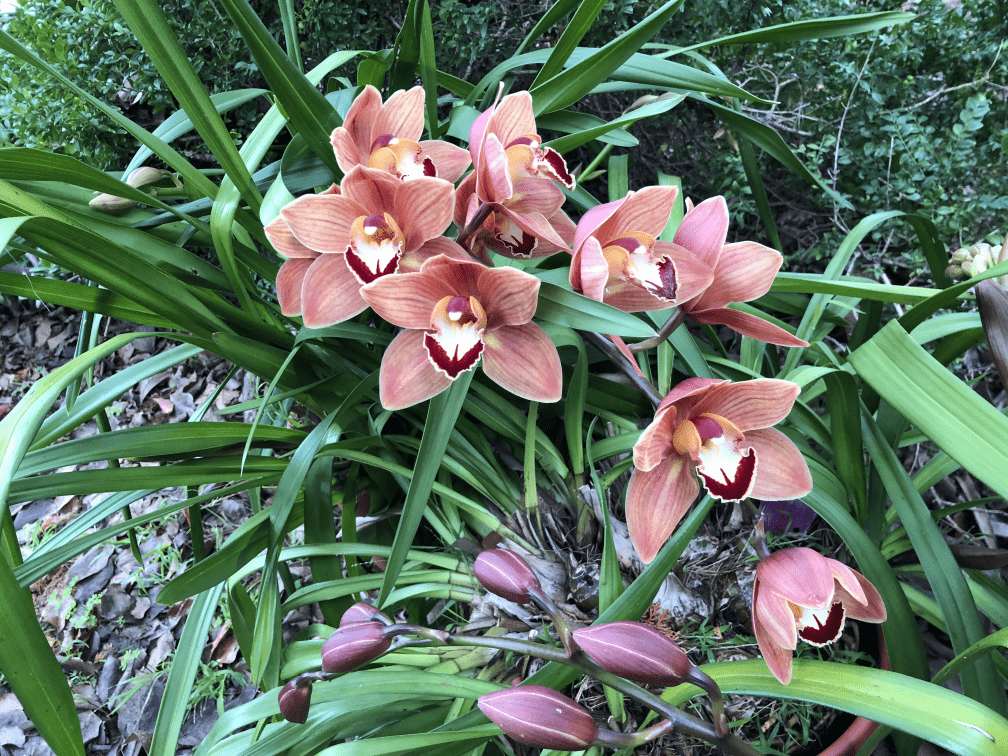}&
      \includegraphics[width=0.24\textwidth]{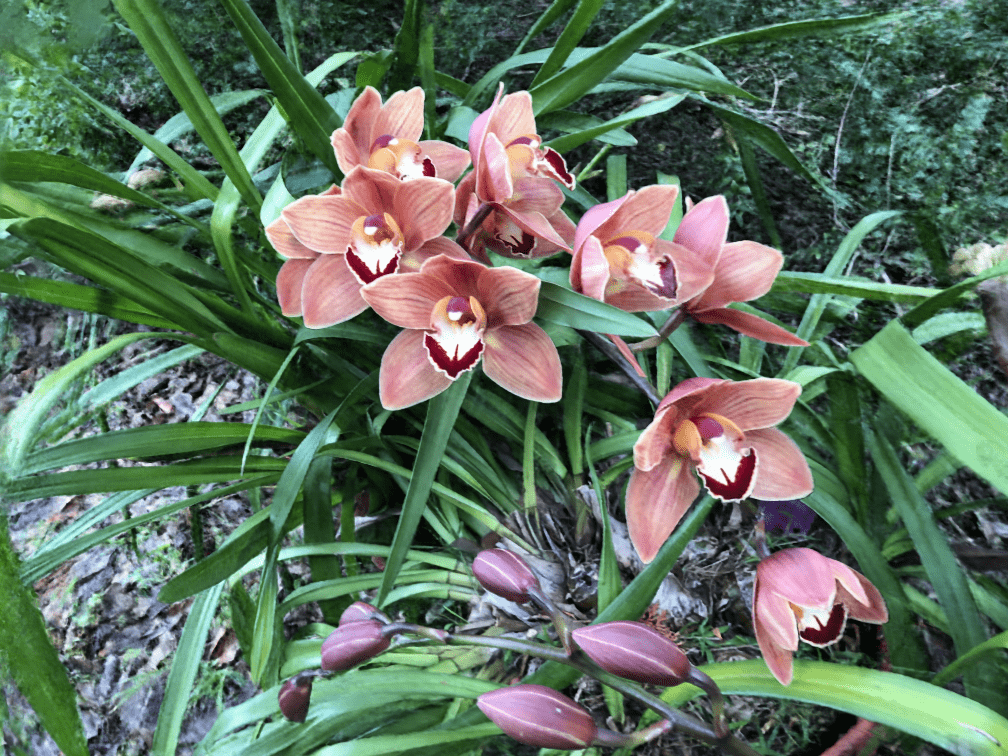} & 
      \includegraphics[width=0.24\textwidth]{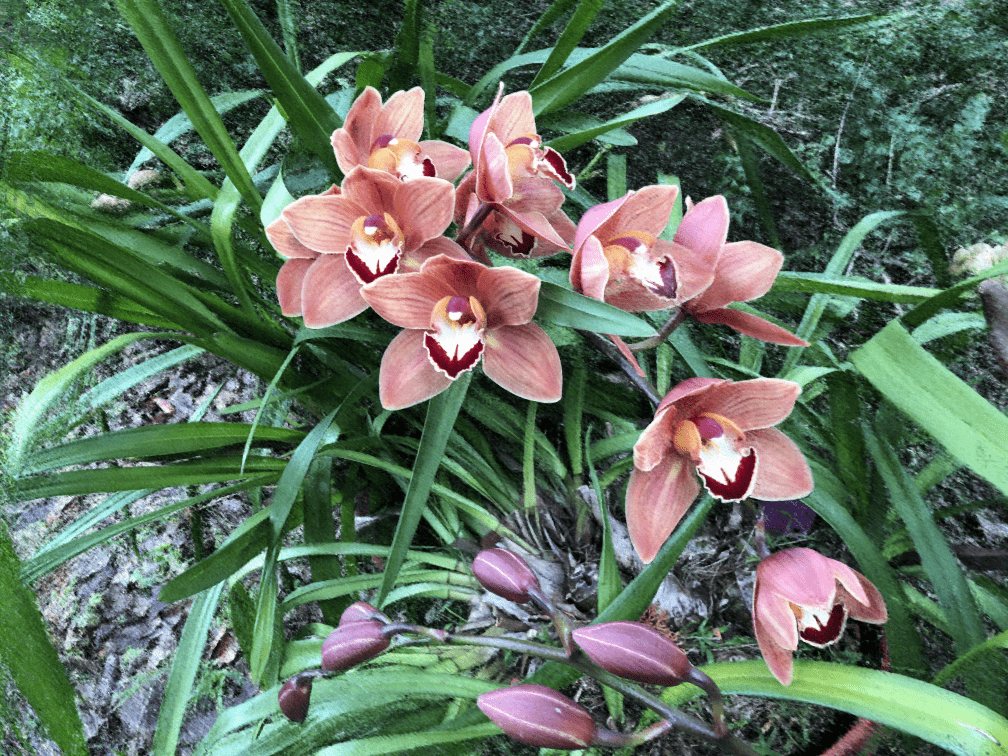} & 
      \includegraphics[width=0.24\textwidth]{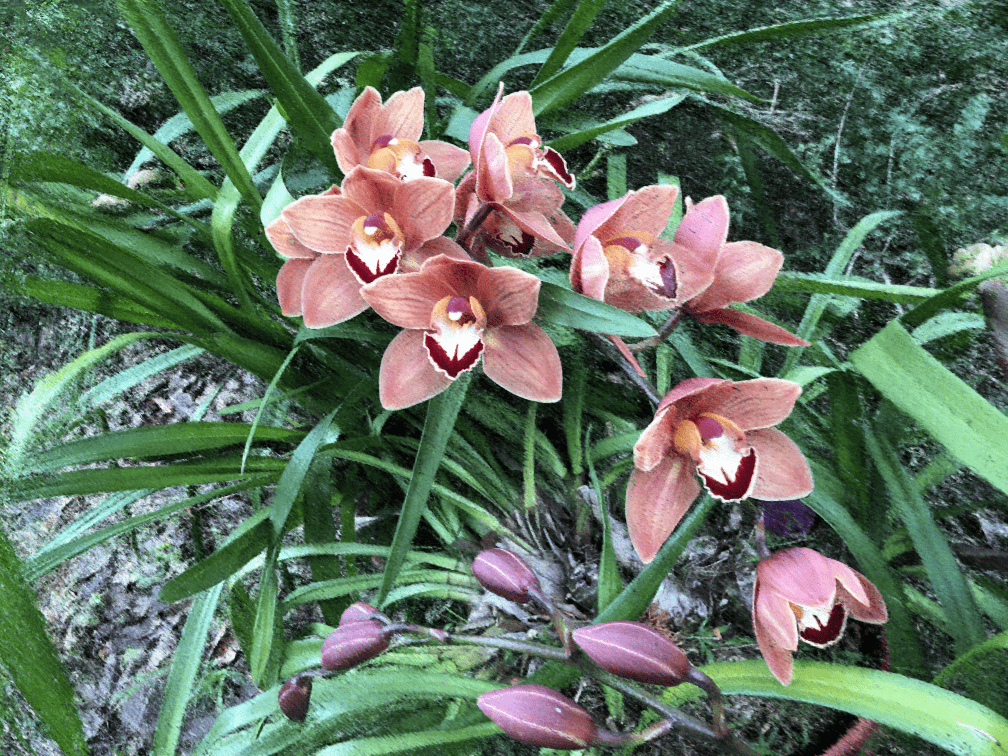} \\
      \midrule
               \includegraphics[width=0.24\textwidth]{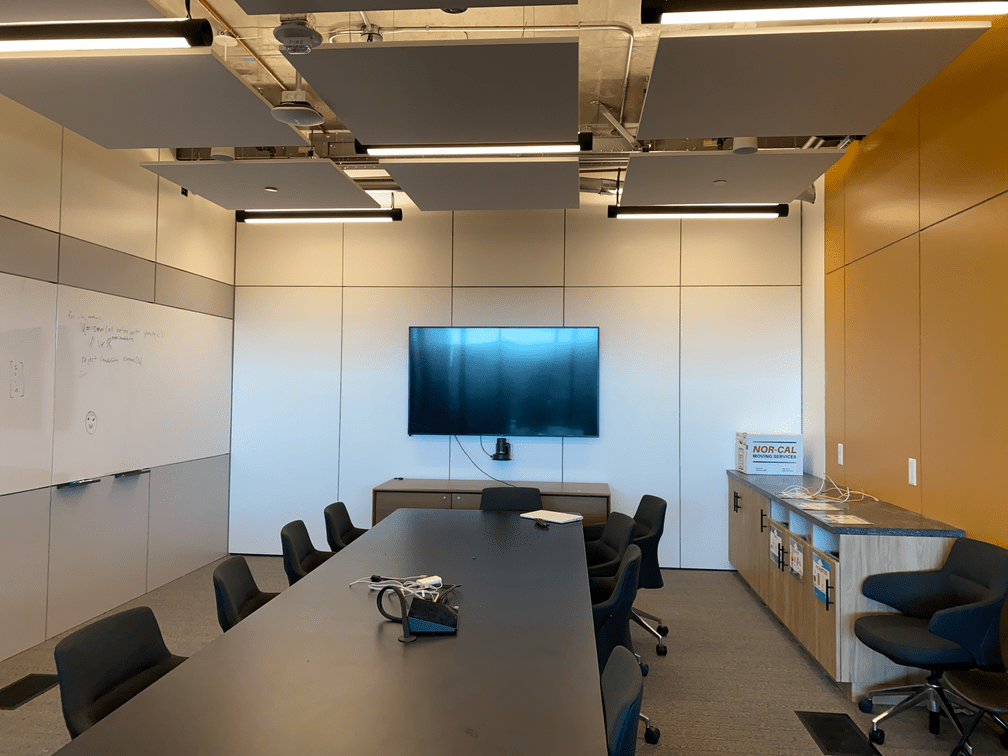}&
      \includegraphics[width=0.24\textwidth]{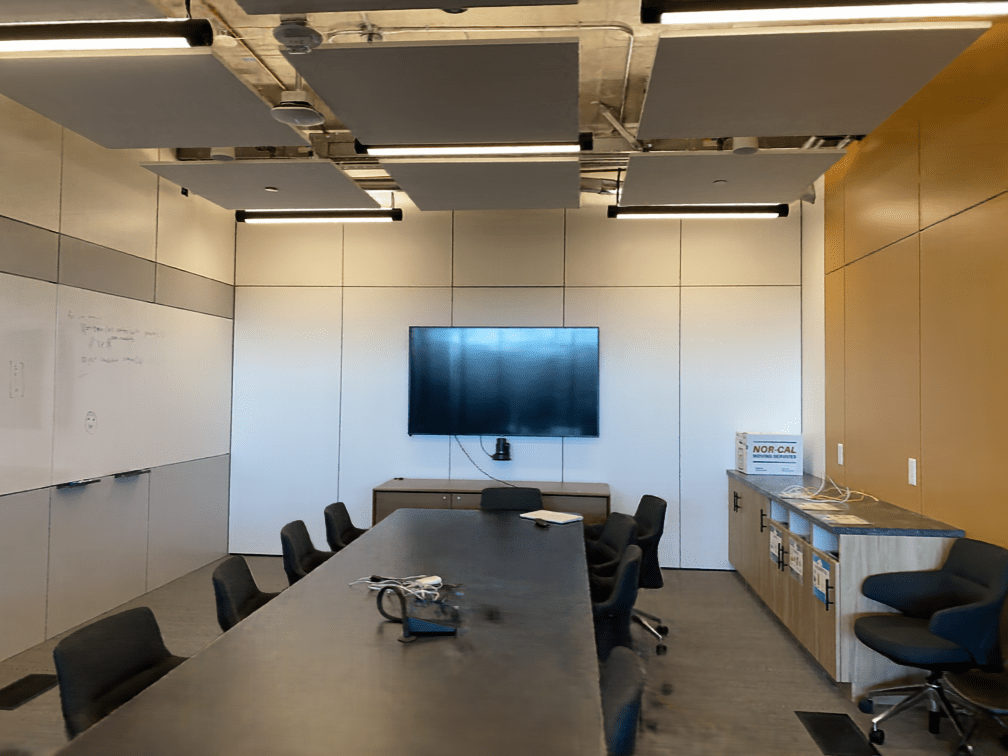} & 
      \includegraphics[width=0.24\textwidth]{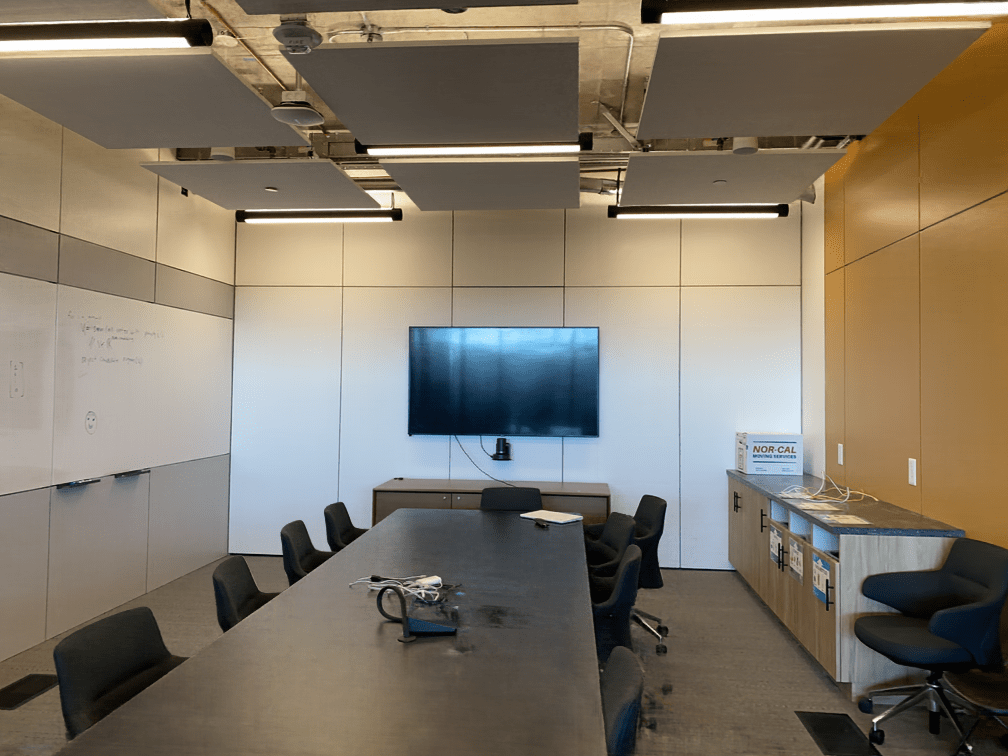} & 
      \includegraphics[width=0.24\textwidth]{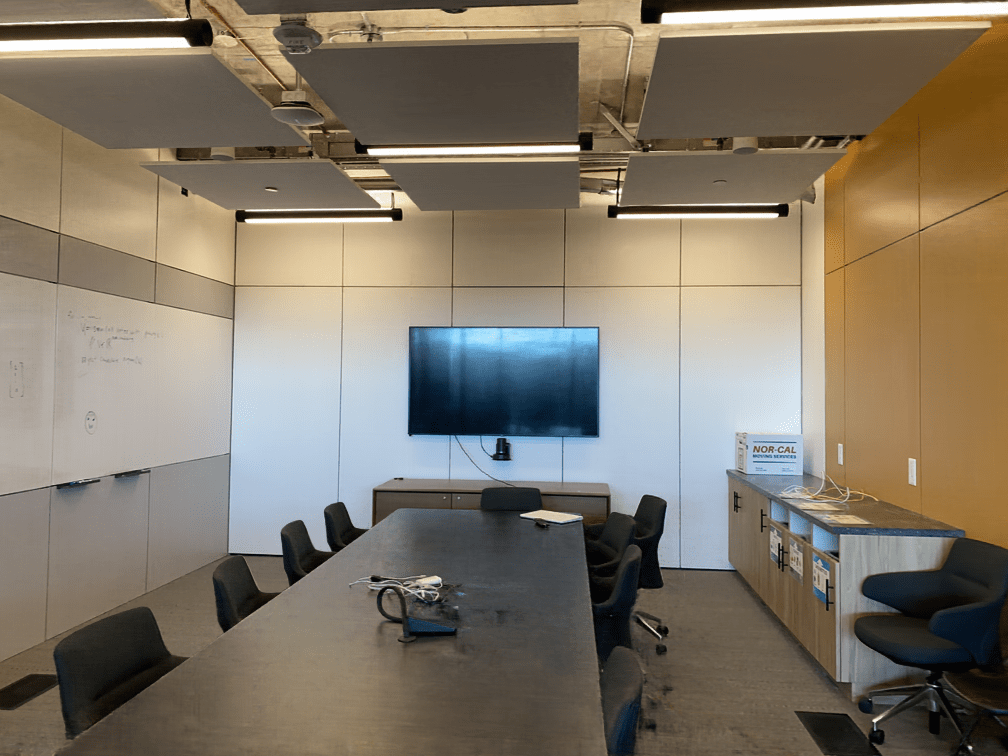} \\
      \midrule
               \includegraphics[width=0.24\textwidth]{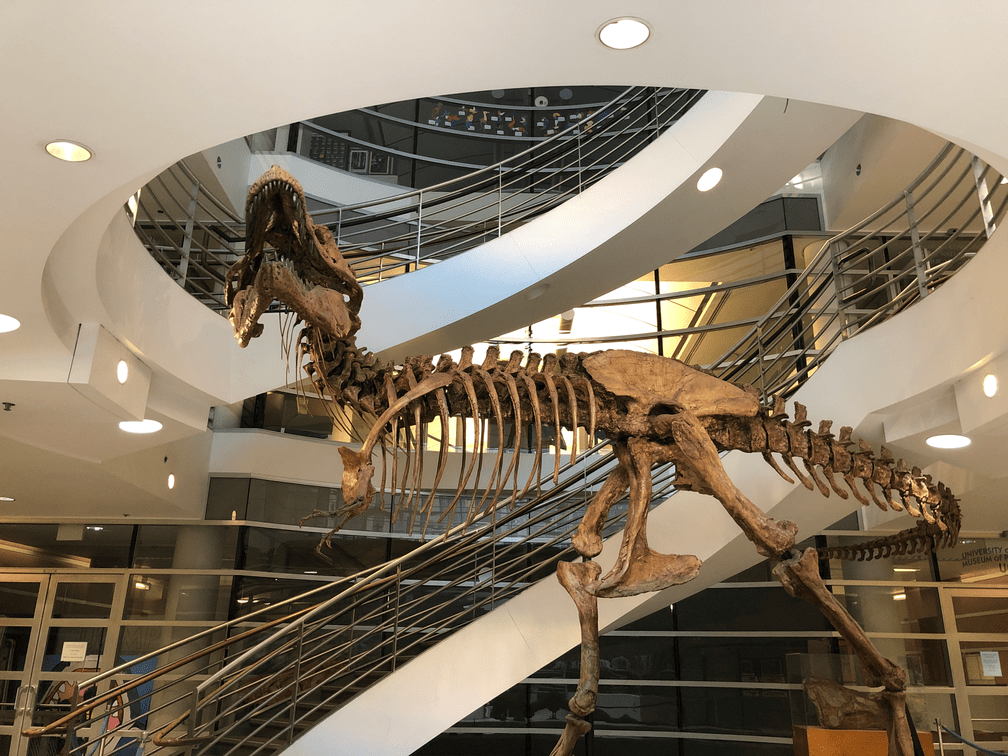}&
      \includegraphics[width=0.24\textwidth]{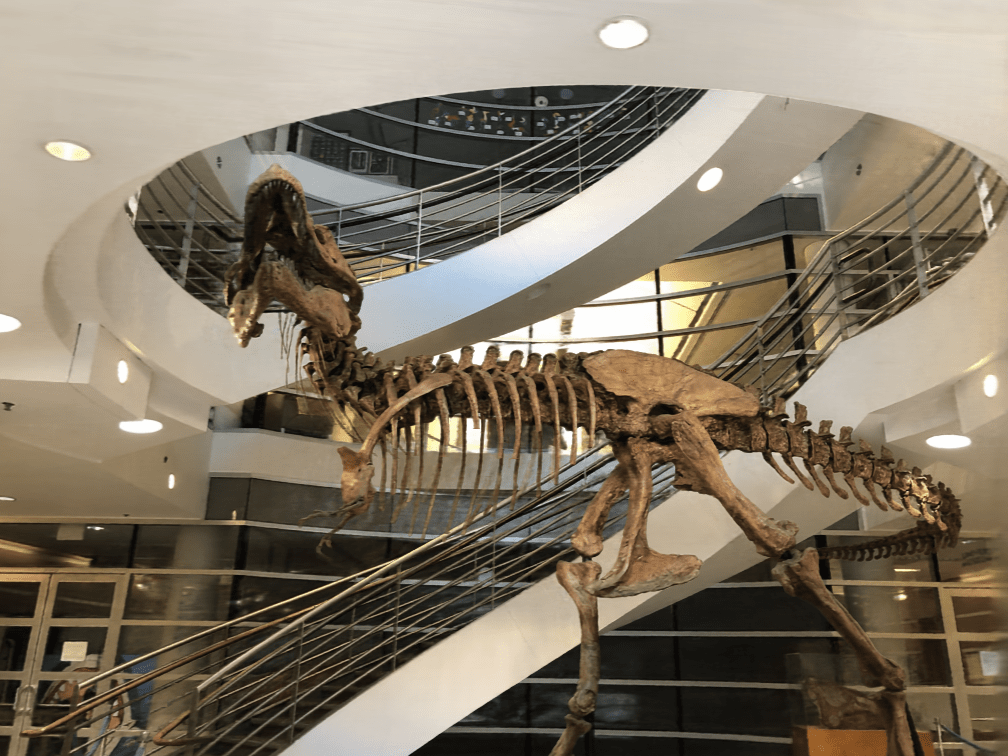} & 
      \includegraphics[width=0.24\textwidth]{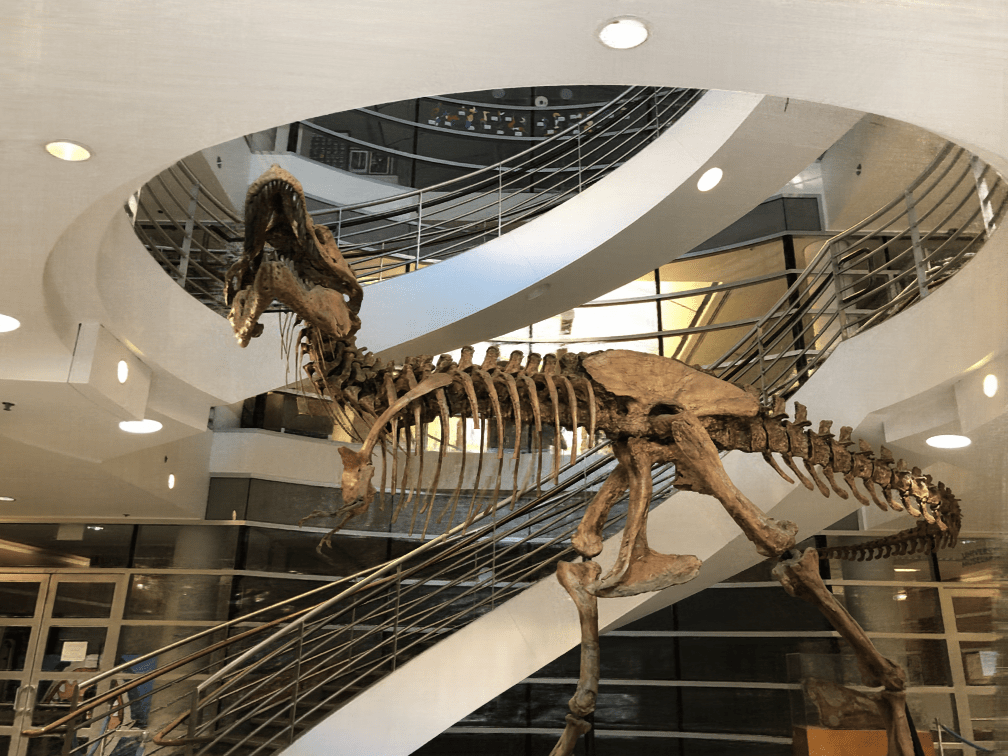} & 
      \includegraphics[width=0.24\textwidth]{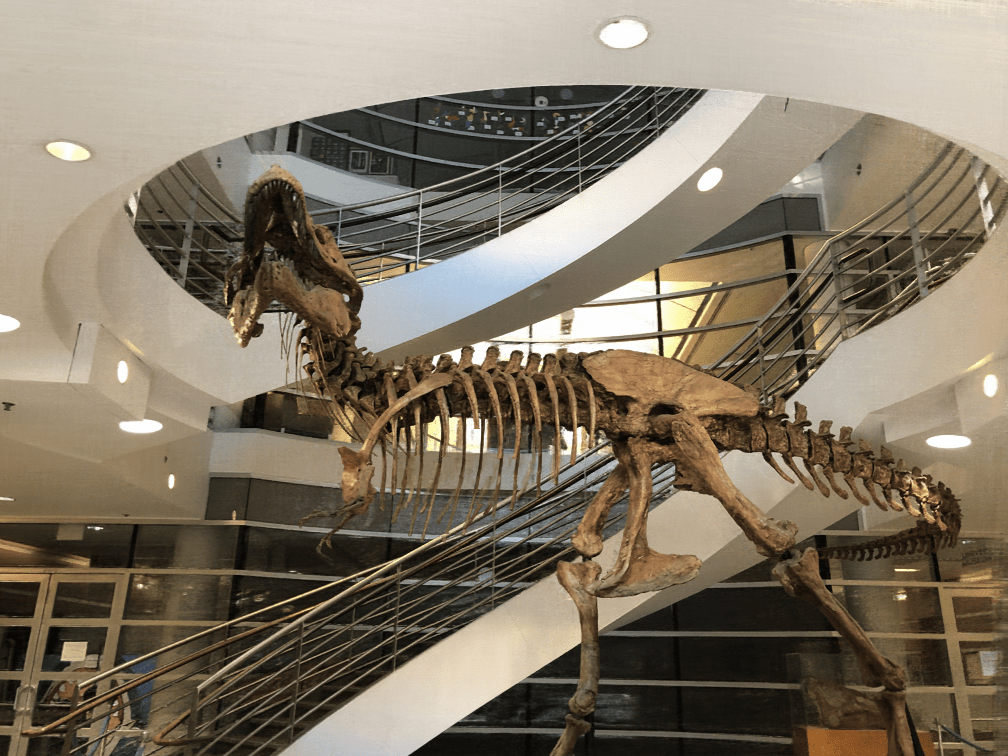} \\
     \bottomrule

     \end{tabular}
     }
 \end{table*}

\end{document}